%% file: main.tex
\documentclass[11pt]{extarticle}
\usepackage{amsmath}
\usepackage{amssymb}
\usepackage{latexsym}
\usepackage{graphics}
\usepackage{epsfig,epsf,rotating}
\usepackage{subfigure}
\usepackage{subcaption}
\usepackage{epsf}

\usepackage[margin=1in]{geometry}

\usepackage{authblk}
\usepackage{fancyhdr}
\usepackage{url}
\usepackage{float}
\usepackage{booktabs}
\usepackage{algorithm}
\usepackage{algpseudocode}
\usepackage{xcolor}
\usepackage{caption}
\usepackage{setspace}
\usepackage[dvipsnames]{xcolor}

\usepackage{placeins}

\usepackage{theorem}

\theoremheaderfont{\itshape} {\theoremstyle{break}
} \theoremstyle{break}
\newtheorem{lemma}{Lemma}[section] \theoremstyle{break}
 {\theoremstyle{plain}
  \theorembodyfont{\rmfamily}  \newtheorem{proof}{Proof}[section]}
{\theoremstyle{plain}
  \theorembodyfont{\rmfamily}  \newtheorem{definition}{Definition}[section]}
\input{macros}
\newcommand{\qed}{$\blacksquare$}

\title{TIDE: A~Trace-Informed Depth-First~Exploration\\ for Planning with Temporally Extended Goals}
\author{Yuliia Suprun$^{1}$, Khen Elimelech$^{1}$, Lydia E. Kavraki$^{1,2}$, and~Moshe~Y.~Vardi$^{1,2}$}
\affil{$^{1}$Department of Computer Science, Rice University, USA \\
$^{2}$Ken Kennedy Institute, Rice University, USA\\
}

\date{May 2025}

\begin{document}
\maketitle
\thispagestyle{fancy}

\input{abstract}

\input{intro}
\input{related_work}
\input{problem_definition}

\input{preliminaries}
\input{proposed_method}
\input{experiment_results}

\input{conclusion}
\input{ack}

% \appendix
% \addcontentsline{toc} {chapter}{\numberline {}Appendix A}
% \include{appendix}
% \include{append-b}
% \addcontentsline{toc} {chapter}{\numberline {}Bibliography}{}
% \include{biblio}

\bibliographystyle{ieeetr}
\bibliography{references}
% \bibliography{PhD_bibliography}
% \appendix
\input{appendix}
\end{document}

%% file: macros.tex
 \renewcommand{\P}{\mathcal{P}}

\newcommand{\true}{\mathit{true}}

\newcommand{\Nat}{{\rm I\kern-.23em N}}
\newcommand{\Prop}{\P}

%% file: abstract.tex
\thispagestyle{empty}
\begin{abstract}
Task planning with temporally extended goals (TEGs) is a critical challenge in AI and robotics, enabling agents to achieve complex sequences of objectives over time rather than addressing isolated, immediate tasks. Linear Temporal Logic on finite traces (LTL$_f$) provides a robust formalism for encoding these temporal goals. Traditional LTL$_f$ task planning approaches often transform the temporal planning problem into a classical planning problem with reachability goals, which are then solved using off-the-shelf planners. However, these methods often lack informed heuristics to provide a guided search for temporal goals.

We introduce TIDE (Trace-Informed Depth-first Exploration), a novel approach that addresses this limitation by decomposing a temporal problem into a sequence of smaller, manageable reach-avoid subproblems, each solvable using an off-the-shelf planner. TIDE identifies and prioritizes promising automaton traces within the domain graph, using cost-driven heuristics to guide exploration. Its adaptive backtracking mechanism systematically recovers from failed plans by recalculating costs and penalizing infeasible transitions, ensuring completeness and efficiency. Experimental results demonstrate that TIDE achieves promising performance and is a valuable addition to the portfolio of planning methods for temporally extended goals.
\end{abstract}

%% file: intro.tex
\section{Introduction}
\label{ch:Intro}

\subsection{Background}
Task planning is a foundational problem in robotics and artificial intelligence, where autonomous agents must generate sequences of actions that achieve specific objectives. In traditional planning, these objectives are typically defined as \textit{final-state goals}—conditions that must hold true in a particular state of the world, specifically the final state of an action sequence. While effective for many applications, this model falls short for applications where agents must adhere to behavioral patterns or constraints that span across multiple states (or time steps).

To address such needs, researchers have explored \textit{temporally extended goals} (TEGs), which describe properties that must hold over sequences of states rather than a single terminal state. A widely adopted formalism for encoding TEGs is Linear Temporal Logic (LTL), which augments propositional logic with temporal operators like ``eventually'', ``always'', and ``until''. LTL is traditionally interpreted over infinite sequences, but many real-world tasks are naturally bounded in time. For these finite-horizon scenarios, the variant known as \textit{Linear Temporal Logic on finite traces} (LTL$_f$) has gained traction.

At a high level, the planning problem with an LTL$_f$ goal can be described as follows:  
Given a domain modeled as a transition system (describing how actions move the agent between states) and a temporally extended goal expressed in LTL$_f$, the task is to find a sequence of actions that produces a sequence of states satisfying the goal formula. In other words, rather than aiming for a particular final configuration, the plan must generate an entire execution trace that adheres to the specified temporal constraints.

The standard approach to planning with LTL$_f$ goals typically involves three steps:
\begin{enumerate}
    \item \textbf{Translate} the LTL$_f$ goal formula into a finite automaton that accepts exactly the traces satisfying the goal.
    \item \textbf{Construct} a product graph by computing a cross-product of the automaton and the domain's transition system.
    \item \textbf{Search} the product graph to find a path that leads to an accepting state of the automaton, thereby satisfying the LTL$_f$ formula. 
\end{enumerate}

This approach allows for elegant formal reasoning, but suffers from significant scalability issues when applied to large complex planning domains.

\subsection{Motivation}
The ability to plan for temporally extended goals is critical in domains where tasks are sequential or dependent on some safety and liveness properties. Consider, for example, a household robot tasked with delivering items in a specific order, or a drone that must survey regions while avoiding restricted zones. These tasks cannot be succinctly expressed using simple final-state goals defined with propositional logic; they require richer ``temporal'' constraints.

While the expressive power of LTL$_f$ makes it suitable for such tasks, the computational demands of current planning techniques often limit its practical use. As planning domains grow in complexity, existing methods struggle to scale, primarily due to the combinatorial explosion of the product graph and the limitations of general-purpose search algorithms. Classical off-the-shelf planners, though robust and domain-independent, are typically unaware of the structure imposed by the product automaton and therefore cannot exploit it to guide the search more effectively.

\subsection{The Challenge of Scalability}

A core difficulty in LTL$_f$ planning lies in constructing and exploring the product graph, which represents the cross-product of the domain's state space and the automaton derived from the LTL$_f$ formula. This product graph can be exponentially large, and most traditional approaches rely on some form of breadth-first graph search to find a valid plan. But without informative heuristics, this search tends to be poorly guided and computationally expensive.

What makes things worse is that many existing planners treat the entire product graph as a single, monolithic structure. They convert the whole planning problem into one massive reachability task, which might be fine in theory, but may quickly break down in practice. As the size of the domain grows, or as the automaton admits many accepting paths, the state space becomes overwhelming—especially in domains with high branching factors or loosely structured temporal goals.

\subsection{Proposed Solution: TIDE}
This article introduces \textit{TIDE (Trace-Informed Depth-first Exploration)}, a novel planning framework designed to address the limitations of existing LTL$_f$ planning methods. Instead of constructing and searching the full product graph up front, TIDE incrementally explores the graph in a trace-guided manner. It focuses on one automaton trace at a time—effectively treating the high-level LTL$_f$ goal as a sequence of intermediate reach-avoid goals—and attempts to realize that trace using a classical planner.

The core idea of TIDE is to reduce the planning problem into a series of smaller subproblems, each more manageable and solvable with off-the-shelf classical planners. It uses a domain-independent heuristic to prioritize which automaton traces are likely to be feasible. If a trace fails to be realized, TIDE backtracks and updates its cost model based on the observed failure, thereby refining its search toward more promising alternatives.

This modular, depth-first strategy significantly reduces computational burden compared to full product graph exploration, making TIDE a practical solution for planning with LTL$_f$ goals in complex real-world domains.

To evaluate its performance, TIDE was compared against state-of-the-art approaches for LTL$_f$ goals—Exp \cite{baier_planning_2006}, Poly \cite{torres_polynomial-time_2015} and FOND4LTL$_f$ \cite{fuggitti_fond_2020}—as well as the recent Plan4Past framework \cite{bonassi_planning_2023}, designed for PPLTL goals. Experimental results demonstrate that TIDE is a valuable addition to the portfolio of planning methods for temporally extended goals, excelling in scenarios where decomposing a large problem into smaller subproblems is advantageous. The full implementation of TIDE is publicly available on GitHub at \url{https://github.com/YuliiaSuprun/TIDE} and \url{https://github.com/kavrakilab}.

\subsection{Overview}

The remainder of this article is organized as follows:
\begin{itemize}
    \item \textbf{Section~\ref{ch:RelatedWork}} reviews prior approaches to planning with temporally extended goals. It categorizes existing work into progression-based, compilation-based, and goal-decomposition strategies, highlighting both their contributions and their limitations.

    \item \textbf{Section~\ref{ch:ProblemDefinition}} formalizes the planning problem addressed in this thesis. It begins with classical planning concepts and gradually introduces the extensions needed to accommodate LTL$_f$ goals. This section also discusses classical off-the-shelf planners in the context of planning with TEGs.
    
    \item \textbf{Section~\ref{ch:Preliminaries}} introduces key automata-theoretic foundations used throughout the thesis. It covers the types of finite automata, the process of translating LTL$_f$ formulas into automata, and practical considerations such as the choice of automaton type and the representation of transition conditions.

    \item \textbf{Section~\ref{ch:ProposedMethod}} presents the TIDE planning framework in detail. The section is structured around the three main phases of the algorithm: (1) selecting a promising automaton trace using a heuristic-guided search, (2) realizing this trace through subproblem decomposition and domain planning, and (3) backtracking and refining costs when a trace fails. 

    \item \textbf{Section~\ref{ch:ExperimentResults}} evaluates the effectiveness of TIDE across several benchmarks. It includes experiments on the TB15 suite of “easy” LTL$_f$ problems, scaling benchmarks in the Blocksworld domain, and custom-designed Openstacks problems that stress-test TIDE’s backtracking capabilities. Comparisons with state-of-the-art methods—Exp, Poly, FOND4LTL$_f$ and Plan4Past— demonstrate the advantages of TIDE’s trace-guided strategy.

    \item \textbf{Section~\ref{ch:Conclusion}} concludes the article by summarizing the key findings and outlining potential directions for future research.
\end{itemize}

%% file: related_work.tex
\section{Related Work}
\label{ch:RelatedWork}
There are two primary directions in planning with LTL/LTL$_f$ goals: goal progression-based approaches and compilation-based approaches. Our approach seeks to combine the strengths of both methods by generating an accepting DFA trace that acts as goal-progression and enables us to divide the problem with temporally extended goals into a sequence of classical planning subproblems with reach-avoid goals, which are then solved using off-the-shelf planners.

\subsection{Goal Progression-Based Approaches}
\label{sec:GoalProgressionBasedApproaches}

Goal progression-based approaches tackle planning with temporal goals by reasoning about how the satisfaction of the goal formula evolves over time. Instead of constructing and searching the full product graph upfront, these methods incrementally evaluate the goal as the plan unfolds, allowing for more dynamic and flexible planning.

Early planning systems, such as TLPlan \cite{bacchus_planning_1998} and TALPlanner \cite{kvarnstrom_talplanner_2000}, leveraged domain-specific knowledge to guide the planning process. This knowledge, known as \emph{Temporal Domain Control Knowledge} (TDCK), consists of temporal logic formulas that encode desirable properties of execution traces. By evaluating these formulas during planning, these systems could effectively prune large portions of the search space, leading to significant performance improvements.

TLPlan, for instance, employs a progression algorithm that updates TDCK formulas as actions are applied, determining whether the ongoing execution still satisfies the specified constraints. If a plan prefix violates the TDCK, it is discarded, ensuring that only promising paths are explored. This approach is particularly effective for \emph{safety} conditions—properties that must always hold throughout execution. For example, a TDCK formula like $G(\text{open(door)})$ asserts that the door must remain open at all times. If an action leads to a state where the door is closed, the progression algorithm detects the violation, and the corresponding plan prefix is pruned.

However, TDCK-based progression has limitations when dealing with \emph{liveness} conditions—properties that must eventually hold. For instance, a goal like $F(\text{at(Robot, Home)})$ requires the robot to eventually reach home. The progression algorithm, focusing on the current state and immediate successors, lacks the foresight to guarantee the eventual satisfaction of such conditions, making it less effective for liveness goals.

More recent work explored alternative ways of improving efficiency without relying on manually specified control knowledge. Luo et al. (2019) proposed a sampling-based planning method inspired by RRT* (Rapidly-exploring Random Trees) \cite{luo_transfer_2019}. Their approach grows multiple random trees from predefined starting points by independently sampling both domain and automaton states. These sampled trees are then connected to form a forest of potential executions that satisfy the temporal goal. Although designed for goals expressed in LTL rather than LTL$_f$, this method demonstrates how sampling techniques can complement goal progression to enhance scalability in complex domains.

\subsection{Compilation-Based Approaches}
\label{sec:CompilationBasedApproaches}
Compilation approaches convert the LTL/LTL$_f$ planning problem into an equivalent planning problem with final-state goals, which can be efficiently solved by highly-optimized off-the-shelf planners. These methods typically involve translating the LTL$_f$ formula into an automaton and generating a product graph with the planning domain. Later, they treat a search problem on this product graph as a problem with reachability goals, where off-the-shelf planners can be applied. 

Baier and McIlraith (2006) pioneered the \textbf{Exp} method \cite{baier_planning_2006}, which constructs a Non-deterministic Finite Automaton (NFA) from the LTL$_f$ formula and generates the Cartesian product with the planning domain. 

Building on this, Torres and Baier (2015) developed the \textbf{Poly} method \cite{torres_polynomial-time_2015}, which achieves polynomial-time translations of LTL$_f$ into final-state goals using alternating automata. However, alternating automata do not cover all possible execution paths, necessitating the insertion of additional synchronization actions that add complexity and extend the plan length.

Recent work  \cite{bonassi_planning_2023} has explored planning with Pure-Past Linear Temporal Logic (PPLTL) goals instead of LTL$_f$ goals. Unlike LTL$_f$, which evaluates execution traces forward in time, PPLTL focuses on evaluating traces backward. The proposed \textbf{Plan4Past} framework addresses planning with PPLTL goals by incrementally and implicitly constructing a Deterministic Finite Automaton (DFA) during the planning process. However, this approach lacks a sense of ``direction'' for heuristic-based planners, as it converts the temporal problem into a large reachability problem, which can be harder to solve for certain planners. Moreover, while PPLTL and LTL$_f$ have the same expressive power, translating a formula from one to the other can be prohibitive, as the best-known algorithms for this conversion are 3EXPTIME \cite{giacomo_pure-past_2020}. 
Therefore, certain problems that are straightforward to express in LTL$_f$ may become impractical to represent in PPLTL, limiting applicability of the latter in certain scenarios.

\subsection{Dynamic Generation of Successive Reachability Goals}
\label{sec:DynamicGenerationofSuccessiveReachabilityGoals}
Combining goal progression with off-the-shelf planners for solving subproblems with reachability goals allows leveraging the benefits of both types of approaches. This strategy enables more directed searches while utilizing highly optimized classical planners for solving individual subproblems.

A related approach is explored in the work by Kabanza and Thiebaux (2005) \cite{kabanza_search_2005}. Their method dynamically generates successive reachability goals during planning by analyzing the temporal goal. These reachability goals are passed onto the traditional search control process, with backtracking occurring when an alternative goal appears more promising or when the current strategy does not result in a plan. However, there is no guarantee that these reachability goals will lead to an accepting state. Notably, this method focused on LTL goals, which involve complex Büchi automata, making it challenging to generate the entire sequence of reachability goals leading to an accepting state.

In contrast, our method focuses on LTL$_f$ goals, converting them into deterministic finite automata (DFA). We first generate a promising DFA trace using a heuristic, which provides an exact sequence of DFA states needed to satisfy the LTL$_f$ goal. This allows us to dynamically generate a sequence of subproblems with reach-avoid goals with the assurance that they lead to an accepting state. Then, we use an off-the-shelf planner to solve each of the subproblems. This approach ensures a more directed and potentially more efficient search process by narrowing down the search space to the most promising paths leading to an accepting state in the DFA.

%% file: problem_definition.tex
\section{Problem Definition}
\label{ch:ProblemDefinition}

\subsection{Classical Planning with Final-State Goals}
\label{sec:ClassicalPlanningwithReachabilityGoals}

This section formally defines a task planning domain and two types of classical planning problems: those with simple reachability goals and those with reachability goals and additional constraints on intermediate states. We refer to the latter as \textit{planning with reach-avoid goals}. Additionally, we describe STRIPS-like representations and introduce an interpreted task planning domain, which helps support logic-based goals.

\subsubsection{Illustrative Example: Classical Planning in Blocksworld}
\begin{figure}[H]
    \centering
    \includegraphics[width=0.85\textwidth]{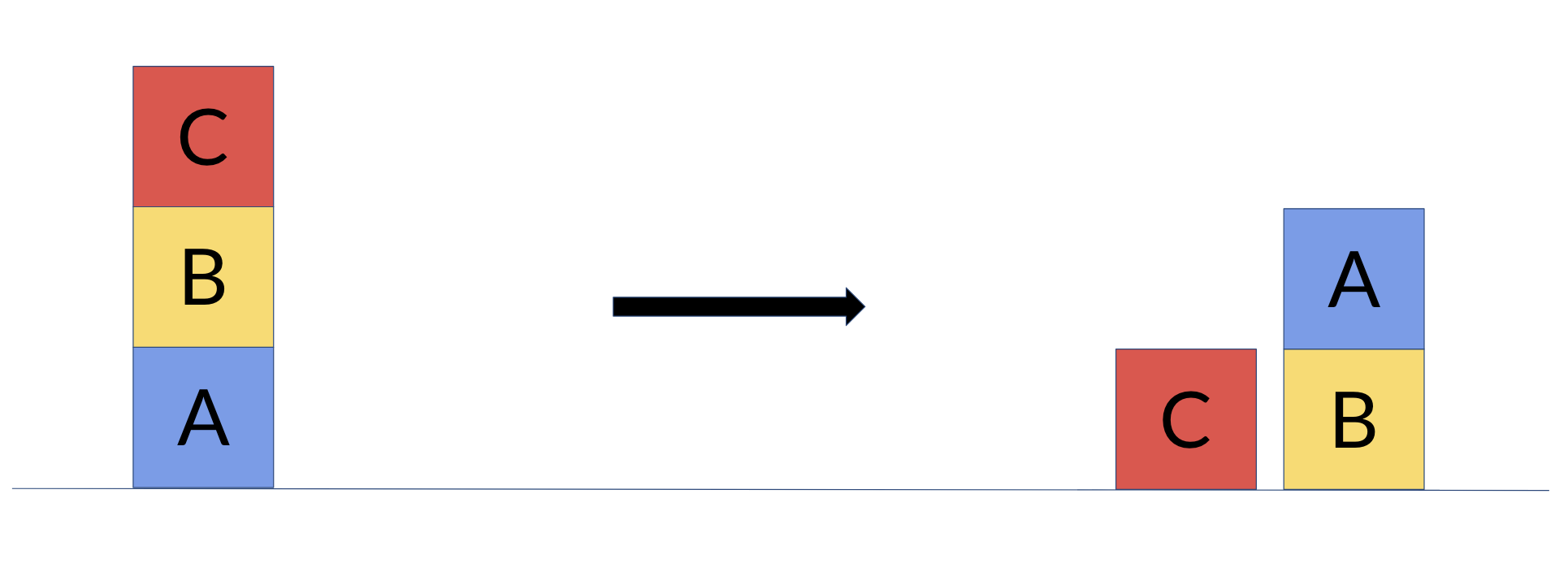}
    \caption{Example of a classical planning problem with final-state goals in the Blocksworld domain.}
    \label{fig:blocksworld-init-goal}
\end{figure}

To ground the formal definitions that follow, we begin with a simple and intuitive example in the well-known \textit{Blocksworld} domain. Consider a scenario where three blocks—A, B, and C—are stacked on top of one another in a single column, as shown in Figure~\ref{fig:blocksworld-init-goal}. The objective is to rearrange the blocks so that block C is placed on the table, and block A is placed on top of block B.

This type of planning problem focuses solely on achieving a specified configuration in the final state. The intermediate states and steps taken during execution are not part of the goal specification. What matters is that the final state satisfies the desired conditions.

To enable automated reasoning, we reformulate this problem using the \textit{Planning Domain Definition Language} (PDDL)—a standardized language used to describe planning problems. The initial and goal states can be expressed as follows:

\begin{itemize}
    \item \textbf{Initial state:} \texttt{(on B A)}, \texttt{(on C B)}, \texttt{(ontable A)}, \texttt{(clear C)}, \texttt{(handempty)}
    \item \textbf{Goal:} (\texttt{(on A B)} $\land$ \texttt{(ontable C)})
\end{itemize}

PDDL also describes the domain—the rules of the environment—using a set of predicates and actions. The Blocksworld domain includes predicates such as \texttt{(on ?x ?y)}, \texttt{(ontable ?x)}, \texttt{(clear ?x)}, and \texttt{(holding ?x)}, as well as actions like \texttt{(pick-up ?x)}, \texttt{(put-down ?x)}, \texttt{(stack ?x ?y)}, and \texttt{(unstack ?x ?y)}. These actions come with preconditions and effects that define how they interact with the current state.

Given this formalization, an off-the-shelf classical planner can automatically generate a valid sequence of actions that transitions the system from the initial state to the desired goal state. For this particular example, a possible plan is:

\texttt{(unstack C B)}, \texttt{(put-down C)}, \texttt{(unstack B A)}, \texttt{(put-down B)}, \texttt{(stack A B)}.

This demonstrates how classical planning problems can be systematically encoded and solved using domain-independent off-the-shelf planners.

\subsubsection{Task Planning Domain}

A task planning domain encapsulates the environment in which planning takes place.
The domain defines states, actions, and transitions. Formally:

\begin{definition}[Task Planning Domain]
A \textit{task planning domain} $D \doteq (\mathcal{S}, \mathcal{A}, \mathcal{T})$ consists of:
\begin{itemize}
    \item $\mathcal{S}$, a state space that can be continuous or discrete,
    \item $\mathcal{A}$, an action space,
    \item $\mathcal{T} \subseteq \mathcal{S} \times \mathcal{A} \times \mathcal{S}$, a set of deterministic transitions such that for any $(\mathbf{s}, a, \mathbf{s}') \in \mathcal{T}$, there does not exist another $\mathbf{s}'' \in \mathcal{S}$ where $(\mathbf{s}, a, \mathbf{s}'') \in \mathcal{T}$.
\end{itemize}
\end{definition}

An \textit{execution} in the domain is a sequence of alternating states and actions that describe how the system evolves over time:

\begin{equation}
E \doteq (\mathbf{s}_0, a_1, \mathbf{s}_1, a_2, \ldots, a_n, \mathbf{s}_n).
\end{equation}

An execution $E$ is considered \textit{feasible} if, for every $i \in \{1, \ldots, n\}$, the transition $(\mathbf{s}_{i-1}, a_i, \mathbf{s}_i) \in \mathcal{T}$. The action sequence $(a_1, \ldots, a_n) \in \mathcal{A}^n$ inducing $E$ is called a \textit{plan}.
The \textit{trace} of states from $E$ is denoted by $\mathfrak{S}(E) \doteq (\mathbf{s}_0, \mathbf{s}_1, \ldots \mathbf{s}_n)$.

\subsubsection{Classical Planning with Reachability Goals}

In classical planning with reachability goals, the objective is to find a plan that transitions the system from an initial state to a desired goal state.

\begin{definition}[Classical Planning Problem with Reachability Goals]
A \textit{classical planning problem with reachability goals} $P \doteq (D, \mathbf{s}_{\text{start}}, \mathbf{s}_{\text{goal}})$ consists of:
\begin{itemize}
    \item $D \doteq (\mathcal{S}, \mathcal{A}, \mathcal{T})$, a task planning domain,
    \item $\mathbf{s}_{\text{start}} \in \mathcal{S}$, the initial state of the system,
    \item $\mathbf{s}_{\text{goal}} \in \mathcal{S}$, the desired goal state.
\end{itemize}
The objective is to find a plan $(a_1, \ldots, a_n) \in \mathcal{A}^n$ such that applying the plan from $\mathbf{s}_{\text{start}}$ induces a feasible execution $E$ that terminates at $\mathbf{s}_{\text{goal}}$.
\end{definition}

This problem formulation only ensures that the system reaches the goal state $\mathbf{s}_{\text{goal}}$ without imposing additional requirements on the intermediate states.
Notably, the solution is concerned solely with feasibility, without requiring any optimization over the plan's length.

\subsubsection{Classical Planning with Reach-Avoid Goals}

Planning problems with reach-avoid goals extend simple reachability problems by introducing constraints that must hold in all intermediate states of the execution. These constraints can represent safety requirements, avoidance of obstacle states, or other
invariants.

\begin{definition}[Planning Problem with Reach-Avoid Goals]
A \textit{planning problem with reach-avoid goals} $P \doteq (D, \mathbf{s}_{\text{start}}, \mathbf{s}_{\text{goal}}, \mathcal{C})$ consists of:
\begin{itemize}
    \item $D \doteq (\mathcal{S}, \mathcal{A}, \mathcal{T})$, a task planning domain,
    \item $\mathbf{s}_{\text{start}} \in \mathcal{S}$, the initial state of the system,
    \item $\mathbf{s}_{\text{goal}} \in \mathcal{S}$, the desired goal state,
    \item $\mathcal{C} \subseteq \mathcal{S}$, a set of \textit{constraint states}, representing the subset of allowed states.
\end{itemize}
The objective is to find a plan $(a_1, \ldots, a_n) \in \mathcal{A}^n$ that induces a feasible execution $E \doteq (\mathbf{s}_0, a_1, \mathbf{s}_1, \ldots, a_n, \mathbf{s}_n)$, where:
\begin{itemize}
    \item $\mathbf{s}_0 = \mathbf{s}_{\text{start}}$,
    \item $\mathbf{s}_n = \mathbf{s}_{\text{goal}}$,
    \item $\mathbf{s}_i \in \mathcal{C}$ for all $i \in \{0, \ldots, n-1\}$.
\end{itemize}
\end{definition}

\subsubsection{Planning with Logic-Based Goals}
\subsubsection{STRIPS-like Representations}
Task planning often involves reasoning about high-level properties of states. To support this reasoning, states in the domain are augmented with logic-based annotations. A common representation in planning is the STRIPS-like representation, whose name is derived from Stanford Research Institute Problem Solver \cite{fikes_strips_1971}. STRIPS-like representations, used extensively in the PDDL (Planning Domain Definition Language) format, provide a structured way to encode planning problems with propositional or temporal logic.

In STRIPS-like representations:
\begin{itemize}
    \item States are defined by a set of literals (e.g., \texttt{On(BlockA, BlockB)}), with
any unspecified literals assumed to be false.
    \item Operators (or actions) are defined by:
        \begin{itemize}
            \item Preconditions: conditions that must hold for the action to be applied,
            \item Effects: conditions that describe how the action modifies the state.
        \end{itemize}
    \item Goals are specified as a set of positive and/or negative literals that must hold in the final state.
\end{itemize}

\subsubsection{Interpreted Task Planning Domain}

To extend STRIPS-like representations, interpreted task planning domains incorporate a set of atomic propositions that describe high-level properties of states:

\begin{definition}[Interpreted Task Planning Domain]
An \textit{interpreted task planning domain} $\mathcal{D} \doteq (\mathcal{S}, \mathcal{A}, \mathcal{T}, \Prop, \ell)$ consists of:
\begin{itemize}
    \item $\mathcal{S}$, $\mathcal{A}$, $\mathcal{T}$ as defined in the task planning domain,
    \item $\Prop$, a finite set of atomic propositions describing high-level properties of states,
    \item $\ell: \mathcal{S} \to 2^{\Prop}$, a labeling function, which maps each state to a subset of atomic propositions that hold true in that state, effectively describing the ``world state''.
\end{itemize}
\end{definition}

\subsubsection{Planning Problem with Propositional Reach-Avoid Goals}
Our proposed method operates by decomposing a problem with LTL$_f$ goals into a sequence of classical problems with propositional reach-avoid goals. Having established the necessary foundations, we are now ready to provide an exact definition of such problems.

\begin{definition}[Planning Problem with Propositional Reach-Avoid Goals]
A planning problem with propositional reach-avoid goals is defined as $P \doteq (\mathcal{D}, \mathbf{s}_{\text{start}}, \varphi_{\text{goal}}, \varphi_{\text{constraint}})$, where:
\begin{itemize}
    \item $\mathcal{D} \doteq (\mathcal{S}, \mathcal{A}, \mathcal{T}, \Prop, \ell)$, an interpreted task planning domain,
    \item $\mathbf{s}_{\text{start}} \in \mathcal{S}$, the initial state of the system,
    \item $\varphi_{\text{goal}}$, a propositional logic formula over $\Prop$ that must be satisfied in the final state of the execution,
    \item $\varphi_{\text{constraint}}$, a propositional logic formula over $\Prop$ that must be satisfied in all intermediate states of the execution.
\end{itemize}
The objective is to find a plan $(a_1, \ldots, a_n) \in \mathcal{A}^n$ that induces a feasible execution $E \doteq (\mathbf{s}_0, a_1, \mathbf{s}_1, \ldots, a_n, \mathbf{s}_n)$, where:
\begin{itemize}
    \item $\ell(\mathbf{s}_n) \vDash \varphi_{\text{goal}}$,
    \item $\ell(\mathbf{s}_i) \vDash \varphi_{\text{constraint}}$ for all $i \in \{0, \ldots, n-1\}$.
\end{itemize}
\end{definition}

\subsection{Off-the-shelf Planners for Final-State Goals}
In this subsection, we discuss how existing classical off-the-shelf planners, such as A* and Fast Downward, can be applied to problems with reachability goals. Additionally, we explain how to apply them to reach-avoid problems by encoding constraints within the PDDL domain.

\subsubsection{Types of Planners Considered}
In our experiments, we have explored the following classical planners for reachability goals.

\paragraph{A* Search:}  
A* is a heuristic-based search algorithm that explores paths in a state space while prioritizing those with the lowest estimated cost. At each step, A* evaluates states using the cost function:
\[
f(n) = g(n) + h(n),
\]
where \( g(n) \) is the cost of reaching state \( n \) from the initial state, and \( h(n) \) is the heuristic estimate of the cost to reach the goal from \( n \) \cite{hart_formal_1968}. A* guarantees optimality if the heuristic \( h(n) \) is admissible (i.e., it never overestimates the true cost). 

\paragraph{Fast Downward with LAMA Search:}  
Fast Downward is a state-of-the-art classical planner for final-state goals designed to solve planning problems encoded in the Planning Domain Definition Language (PDDL) \cite{helmert_fast_2006}. It incorporates multiple search strategies, including LAMA search, which we used for our experiments. LAMA accelerates the search process by identifying landmarks—subgoals that must be satisfied in any valid plan \cite{richter_lama_2010}. LAMA employs a greedy best-first search strategy with landmarks and a heuristic that measures progress toward the goal. While it often generates high-quality plans, it doesn't provide any optimality guarantees.

\subsubsection{How to Apply Planners to Reach-Avoid Problems}
While off-the-shelf planners like A* and Fast Downward are designed for reachability problems, they can be adapted to handle reach-avoid problems by encoding constraints into the PDDL domain. This ensures that all intermediate states satisfy the required constraints, effectively transforming the problem into one that is compatible with standard planners.

\paragraph{Approaches for Encoding Constraints}
To incorporate constraints in reach-avoid problems, we can use one of the following methods:

\begin{itemize}
    \item \textbf{Extend Problem Encoding:}  
    Constraints can be expressed as additional derived predicates in the PDDL domain file and explicitly included in the problem definition using the \texttt{constraints} feature of PDDL 3.0. This approach allows direct specification of constraints as part of the problem. However, not all planners support PDDL 3.0 features (Fast Downward is one of them). 

    \item \textbf{Add Constraints to Action Preconditions:}  
    A more universally supported approach involves modifying the PDDL domain to include constraints in the preconditions of all actions. For example, if a constraint requires that a predicate \( p \) must hold true in all intermediate states, \( p \) is added as a precondition for every action. This ensures that only actions leading to valid intermediate states are considered by the planner.

    To further enhance efficiency in domains with large action spaces, redundant evaluations of constraints across multiple actions can be avoided by using the following strategy:

    \begin{itemize}
    \item Introduce a new predicate, \texttt{constraint\_satisfied}, in the PDDL domain.
    \item Modify all actions to include \texttt{(constraint\_satisfied)} as a precondition and \texttt{(not constraint\_satisfied)} as an effect. 
    \item Add a dedicated ``constraint-checking'' action to the domain. This action sets \texttt{(constraint\_satisfied)} to true if the current state satisfies the constraint, effectively validating the state before proceeding.
\end{itemize}
    This approach guarantees that all intermediate states satisfy the specified constraints.
\end{itemize}

Since Fast Downward does not support PDDL 3.0 features, our method employs the second approach by modifying action preconditions to enforce constraints indirectly.

\subsection{Planning with Temporally Extended Goals}
\label{sec:PlanningwithTemporallyExtendedGoals}

\subsubsection{Illustrative Example: Temporally Extended Goals in Blocksworld}
\label{sec:ExampleTEG}

So far, we have considered problems with final-state goals, where the objective is to reach a specific configuration regardless of the intermediate steps taken. However, many real-world tasks impose additional requirements along the way, where certain conditions must hold at intermediate stages of execution. \textit{Temporally extended goals} (TEGs) allow for richer and more expressive goal specifications.

To illustrate this, consider a task in the well-known Blocksworld domain involving three blocks—A, B, and C. The blocks initially form a single vertical stack, with C on top of B and B on top of A. The goal is not merely to reach a final configuration, but to achieve two temporally ordered subgoals:
\begin{itemize}
    \item First, reverse the order of the blocks so that block A is on block B, and block B is on block C.
    \item Then, place all three blocks on the table.
\end{itemize}

This problem introduces a form of sequencing: one subgoal must be satisfied before another, making it impossible to evaluate plan success by examining only the final state.

\begin{figure}[h]
    \centering
    \includegraphics[width=0.9\textwidth]{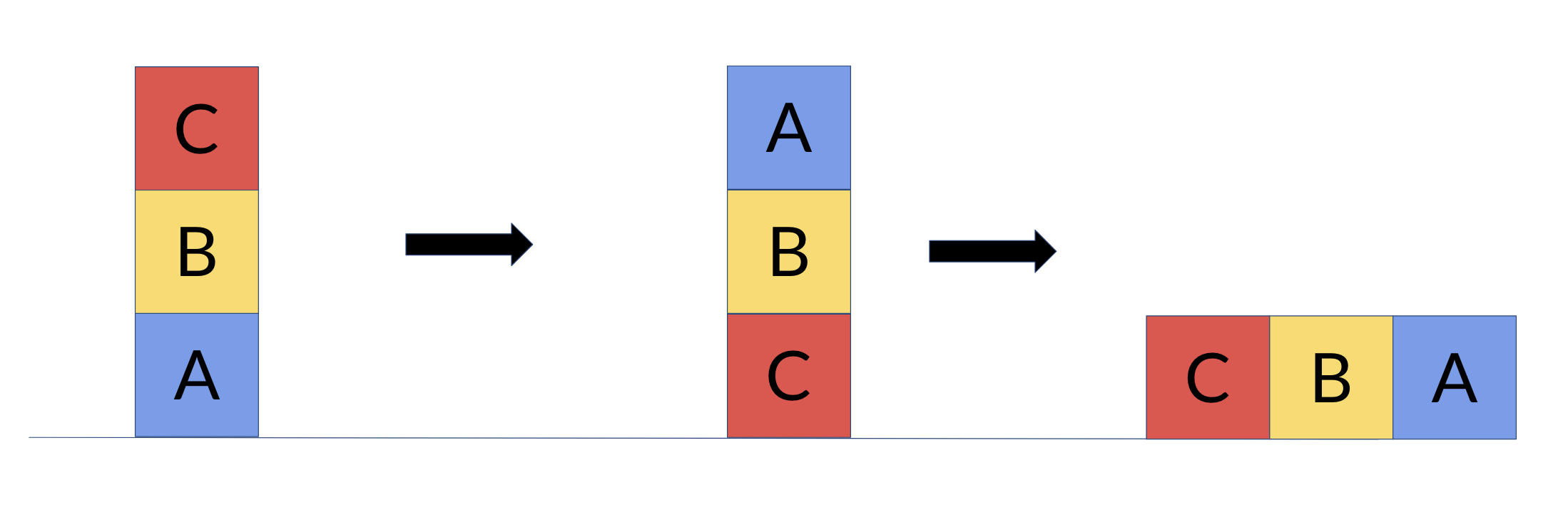}
    \caption{Problem with a Temporally Extended Goal: reverse the order of the blocks and then place them all on the table.}
    \label{fig:teg_example}
\end{figure}

Let us now consider how to formally define this problem using LTL$_f$. 

The initial states and a goal expressed in LTL$_f$ can be formulated as follows:

\begin{itemize}
    \item \textbf{Initial state:} \texttt{(on B A)}, \texttt{(on C B)}, \texttt{(ontable A)}, \texttt{(clear C)}, \texttt{(handempty)}
    \item \textbf{Goal:} \(
F \left(((\texttt{on A B}) \land (\texttt{on B C})) \land (X \, (F ((\texttt{ontable A}) \land (\texttt{ontable B}) \land (\texttt{ontable C}))) \right))
\)
\end{itemize}

This goal formula specifies that:
\begin{itemize}
    \item Eventually, the blocks must be stacked in reverse order (A on B and B on C);
    \item Then, in the next step, it must eventually become true that all three blocks are placed on the table.
\end{itemize}

This example highlights the expressive power of LTL$_f$ in capturing rich temporal behaviors that cannot be represented by traditional final-state goals. It also demonstrates the need for planning techniques capable of reasoning over entire execution traces, rather than evaluating only the final outcome.

\subsubsection{Linear Temporal Logic over Finite Traces}
Linear Temporal Logic over finite traces (LTL$_f$) extends propositional logic by adding temporal operators \cite{baier_planning_2006, de_giacomo_linear_2013}. Unlike traditional LTL \cite{pnueli_temporal_1977}, which is defined on infinite traces, LTL$_f$ focuses on traces of finite length.

\begin{definition}[Syntax of LTL$_f$]
Let $\Prop$ denote a finite set of atomic propositions. The syntax of an LTL$_f$ formula $\varphi$ is identical to that of LTL and is defined as follows:

\[
\varphi ::= \top \, | \, p \, | \, (\neg \varphi) \, | \, (\varphi_1 \land \varphi_2) \, | \, (X \varphi) \, | \, (\varphi_1 U \varphi_2) 
\]

where $p \in \Prop$ and $\top$ represents a tautology. The symbols $X$ (Next) and $U$ (Until) denote temporal operators \cite{de_giacomo_linear_2013}.

\end{definition}
The additional temporal operators ``Eventually'' ($F$) and ``Globally'' ($G$) are expressed as follows:

\begin{itemize}
    \item $F\varphi = \top U \varphi$
    \item $G\varphi = \neg F \neg \varphi$
\end{itemize}

\begin{definition}[Semantics of LTL$_f$]
The semantics of LTL$_f$ is established over finite traces. A trace $\rho$ represents a word in $(2^{\Prop})^*$. Let $|\rho|$ be the length of trace $\rho$ and $\rho[i]$ denote the $i^{th}$ symbol of $\rho$. Additionally, $\rho, i \vDash \varphi$ is interpreted as ``the $i^{th}$ step of trace $\rho$ satisfies $\varphi$.'' The semantics are defined as:

\begin{itemize}
    \item $\rho, i \vDash \top$;
    \item $\rho, i \vDash p \iff p \in \rho[i]$;
    \item $\rho, i \vDash (\neg \varphi) \iff \rho, i \nvDash \varphi$;
    \item $\rho, i \vDash (\varphi_1 \land \varphi_2) \iff \rho, i \vDash \varphi_1\; and\; \rho, i \vDash \varphi_2$;
    \item $\rho, i \vDash (X \varphi) \iff |\rho| > i + 1 \; and \; \rho, i+1 \vDash \varphi$;
    \item $\rho, i \vDash (\varphi_1 U \varphi_2) \iff \exists j \; s.t. \; i \le j < |\rho|\; and \;\\ \rho, j \vDash \varphi_2 \; and\; \forall k, \; i \le k< j, \; \rho, k \vDash \varphi_1$;
\end{itemize}

\end{definition}
A finite trace $\rho$ is said to satisfy $\varphi$, denoted by $\rho \vDash \varphi$, if and only if $\rho, 0 \vDash \varphi$.

An LTL$_f$ formula $\varphi$ generates a language $\mathcal{L}(\varphi)$ over the alphabet $2^{\Prop}$, defined as follows:

\[
\mathcal{L}(\varphi) = \{\rho \in (2^{\Prop})^* \, |\, \rho \vDash \varphi\}
\]

\subsubsection{Planning with LTL$_f$ Goals}

Linear Temporal Logic over finite traces (LTL$_f$) provides a powerful framework for specifying goals that involve temporal constraints over sequences of states. Unlike propositional reach-avoid goals, which focus on specific states, LTL$_f$ allows for reasoning about entire execution traces. 

\begin{definition}[Planning Problem with LTL$_f$ Goals]
A planning problem with LTL$_f$ goals is a tuple $(\mathcal{D}, \mathbf{s}_{start}, \varphi_G)$, where: \begin{itemize}
    \item $\mathcal{D} \doteq (\mathcal{S}, \mathcal{A}, \mathcal{T}, \Prop, \ell)$ is an interpreted task planning domain,
    \item $\mathbf{s}_{start}$ is the initial state of the system,
    \item $\varphi_G$ is the LTL$_f$ formula defined over a finite set of atomic propositions $\Prop$, which describes the goal of an agent.
\end{itemize}
The objective is to find a plan $(a_1, \ldots, a_n) \in \mathcal{A}^n$ that induces a feasible execution $E \doteq (\mathbf{s}_0, a_1, \mathbf{s}_1, \ldots, a_n, \mathbf{s}_n)$, where: \begin{itemize} \item $\ell(\mathfrak{S}(E)) \vDash \varphi_G$. \end{itemize} Here, $\mathfrak{S}(E)$ represents the trace of domain states $(\mathbf{s}_0, \mathbf{s}_1, \ldots, \mathbf{s}_n)$ in the execution, and $\ell(\mathfrak{S}(E))$ is the corresponding sequence of world states determined by the labeling function $\ell$.
\end{definition}

Note that in this context, the focus is again on finding a feasible plan rather than optimizing plan length or cost.

LTL$_f$ goals enable the specification of complex task requirements, such as: \begin{itemize} \item \textbf{Sequencing:} Certain tasks must occur in a specified order \\ Example: $F (\texttt{On(BlockA, BlockB)} \land X (\texttt{On(BlockB, BlockC)}))$. \item \textbf{Safety:} Specific conditions must hold throughout the execution \\ Example: $G (\neg \texttt{Danger})$. \item \textbf{Liveness:} Certain states must eventually be reached \\ Example: $F (\texttt{Complete})$. \end{itemize}

To address the complexity of LTL$_f$ goals, our proposed method decomposes the problem with LTL$_f$ goals into a sequence of planning problems with propositional reach-avoid goals. This decomposition enables the efficient application of existing off-the-shelf planners for problems with final-state goals.

%% file: preliminaries.tex
\section{Preliminaries}
\label{ch:Preliminaries}

Planning with temporally-extended goals often involves translating temporal logic formulas into automata. This translation is helpful because it converts temporal logic specification into a state-based model that can be
efficiently searched. This chapter defines the key types of finite automata used in this context—Deterministic Finite Automata (DFA), Non-deterministic Finite Automata (NFA), and Alternating Finite Automata (AFA)—and discusses their relevance to translating LTL$_f$ formulas.

\subsection{Finite Automata Definitions}

\subsubsection{Deterministic Finite Automaton (DFA)}
A DFA is a state-based model with deterministic transitions, where each state and input symbol pair maps to a unique successor state.

\begin{definition}[Deterministic Finite Automaton]
A DFA is defined as a tuple \( D = (Q, q_0, \Sigma, \delta, F) \), where:
\begin{itemize}
    \item \( Q \): a finite set of states,
    \item \( q_0 \in Q \): the initial state,
    \item \( \Sigma = 2^\Prop \): the alphabet, where \( \Prop \) is the set of atomic propositions,
    \item \( \delta: Q \times \Sigma \to Q \): the transition function, mapping a state and symbol to a single state,
    \item \( F \subseteq Q \): the set of accepting states.
\end{itemize}
\end{definition}

A finite word \( w = w_0, w_1, \ldots, w_n \in \Sigma^* \) is said to have a run \( \rho = q_0, q_1, \ldots, q_{n+1} \in Q^+ \) in \( D = (Q, q_0, \Sigma, \delta, F)  \) if \( q_0 \) is the initial state and
\( q_{i+1} = \delta(q_i, w_i) \) for all \( i \in \{0, \ldots, n\} \).

A DFA accepts a word \( w \) if there exists a \textbf{unique} run \( \rho = q_0, q_1, \ldots, q_{n+1} \) such that the final state belongs to the set of accepting states: \( q_{n+1} \in F \).

The language of \( D \), denoted \( \mathcal{L}(D) \), is the set of all words \( w \) that have an accepting run in \( D \).

A DFA is said to be \textit{minimal} if the language represented by this DFA cannot be represented by another DFA with fewer states.

\subsubsection{Non-deterministic Finite Automaton (NFA)}
An NFA is a state-based model where transitions are non-deterministic, meaning that from a given state, multiple transitions may be possible for the same input.

\begin{definition}[Non-deterministic Finite Automaton]
An NFA is defined as a tuple \( N = (Q, Q_0, \Sigma, \delta, F) \), where:
\begin{itemize}
    \item \( Q \): a finite set of states,
    \item \( Q_0 \subseteq Q \): a set of initial states,
    \item \( \Sigma = 2^\Prop \): the alphabet, where \( \Prop \) is the set of atomic propositions,
    \item \( \delta: Q \times \Sigma \to 2^Q \): the transition function, mapping a state and symbol to a set of states,
    \item \( F \subseteq Q \): the set of accepting states.
\end{itemize}
\end{definition}

A finite word \( w = w_0, w_1, \ldots, w_n \in \Sigma^* \) is said to have a run \( \rho = q_0, q_1, \ldots, q_{n+1} \in Q^+ \) in \( N = (Q, Q_0, \Sigma, \delta, F)\) if \( q_0 \in Q_0 \) and \( q_{i+1} \in \delta(q_i, w_i) \) for all \( i \in \{0, \ldots, n\} \).

An NFA accepts a word \( w = w_0, w_1, \ldots, w_n \in \Sigma^* \) if there exists \textbf{at least one} run \( \rho = q_0, q_1, \ldots, q_{n+1} \) such that the final state belongs to the set of accepting states: \( q_{n+1} \in F \).

The language of \( N \), denoted \( \mathcal{L}(N) \), is the set of all words \( w \) that have at least one accepting run in \( N \).

\subsubsection{Alternating Finite Automaton (AFA)}

An AFA generalizes NFAs by introducing alternation, which enables a single state to represent multiple possible transitions simultaneously using Boolean formulas. This makes AFAs particularly compact and efficient to construct from logical formulas.

\begin{definition}[Alternating Finite Automaton]
An AFA is defined as a tuple \( A = (Q, q_0, \Sigma, \delta, F) \), where:
\begin{itemize}
    \item \( Q \): a finite, non-empty set of states,
    \item \( q_0 \in Q \): the initial state,
    \item \( \Sigma = 2^\Prop \): the alphabet, where \( \Prop \) is the set of atomic propositions,
    \item \( \delta: Q \times \Sigma \to B^+(Q) \): the transition function, where \( B^+(Q) \) denotes the set of positive Boolean formulas whose variables are states in \( Q \).
    \item \( F \subseteq Q \): the set of accepting states.
\end{itemize}
\end{definition}
A run of an AFA on a word \( w = w_0, w_1, \ldots, w_n \in \Sigma^* \) is represented as a tree rather than a sequence. Each node of the tree corresponds to a state in \( Q \), and the edges correspond to transitions defined by the Boolean formulas in \( \delta \).

A finite word \( w = w_0, w_1, \ldots, w_n \in \Sigma^* \) is said to have a run in \( A = (Q, q_0, \Sigma, \delta, F) \) if it satisfies the following conditions:
\begin{enumerate}
    \item The root of the tree is labeled with the initial state \( q_0 \).
    \item If a node at level \( i \) is labeled with state \( q \in Q \) and the current input symbol is \( a_i \), then the transition function \( \delta(q, a_i) = \Theta \) determines the set of child nodes:
        \begin{itemize}
            \item If \( \Theta \) is \texttt{true}, the node has no children.
            \item Otherwise, the children correspond to the states in \( Q \) that satisfy \( \Theta \).
        \end{itemize}
\end{enumerate}

An AFA accepts a word \( w = w_0, w_1, \ldots, w_n \in \Sigma^* \) if there exists a run such that all leaves at depth \((n + 1)\) (the length of the input word $w$) are
labeled with states in \(F\). In other words, every branch in an accepting run has to either hit the $\true$ transition or hit an accepting state after reading the entire input word $w$.

Similar to NFAs and DFAs, AFAs are equivalent in expressive power to regular expressions \cite{brzozowski_equations_1980}.

\subsection{Translating LTL$_f$ Formulas into Automata}

\subsubsection{Translation to AFA}
Any LTL$_f$ formula \( \varphi \) can be systematically translated into an equivalent AFA \( A_{\varphi} \), such that for any word \( w \), \( w \in \mathcal{L}(A_{\varphi}) \iff w \vDash \varphi \). The translation process begins by decomposing the LTL$_f$ formula into its subformulas. Each subformula corresponds to a state in the AFA, while transitions between states are derived based on the logical operators present in the formula. This process captures the semantics of the LTL$_f$ formula and runs in \textit{polynomial time} with respect to the size of the formula \cite{de_giacomo_linear_2013, torres_polynomial-time_2015}.

\subsubsection{Translation to NFA}
An AFA \( A \) can be converted into an equivalent NFA \( N \), which accepts the same language (\( \mathcal{L}(A) = \mathcal{L}(N) \)). This conversion involves an exponential blowup in the size of the automaton. Consequently, directly translating an LTL$_f$ formula into an NFA is a \textit{single-exponential} process in terms of the size of the formula.

\subsubsection{Translation to DFA}
An NFA \( N \) can then be converted into a DFA \( D \), accepting the same language (\( \mathcal{L}(N) = \mathcal{L}(D) \)). This conversion, known as \textit{determinization}, is typically achieved using the Rabin–Scott powerset construction \cite{rabin_finite_1959}. The process involves creating DFA states where each state corresponds to a subset of NFA states, resulting in a single-exponential increase in size relative to the NFA. Thus, when the entire sequence of conversions—LTL$_f$ to AFA, AFA to NFA, and NFA to DFA—is considered, the total computational complexity becomes \textit{double-exponential} in the size of the original LTL$_f$ formula.

\paragraph{Remarks on Translation Methods}
It is important to note that translating LTL$_f$ formulas into automata does not always follow the AFA-to-NFA-to-DFA pipeline. Different approaches exist, with varying intermediate stages, depending on the tools and methods used. For instance, our method utilizes the third-party library Spot \cite{artho_spot_2016, shoham_spot_2022}, which is optimized for handling LTL and $\omega$-automata. Spot internally converts an LTL$_f$ formula into an equivalent LTL formula, which is then transformed into a Büchi automaton before determinization into a DFA. 

While the specific stages of conversion may vary, the computational complexity analysis presented here provides intuition for why translating an LTL$_f$ formula into different automata types has varying complexities with respect to the formula size: 
\begin{itemize}
    \item LTL$_f$-to-AFA translation is polynomial-time (PTIME),
    \item LTL$_f$-to-NFA translation is single-exponential time (EXPTIME),
    \item LTL$_f$-to-DFA translation is double-exponential time (2EXPTIME).
\end{itemize}

\subsection{Product Automata}
\label{sec:ProductAutomata}
In the context of planning, an automaton derived from an LTL$_f$ formula can directly guide the search process. By integrating the goal automaton with an interpreted planning domain $\mathcal{D} = (\mathcal{S}, \mathcal{A}, \mathcal{T}, \Prop, \ell)$, planners can ensure that the resulting plans not only achieve the desired temporal goals specified by the LTL$_f$ formula but also adhere to the constraints of the domain. This integration often takes the form of a product automaton, where each state represents a combination of a domain state and an automaton state. Transitions in this product automaton are allowed only if they are valid under both the domain transition system and the automaton. The formal definition of the product automaton is given below. Note that the terms Product Automaton and Product Graph are used interchangeably. 
\begin{definition}[Product Automaton]
Let the interpreted planning domain be represented as \(\mathcal{D} = (\mathcal{S}, \mathcal{A}, \mathcal{T}, \Prop, \ell)\), where:
\(\mathcal{S}\) is the set of domain states, \(\mathcal{A}\) is the set of actions, \(\mathcal{T} \subseteq \mathcal{S} \times \mathcal{A} \times \mathcal{S}\) is the transition relation, \(\Prop\) is the set of atomic propositions, and \(\ell: \mathcal{S} \to 2^{\Prop}\) is the labeling function mapping each state to a set of propositions.

Let the automaton be represented as \(A = (Q, q_{\text{start}}, \Sigma, \delta, F)\), where \(Q\) is the set of automaton states, \(q_{\text{start}} \in Q\) is the initial automaton state, \(\Sigma = 2^{\Prop}\) is the alphabet, \(\delta: Q \times \Sigma \to 2^Q\) is the transition function, and \(F \subseteq Q\) is the set of accepting automaton states.

The Product Automaton \(PA\) is defined as the tuple \(PA = (S_{\text{PA}}, p_{\text{start}}, \to_{\text{PA}}, S_F)\), where:
\begin{itemize}
    \item \(S_{\text{PA}} = \mathcal{S} \times Q\): the set of product states, each representing a combination of a domain state and an automaton state.
    \item \(p_{\text{start}} = (\mathbf{s}_{\text{start}}, q_{\text{start}}) \in S_{\text{PA}}\): the initial product state, formed by the initial domain state \(\mathbf{s}_{\text{start}}\) and the initial automaton state \(q_{\text{start}}\).
    \item \(\to_{\text{PA}} \subseteq S_{\text{PA}} \times S_{\text{PA}}\): the product transition relation, defined as:
    \[
    (\mathbf{s}, q) \to_{\text{PA}} (\mathbf{s}', q') \iff \exists a \in \mathcal{A}, (\mathbf{s}, a, \mathbf{s}') \in \mathcal{T} \text{ and } q' \in \delta(q, \ell(\mathbf{s})).
    \]
    \item \(S_F = \mathcal{S} \times F\): the set of accepting product states, where the domain state is paired with an accepting automaton state.
\end{itemize}
\end{definition}

An accepted run in the Product Automaton \(PA\) is a sequence of product states starting from the initial product state \((\mathbf{s}_{\text{start}}, q_{\text{start}})\) and ending in one of the accepting product states in \(S_F\). Formally, an \textit{accepted} run \(\pi\) is a sequence \((\mathbf{s}_0, q_0), (\mathbf{s}_1, q_1), \ldots, (\mathbf{s}_n, q_n)\), where:
\begin{itemize}
    \item \((\mathbf{s}_0, q_0) = (\mathbf{s}_{\text{start}}, q_{\text{start}})\)
    \item \((\mathbf{s}_i, q_i) \to_{\text{PA}} (\mathbf{s}_{i+1}, q_{i+1})\) for all \(i \in \{0, \ldots, n-1\}\),
    \item \((\mathbf{s}_n, q_n) \in S_F\).
\end{itemize}

The significance of an accepted path in \(PA\) is that the sequence of domain states \(\mathfrak{S}(E) = (\mathbf{s}_0, \mathbf{s}_1, \ldots \mathbf{s}_n)\) along this product path satisfies the LTL$_f$ formula \(\varphi_G\) from which the automaton was constructed: \(\ell(\mathfrak{S}(E)) \vDash \varphi_G\). 

\subsection{Practical Considerations: DFA vs. NFA vs. AFA}
\label{sec:NFAvsDFAvsAFA}

When translating LTL$_f$ formulas into automata for planning, the choice of automaton type—DFA, NFA, or AFA—significantly impacts both computational complexity and practical performance. Each automaton type has its advantages and limitations, described below:
\begin{itemize}
    \item \textbf{Deterministic Finite Automaton:}
    \begin{itemize}
        \item \textbf{Pros:} Deterministic transitions ensure that every input word has at most one accepting path, simplifying the search process. Additionally, DFAs can be efficiently minimized, often resulting in a compact representation that is convenient for planning.
        \item \textbf{Cons:} Translating an LTL$_f$ formula into a DFA has double-exponential complexity with respect to the formula size.
    \end{itemize}

    \item \textbf{Non-deterministic Finite Automaton:}
    \begin{itemize}
        \item \textbf{Pros:} Translating an LTL$_f$ formula into an NFA has single-exponential complexity with respect to the formula size, making it computationally less expensive than DFA construction.
        \item \textbf{Cons:} Retains non-determinism, meaning that a single input word can have multiple potential accepting paths. This complicates planning algorithms, as they must handle multiple simultaneous transitions, leading to increased computational overhead during the planning phase.
    \end{itemize}

    \item \textbf{Alternating Finite Automaton:}
    \begin{itemize}
        \item \textbf{Pros:} Translating an LTL$_f$ formula into an AFA has polynomial complexity with respect to the formula size. AFAs also offer a compact representation of transitions through Boolean formulas.
        \item \textbf{Cons:} Alternating transitions require additional synchronization mechanisms, often introducing ``spurious'' actions that lead to longer plans and make a planning process significantly slower.
    \end{itemize}
\end{itemize}

Despite the higher theoretical complexity of DFA construction, DFAs often prove to be more practical in real-world applications \cite{tabakov_optimized_2012}. Their deterministic nature simplifies planning algorithms. Furthermore, DFAs can be minimized using well-established algorithms, such as Hopcroft's \cite{hopcroft_n_1971} or Brzozowski's \cite{brzozowski_canonical_1962} methods, resulting in a compact and more efficient representation. In contrast, while AFAs and NFAs are computationally easier to generate in theory, they can be harder to use in practice. 

Given these considerations, our method converts LTL$_f$ formulas into DFAs. To evaluate our approach, we compare it to other state-of-the-art methods for planning with LTL$_f$ goals:
\begin{itemize}
    \item \textbf{FOND4LTL$_f$} \cite{fuggitti_fond_2020}, which translates LTL$_f$ formulas into DFAs,
    \item \textbf{Exp} \cite{baier_planning_2006}, which translates LTL$_f$ formulas into NFAs,
    \item \textbf{Poly} \cite{torres_polynomial-time_2015}, which translates LTL$_f$ formulas into AFAs.
\end{itemize}
The names Exp and Poly reflect the complexity of the automata construction these methods use: exponential for NFAs and polynomial for AFAs. 
Additionally, we compare our method against the \textbf{Plan4Past} \cite{bonassi_planning_2023}, which expresses goals in PPLTL (Pure-Past Linear Temporal Logic) and implicitly constructs a DFA from PPLTL formulas. Notably, due to the property of reverse languages \cite{chandra_alternation_1981}, the corresponding DFA can be computed directly from the PPLTL formula in \textit{single-exponential} time with respect to the size of the formula.

\subsection{Representing Edge Conditions in Automata}

\subsubsection{Edge Conditions and the Alphabet of Automata}
In automata constructed from LTL$_f$ formulas, transitions between states are governed by edge conditions, also referred to as \textit{edge guards}. These conditions specify the logical predicates that must be satisfied for the transition to occur. 

The alphabet of the automaton, denoted by \( \Sigma \), is typically defined as \( \Sigma = 2^{\Prop} \), where \( \Prop \) is the set of atomic propositions. Each symbol in \( \Sigma \) represents a subset of atomic propositions, and transitions are labeled with such subsets. However, this representation has an inherent limitation: the size of the alphabet \( \Sigma \) grows exponentially with the number of atomic propositions (\( |\Sigma| = 2^{|\Prop|} \)).

An alternative way to represent edge conditions is to use Binary Decision Diagrams (BDDs). BDDs offer a compact and efficient data structure for encoding Boolean formulas, reducing the overhead associated with an exponential alphabet size.

As mentioned before, our method leverages the Spot library \cite{artho_spot_2016, shoham_spot_2022} for LTL$_f$-to-DFA translation, which employs a customized version of the BuDDy library to manipulate BDDs and uses them to label edges in automata.

\subsubsection{Binary Decision Diagrams (BDDs)}
A Binary Decision Diagram is a data structure designed for the efficient manipulation of Boolean formulas. It represents Boolean formulas as directed acyclic graphs, where:
\begin{itemize}
    \item Non-terminal nodes (called \textit{decision nodes}) are labeled with Boolean variables.
    \item Each decision node has two child nodes: \begin{itemize}
        \item a \textit{low} child for the \texttt{false} branch,
        \item a \textit{high} child for the \texttt{true} branch.
    \end{itemize}
    \item Terminal nodes (called \textit{leaf nodes}) represent the truth values \texttt{true} or \texttt{false}.
\end{itemize}

Let \( \textbf{x} \) denote a vector of Boolean variables \( x_1, x_2, \ldots, x_n \). Boolean functions over these variables are written as \( f(\textbf{x}) \) or simply \( f \) when the arguments are clear. Let \( \textbf{a} \) denote a vector of values \( a_1, a_2, \ldots, a_n \), where each \( a_i \in \{0, 1\} \). The \textit{valuation} of a function \( f \) applied to \( \textbf{a} \) is denoted \( f(\textbf{a}) \), where \( f(\textbf{a}) \in \{0, 1\} \).

Let \textbf{1} denote the function that always yields 1 (\texttt{true}), and \textbf{0} the function that always yields 0 (\texttt{false}).

\begin{definition}[Restriction]
For a function \( f \), variable \( x_i \), and binary value \( b \in \{0, 1\} \), the \textit{restriction} of \( f \) is the function resulting from setting \( x_i \) to \( b \):
\[
f |_{x_i \gets b}(\textbf{x}) = f(x_1, \ldots, x_{i-1}, b, x_{i+1}, \ldots, x_n),
\]
where \( b \in \{0, 1\} \) specifies whether the variable is set to \texttt{true} (\( b = 1 \)) or \texttt{false} (\( b = 0 \)).
\end{definition}
\begin{definition}[Cofactors]
The two restrictions of \( f \) with respect to \( x_i \) are referred to as the \textit{cofactors}:
\[
f |_{x_i \gets 1} \quad \text{and} \quad f |_{x_i \gets 0}.
\]

Using the Shannon expansion, the function \( f \) can be reconstructed as:
\[
f = \left(x_i \land f |_{x_i \gets 1}\right) \lor \left(\lnot x_i \land f |_{x_i \gets 0}\right).
\]
\end{definition}

\begin{figure}[ht]
    \centering
    \includegraphics[width=0.3\textwidth]{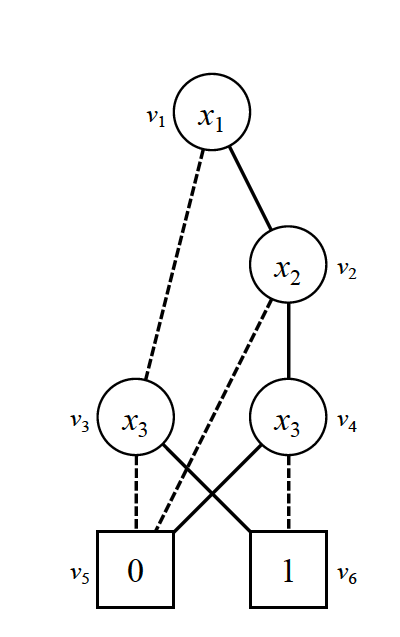}
    \caption{BDD representation of the function \( f(x_1, x_2, x_3) = (x_1 \land x_2 \land \neg x_3) \lor (\neg x_1 \land x_3) \). Dashed lines represent transitions to the low child (\texttt{false}), and solid lines represent transitions to the high child (\texttt{true}).}
    \label{fig:bdd_example}
\end{figure}

\paragraph{Example:}
Figure~\ref{fig:bdd_example} illustrates a BDD representation of the Boolean function:
\[
f(x_1, x_2, x_3) = (x_1 \land x_2 \land \neg x_3) \lor (\neg x_1 \land x_3).
\]
In this diagram, the dashed lines represent transitions to the \textit{low} child (a branch where the variable is \texttt{false}), while the solid lines represent transitions to the \textit{high} child (a branch where the variable is \texttt{true}).

To see the correspondence between the BDD and the Boolean formula, observe the two paths from the root node (\( v_1 \)) to the \( \textbf{1} \)-leaf (\( v_6 \)):
\begin{enumerate}
    \item One path goes through \( v_2 \) and \( v_4 \), representing the assignment \( x_1 = 1 \), \( x_2 = 1 \), and \( x_3 = 0 \). This corresponds to the term \((x_1 \land x_2 \land \neg x_3) \).
    \item The other path goes through \( v_3 \), representing the assignment \( x_1 = 0 \), \( x_3 = 1 \), and \( x_2 \) is not constrained. This corresponds to the term \( (\neg x_1 \land x_3) \).
\end{enumerate}

Thus, this BDD compactly encodes the disjunction of these two conjunctive terms. This example was adapted from \cite{bryant_binary_2018}.

\subsubsection{Converting BDDs into Goal Expressions}
Our method, TIDE, decomposes a problem with LTL$_f$ goals into a sequence of smaller subproblems with reach-avoid goals. These subproblems correspond to specific transitions in the automaton. To achieve this, TIDE translates the edge conditions represented as BDDs into goal expressions in propositional logic, which is suitable for defining goals and constraints in the PDDL (Planning Domain Definition Language) format. This process involves analyzing the logical structure of the BDD, particularly focusing on its disjunctive normal form (DNF).

\noindent
{\bf Procedure for Conversion}

\paragraph{Step 1: Identify Required Predicates}
\label{subsec:identify_required_predicates}

The first step in converting a BDD into a goal expression involves identifying \textit{required predicates}, which are predicates common to all conjunctive terms in the formula represented by the BDD. This process is formalized in the following lemma.

\begin{lemma}
Let \( f \) be a Boolean function over variables \( x_1, x_2, \ldots, x_n \). For a variable \( x_i \):
\begin{enumerate}
    \item \( x_i \) must be \texttt{true} for \( f \) to evaluate to \( 1 \) if and only if \( f |_{x_i \gets 0} = \textbf{0} \)\\ (the cofactor of \( f \) with \( x_i = 0 \) is identically false).
    \item \( x_i \) must be \texttt{false} for \( f \) to evaluate to \( 1 \) if and only if \( f |_{x_i \gets 1} = \textbf{0} \)\\ (the cofactor of \( f \) with \( x_i = 1 \) is identically false).
\end{enumerate}
\end{lemma}

\begin{proof}
The proof is based on the Shannon expansion of \( f \) with respect to \( x_i \):
\[
f = \left(x_i \land f |_{x_i \gets 1}\right) \lor \left(\lnot x_i \land f |_{x_i \gets 0}\right).
\]

\begin{enumerate}
    \item \textbf{ \( f |_{x_i \gets 0} = \textbf{0} \)\\}
        \(
        \iff f = \left(x_i \land f |_{x_i \gets 1}\right) \lor \left(\lnot x_i \land \textbf{0}\right)\\
        \)
        \(
        \iff f = x_i \land f |_{x_i \gets 1}\\
        \)
        \(\iff\) For \( f \) to evaluate to \( 1 \), \( x_i \) must be \texttt{true}. \qed

    \item \textbf{\( f |_{x_i \gets 1} = \textbf{0} \)\\}
        \(
        f = \left(x_i \land \textbf{0}\right) \lor \left(\lnot x_i \land f |_{x_i \gets 0}\right)\\
        \)
        \(
        f = \lnot x_i \land f |_{x_i \gets 0}\\
        \)
        \(\iff\) For \( f \) to evaluate to \( 1 \), \( x_i \) must be \texttt{false}. \qed

\end{enumerate}
\end{proof}
\paragraph{Procedure}

For each Boolean variable \( x_i \) in the BDD:
\begin{enumerate}
    \item Compute the cofactors \( f |_{x_i \gets 1} \) and \( f |_{x_i \gets 0} \).
    \item Evaluate whether one of the cofactors is \( \textbf{0} \):
    \[
    f |_{x_i \gets b} = \textbf{0}, \quad b \in \{0, 1\}.
    \]
    \begin{itemize}
       \item If \( f |_{x_i \gets 1} = \textbf{0} \), then \( x_i \) must be \texttt{false} for \( f \) to evaluate to \( 1 \).
       \item If \( f |_{x_i \gets 0} = \textbf{0} \), then \( x_i \) must be \texttt{true} for \( f \) to evaluate to \( 1 \).
   \end{itemize}
   \item If either of the above conditions is satisfied, \( x_i \) is added into a set of \textit{required} predicates with its corresponding truth value.
\end{enumerate}

The required predicates are then combined into a conjunction:
\[
\varphi_{\text{required}} = \bigwedge_{i \in \text{required}} \begin{cases} 
x_i & \text{if } f |_{x_i \gets 0} = \textbf{0}, \\
\lnot x_i & \text{if } f |_{x_i \gets 1} = \textbf{0}.
\end{cases}
\]

\paragraph{Step 2: Separate Disjunctive Conditions}
\label{subsec:separate_disjunctive_conditions}

The goal of this step is to simplify the BDD by ``removing'' the required predicates and isolating any remaining disjunctive structure. If the resulting BDD becomes identically \( \textbf{1} \) (\texttt{true}), the edge condition is purely conjunctive, and no further processing is needed. Otherwise, we must extract and represent the disjunctive structure as a logical goal expression.

\paragraph{1) Restricting Required Predicates.}
To simplify the BDD, the required predicates identified in Step 1 are restricted to their corresponding truth values.\\ Formally, let \( \varphi_{\text{required}} \) represent the conjunction of required predicates:
\[
\varphi_{\text{required}} = \bigwedge_{i \in \text{required}} \begin{cases} 
x_i & \text{if } f |_{x_i \gets 0} = \textbf{0}, \\
\lnot x_i & \text{if } f |_{x_i \gets 1} = \textbf{0}.
\end{cases}
\]
The simplified BDD is then obtained using the restriction operation:
\[
f_{\text{disjunctive}} = f |_{\varphi_{\text{required}}}.
\]
If \( f_{\text{disjunctive}} = \textbf{1} \), the BDD is fully simplified, and no disjunctive terms remain. Otherwise, the remaining structure of \( f_{\text{disjunctive}} \) must be analyzed to extract disjunctive conditions.

\paragraph{2) Extracting Disjunctive Predicates.}
When \( f_{\text{disjunctive}} \neq \textbf{1} \), it represents a disjunction of conjunctive terms. To simplify the process of extracting these terms, we first identify the set of all predicates that appear in \( f_{\text{disjunctive}} \). This is accomplished using the following lemma:

\begin{lemma}
A predicate \( x_i \) is part of the disjunctive structure of \( f \) if and only if the simplified BDD \( f_{\text{disjunctive}} \) is affected by the restriction of \( x_i \) to \( \texttt{true} \) or \( \texttt{false} \). Formally:
\[
x_i \text{ is part of } f_{\text{disjunctive}} \iff f_{\text{disjunctive}} \neq f_{\text{disjunctive}} |_{x_i \gets b}, \quad \text{for some } b \in \{0, 1\}.
\]
\end{lemma}

\begin{proof}
To prove this claim, we show its contrapositive:
\[
x_i \text{ is not part of } f_{\text{disjunctive}} \iff f_{\text{disjunctive}} = f_{\text{disjunctive}} |_{x_i \gets 0} = f_{\text{disjunctive}} |_{x_i \gets 1}.
\]
This follows directly from the Shannon expansion:
\[
f_{\text{disjunctive}} = \left(x_i \land f_{\text{disjunctive}} |_{x_i \gets 1}\right) \lor \left(\lnot x_i \land f_{\text{disjunctive}} |_{x_i \gets 0}\right).
\]
Suppose \( f_{\text{disjunctive}} |_{x_i \gets b} = f_{\text{disjunctive}} \) for all \( b \in \{0, 1\} \),\\
\(\iff f_{\text{disjunctive}} = \left(x_i \land f_{\text{disjunctive}}\right) \lor \left(\lnot x_i \land f_{\text{disjunctive}}\right)\)\\
\(\iff\) the value of \( x_i \) does not affect \( f_{\text{disjunctive}} \), and \( x_i \) is not part of the formula. \qed
\end{proof}

\paragraph{Procedure}

For each Boolean variable \( x_i \) in the BDD:
\begin{enumerate}
    \item Compute the cofactors \( f_{\text{disjunctive}} |_{x_i \gets 1} \) and \( f_{\text{disjunctive}} |_{x_i \gets 0} \).
    \item If \( f_{\text{disjunctive}} |_{x_i \gets 1} \neq f_{\text{disjunctive}} \) or \( f_{\text{disjunctive}} |_{x_i \gets 0} \neq f_{\text{disjunctive}} \),  \( x_i \) is added into a set of \textit{disjunction predicates} with its corresponding truth value.
\end{enumerate}

\paragraph{3) Iterating Over Subsets of Disjunction Predicates.}
Once the relevant \textit{disjunction predicates} are identified, we iterate over all subsets of these predicates to extract conjunctive terms. For a subset \( S \) of predicates, define:
\[
\psi_S = \bigwedge_{x_i \in S} \begin{cases}
x_i & \text{if } x_i \text{ is positive in } S, \\
\lnot x_i & \text{if } x_i \text{ is negative in } S.
\end{cases}
\]
Using the restriction operation, we evaluate:
\[
f_{\text{restricted}} = f_{\text{disjunctive}} |_{\psi_S}.
\]
If \( f_{\text{restricted}} = \textbf{1} \), then \( \psi_S \) represents a valid conjunctive term in the disjunction, and is added to the set of \textit{valid conjuncts}.

\paragraph{4) Constructing the Final Disjunction.}
The extracted conjunctive terms are combined into a disjunctive expression:
\[
\varphi_{\text{disjunctive}} = \bigvee_{S \subseteq \text{valid conjuncts}} \psi_S.
\]

\paragraph{Step 3: Combine into a Goal Expression}
\label{subsec:combine_goal_expression}

The third and final step in the process involves constructing the complete propositional goal expression by combining the required predicates \( \varphi_{\text{required}} \) (from Step 1) with the disjunctive conditions \( \varphi_{\text{disjunctive}} \) (from Step 2). 

The final goal expression \( \varphi_{\text{goal}} \) is constructed as following:
\[
\varphi_{\text{goal}} = \varphi_{\text{required}} \land \varphi_{\text{disjunctive}}.
\]

\paragraph{Example}
Consider a BDD encoding the following DNF formula:
\[
(a \land b \land \neg d) \lor (a \land c \land \neg d) \lor (a \land e)
\]
The conversion process is as follows:
\begin{enumerate}
    \item \textbf{Identify required predicates:} Extract predicates that are common across all conjunctive terms. In this case, \( a \) appears in every conjunct, so it is a required predicate.
    \item \textbf{Separate disjunctive conditions:} Remove the common predicate (\( a \)) from each conjunct, leaving:
    \[
    (b \land \neg d) \lor (c \land \neg d) \lor e
    \]
    \item \textbf{Combine into a goal expression:} The final goal expression is constructed by combining the required predicate with the disjunctive conditions:
    \[
    a \land \left((b \land \neg d) \lor (c \land \neg d) \lor e\right)
    \]
\end{enumerate}

\paragraph{Computational Complexity}

Let \( n \) denote the number of Boolean variables in the BDD and \( m \) the number of nodes in the BDD. Each restriction operation (\( f |_{x_i \gets b} \)) has a worst-case complexity of \( O(m) \).

\begin{itemize}
    \item \textbf{Identifying Required Predicates}: \( O(nm) \)\\
    This step involves iterating over all Boolean variables in the BDD and computing their cofactors \( f |_{x_i \gets 1} \) and \( f |_{x_i \gets 0} \). Each cofactor computation requires evaluating a restriction operation. The complexity of this step is \( O(nm) \), as we perform \( 2n \) restrictions, each taking \( O(m) \) time.
    \item \textbf{Identifying Disjunctive Predicates}: \( O(nm) \)\\
    This involves iterating over the remaining BDD variables and checking if the restricted BDD changes when each variable is set to \( \texttt{true} \) or \( \texttt{false} \). This process is similar to Step 1 and has a complexity of \( O(n' m') \), where \( n' \) is the number of remaining variables, and \( m' \) is the size of the simplified BDD. In the worst case, where all variables are retained (\( n' = n \)) and the BDD size remains unchanged (\( m' = m \)), the complexity of this step remains \( O(nm) \).
    \item \textbf{Iterating Over Subsets of Disjunctive Predicates}: \( O(2^d \cdot m) \)\\
    Let \( d \) denote the number of disjunctive predicates. The number of subsets is \( 2^d \), and for each subset, a restriction operation is performed on the BDD, which has a complexity of \( O(m') \). Therefore, the complexity of this step is \( O(2^d \cdot m') \). In the worst case, where the BDD size remains unchanged (\( m' = m \)), the complexity of this step is \( O(2^d \cdot m) \).
\end{itemize}

The overall complexity of the procedure depends on \( n \) (the number of variables), \( m \) (the size of the BDD), and \( d \) (the number of disjunctive predicates). The dominating term is the exponential component \( O(2^d \cdot m) \), which highlights the necessity of minimizing \( d \). Extracting required predicates first is efficient because it reduces the size of the remaining disjunctive structure. Furthermore, in practical scenarios, most edge conditions in DFAs are simple conjunctions, meaning the BDD corresponding to \(f_{\text{disjunctive}}\) becomes identically \( \textbf{1} \) after Step 1, and the process terminates early avoiding the need to iterate over subsets of variables.

%% file: proposed_method.tex
\section{Proposed Method}
\label{ch:ProposedMethod}
Our method, TIDE (Trace-Informed Depth-first Exploration), combines the best of two worlds: the goal-progression provided by the automaton trace and the efficiency of classical off-the-shelf planners from compilation-based approaches. By dynamically generating and selecting promising DFA traces, we decompose a problem with LTL$_f$ goal into manageable subproblems with propositional reach-avoid goals. These subproblems can then be solved using highly-optimized classical planners. The efficiency of our approach relies on two key elements: a well-designed cost function to guide the selection of promising automaton traces, and a systematic backtracking mechanism that incorporates feedback from failed attempts to refine future trace selections.

The key steps of our method are as follows:
\begin{enumerate}
    \item \textbf{Convert the LTL$_f$ formula into a DFA}.\\
    This step translates the temporal logic specification into a Deterministic Finite Automaton (DFA), which serves as a state-based model that can be efficiently searched. We achieve this transformation using Spot, a powerful C++17 library for handling LTL and $\omega$-automata \cite{artho_spot_2016, shoham_spot_2022}. 

    \item \textbf{Search: The Selection-Realization-Backtracking Loop}.
    \begin{enumerate}
        \item \textbf{Phase 1: Select the most promising DFA trace}.\\
        The first step in solving the problem is to select a promising trace from the DFA using a variant of Uniform Cost Search. This systematic exploration evaluates all feasible paths from the initial DFA state, considering both straight paths and potential cycles. 
        Our search is guided by a cost function that prioritizes traces with a high likelihood of being successfully realized in the planning domain 
        (the definition and rationale behind this cost function are detailed later in Section~\ref{sec:EdgeCostCalculations}).

        Notably, we employ a variant of Hill Climbing: the search returns a trace as soon as it finds an accepting state with a lower cost than a subsequent accepting trace, opting for a locally optimal solution rather than merely the first found solution (see Section~\ref{sec:HillClimbingHeuristic} for details). Once a trace is selected, the search progress is ``frozen'' so that we can easily resume the process if we need to backtrack and select a different trace.
    
        \item \textbf{Phase 2: Realize the selected DFA trace in the domain}.\\
        After selecting a DFA trace, the next step is to realize this trace in the domain. Each transition in the trace is treated as a classical reach-avoid problem, allowing us to decompose the temporal planning problem into a sequence of simpler subproblems. We use off-the-shelf planners to solve these subproblems, progressively constructing a product graph in a depth-first manner. The idea of goal-progression is central here: the final state of the current subproblem becomes the initial state of the next subproblem, enabling us to concatenate the subplans smoothly into a single, cohesive plan for the entire temporal goal.

        \item \textbf{Phase 3: Backtrack if necessary}.\\
        If the realization of the selected trace is successful, the process is complete. However, if the realization fails, we employ a systematic backtracking strategy:
        \begin{itemize}
            \item Cache plans corresponding to the automaton trace’s prefix, which was successfully realized in the domain. This allows us to reuse these plans efficiently in future steps.
            \item Assign a significantly lower cost to the DFA transitions that have been successfully realized, encouraging the algorithm to reuse these transitions in future trace selections.
            \item Penalize the DFA transition that caused the failure by assigning it a high cost. If the search for a plan on this transition was exhaustive, we can completely discard the corresponding trace prefix from further consideration.
            \item Recalculate the total costs of all traces in the frozen priority queue from the selection phase, incorporating the updated edge costs from successful and failed transitions.
        \end{itemize}
        After adjusting the trace costs based on the feedback from the failed attempt, the algorithm backtracks and selects a new trace for exploration.
    \end{enumerate}
\end{enumerate}

The pseudocodes provided in the Appendix illustrate the core steps of the proposed method in detail. Algorithm 1 presents the high-level view of the TIDE approach, outlining how it iteratively selects and attempts to realize DFA traces until a valid plan is found or all options are exhausted. Algorithm 2, \textit{generate\_dfa\_trace()}, describes a variant of Uniform Cost Search strategy for selecting the most promising DFA trace. Algorithm 3, \textit{realize\_dfa\_trace\_without\_planner()}, demonstrates the procedure for realizing a DFA trace using hierarchical Breadth-First Search (BFS), incrementally building a product graph.  This variant uses BFS in place of the off-the-shelf planner, which offers a clearer understanding of how the method works in its simplest form. Finally, Algorithm 4, \textit{realize\_dfa\_trace\_with\_planner()}, depicts an alternative realization method that uses an off-the-shelf planner to solve classical reach-avoid planning problems corresponding to DFA transitions.

\subsection{Phase 1: Automaton Trace Selection}
\label{sec:Phase1}
\paragraph{Uniform Cost Search in DFA}
The trace selection phase uses a variant of Uniform Cost Search (UCS) to explore all possible DFA traces starting from the initial state. Each node in the search represents a state in the DFA and a trace from the start state to that node. Importantly, there could be multiple nodes that represent the same DFA state because they can correspond to different traces (paths in the automaton graph) leading to the same DFA state. Thus, each node corresponds to a unique trace from the start state to the current state, and it is this trace that we are interested in realizing within the domain.

The UCS algorithm maintains a priority queue of nodes, ordered by their trace cost. At each step, the algorithm pops the node with the lowest cost from the queue and explores its successors by expanding the trace to include outgoing DFA transitions. Each newly generated node is added to the priority queue, with its associated cost reflecting the cumulative cost of the trace up to that point. This approach is complete because UCS explores all possible traces from the start state, even those that involve cycles. For simplicity, we avoid self-transitions (i.e., DFA transitions from a state to itself), as they do not contribute to the progression of the trace and are equivalent to remaining in the same state.

Algorithm 2 in the Appendix provides a detailed illustration of this UCS-based trace selection process.

\paragraph{Hill Climbing Heuristic}
\label{sec:HillClimbingHeuristic}
To improve efficiency, we integrate a Hill Climbing heuristic into the UCS search. Rather than returning the first accepting trace that we find, we continue exploring for better solutions. The algorithm evaluates accepting traces based on cost, and it compares the cost of the current trace with a subsequently discovered accepting trace. When an accepting trace is found, it is cached along with its cost. If a subsequent trace has a lower cost, the algorithm replaces the previously cached trace by pushing the old one back onto the priority queue in favor of the new, better trace. This new trace is then cached, and the search proceeds. Otherwise, if the new trace has a higher cost, the previously cached trace is deemed ``locally optimal,'' and the algorithm returns it.

Additionally, we use a stopping mechanism to handle cases where no better trace is found within a certain length threshold. For example, if we discover an accepting trace of length 3 and we assume a length threshold of 4, the algorithm will only return that trace after it has explored all traces with a length of up to 7 (3 + 4) and found no better solution. This heuristic strikes a balance between exploration and efficiency: it prevents returning suboptimal traces too early while also establishing a stopping condition.

\paragraph{Edge-Cost Calculations}
\label{sec:EdgeCostCalculations}
The cost function for DFA transitions is a crucial component of our method, as it determines which DFA traces are prioritized during the Uniform Cost Search process. If we were to use a naive constant cost function—where each DFA edge has the same cost (e.g., 1)—the algorithm would always prioritize shorter traces. However, shorter traces do not necessarily result in easier or faster solutions in the planning domain. In fact, this problem is analogous to the ``sparsity of rewards'' issue in Reinforcement Learning, where intermediate rewards often simplify the search process for achieving long-horizon goals. 

In the context of our method, each DFA state represents an intermediate goal or checkpoint. Having more checkpoints in a trace can make subproblems easier to solve, as they provide guidance to the planner. For example, in the Blocksworld domain, if the goal is to reverse the order of two blocks (e.g., placing the yellow block on top of the red block), an intermediate checkpoint of placing the red block on the table can be extremely helpful. Similarly, in temporal problems requiring a sequence of objectives, intermediate goals serve as stepping stones toward the overall goal. Consequently, slightly longer traces with more checkpoints can often lead to faster solutions in the planning domain. This motivates the development of a more nuanced cost function.

\paragraph{Defining the Edge Cost Function}
\label{subsec:edge_cost}
We define the cost of a particular DFA edge based on the amount of ``work'' required to transition between DFA states. Specifically, we define the cost as the number of required predicates with differing truth assignments between the self-edge guard of the current DFA state and the guard of the transition edge.\\ \textbf{Note}: terms ``edge guard'' and ``edge condition'' are used interchangeably.
\begin{definition}[Edge Cost]
For a given DFA edge \( q_i \xrightarrow{g_{i,j}} q_j \), the cost is computed as:
\[
\text{cost}(q_i \rightarrow q_j) =
\begin{cases}
|\Delta(g_{i,j} \setminus g_{i,i})| & \text{if the self-edge } (q_i \xrightarrow{g_{i,i}} q_i) \text{ exists}, \\
|\text{req}(g_{i,j})| & \text{otherwise,}
\end{cases}
\]
where:
\begin{itemize}
    \item \( g_{i,j} \): the guard (edge condition) of the transition \( q_i \rightarrow q_j \).
    \item \( g_{i,i} \): the guard of the self-edge of state \( q_i \), representing conditions under which the DFA can remain in the same state. If no self-edge exists, \( g_{i,i} = \textbf{0} \) (\texttt{false}).
    \item \( \text{req}(g) \): the conjunction of literals (required predicates) extracted from the BDD representing \( g \), as defined in Section~\ref{subsec:identify_required_predicates}. These literals represent the conditions that must hold for the edge guard to be satisfied.
    \item \( \Delta(g_{i,j} \setminus g_{i,i}) \): the set of literals in \( \text{req}(g_{i,j}) \) that are not satisfied by \( \text{req}(g_{i,i}) \), formally defined as:
    \[
    \Delta(g_{i,j} \setminus g_{i,i}) = 
    \{ l \in \text{Literals}(\text{req}(g_{i,j})) \mid 
    \text{req}(g_{i,j}) \vDash l \text{ and }\text{req}(g_{i,i}) \nvDash l \},
    \]
    where \( \text{Literals}(\varphi) \) is the set of all literals (e.g., \( p \) or \( \neg p \)) appearing in the propositional formula \( \varphi \).
\end{itemize}
\end{definition}

The intuition behind this cost function is that \textit{required predicates} typically represent goals that necessitate explicit ``work'': changes in their truth values indicate that specific actions must be performed. In contrast, the remaining disjunctive conditions often represent constraints that do not necessarily demand direct effort but must still be satisfied (e.g., safety constraints). Thus, the cost function focuses on changes in the required predicates, estimating the ``work'' needed to achieve transitions between states. The Uniform Cost Search algorithm uses these edge costs to prioritize traces with easier transitions.

\paragraph{Example:}
\begin{figure}[H]
    \centering
    \includegraphics[width=0.7\textwidth]{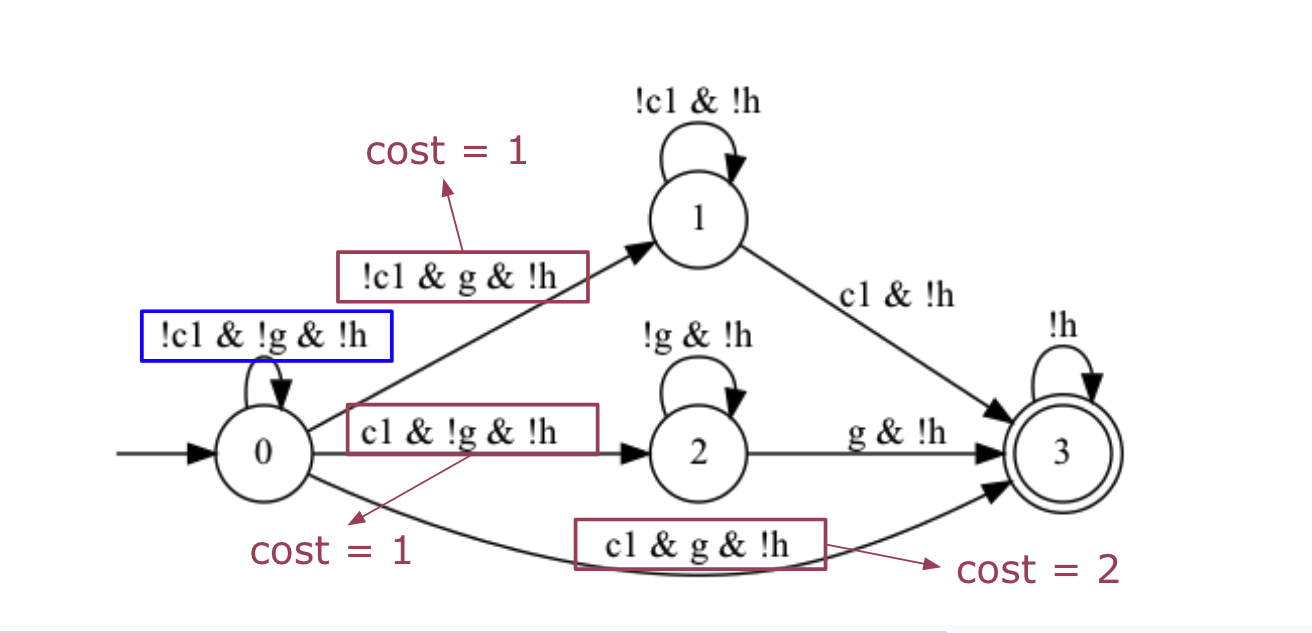} 
    \caption{DFA for Cost Calculation Example. Each edge is labeled with its guard, and edge costs are determined by the difference in required predicates (\( \Delta \)) between the guard of the edge leading to the new state (in red) and the guard of the current state's self-edge (in blue).}
    \label{fig:dfa-hamming-example}
\end{figure}
Consider the DFA in Figure~\ref{fig:dfa-hamming-example}, where each edge is labeled with its guard. 

We compute the edge cost of \((0 \rightarrow 2)\) as follows:
\begin{enumerate}
    \item \textbf{Extract Required Predicates}:\\ 
\(
\text{req}(g_{0,0}) = (\neg c1 \land \neg g \land \neg h), \quad \text{req}(g_{0,2}) = (c1 \land \neg g \land \neg h).
\)
\item \textbf{Compute Difference:}
\[\Delta(g_{0,2} \setminus g_{0,0}) = 
    \{ l \in \{c1, \neg g, \neg h\} \mid 
    (c1 \land \neg g \land \neg h) \vDash l \text{ and } (\neg c1 \land \neg g \land \neg h) \nvDash l \} = \{c1\}
\]
\item  \textbf{Compute Cost}: the edge cost is the size of the difference set
\[
\text{cost}(0 \rightarrow 2) = |\Delta(g_{0,2} \setminus g_{0,0})| = \{c1\} = 1.
\]
\end{enumerate}

Similarly, we compute costs of other DFA transitions as follows:
\begin{align*}
\text{cost}(0 \rightarrow 1) &= 1 \text{ ($g$ flips)}\\
\text{cost}(2 \rightarrow 3) &= 1 \text{ ($g$ flips)}\\
\text{cost}(0 \rightarrow 3) &= 2 \text{ (both $c1$ and $g$ flip)}
\end{align*}

\paragraph{Trace Ranking System}

In our approach, the ranking of DFA traces is based on the \textit{average cost per transition}. This metric plays a key role in the Uniform Cost Search (UCS) algorithm, which uses it as the ``cost'' value to prioritize candidate traces in its priority queue. To align with UCS terminology, we might refer to this metric as the ``trace cost.'' However, the term ``trace cost'' is commonly associated with the total cost (i.e., the sum of individual transition costs), whereas in our context, it specifically refers to the average cost. To avoid this potential confusion, we will use the term \textbf{trace rank} throughout this discussion.

\begin{definition}[Trace Rank]
For a given DFA trace consisting of \( n \) transitions \( T_1, T_2, \ldots, T_n \), the \textit{trace rank} is defined as:
\[
\text{trace rank} = \frac{\sum_{i=1}^{n} \text{cost}(T_i)}{n},
\]
where \( \text{cost}(T_i) \) is the edge cost of the \( i \)-th transition in the trace, and \( n \) is the trace length.
\end{definition}

This ranking system favors traces that are longer but composed of relatively low-cost transitions. Such traces often provide intermediate goals that simplify the decomposition of the problem into smaller subproblems, which makes planning more efficient. By averaging the transition costs over the trace length, we balance the trade-off between having more intermediate goals and ensuring that the transitions within the trace are not too effort-intensive.

\paragraph{Example:} 
Consider the same example shown in Figure~\ref{fig:dfa-hamming-example}. We compute the trace ranks for two possible traces as follows:
\begin{align*}
&\text{rank}(0 \rightarrow 2 \rightarrow 3) = \frac{\text{cost}(0 \rightarrow 2) + \text{cost}(2 \rightarrow 3)}{\text{length}(0 \rightarrow 2 \rightarrow 3)} = \frac{1 + 1}{2} = 1 \\
&\text{rank}(0 \rightarrow 3) = \frac{\text{cost}(0 \rightarrow 3)}{\text{length}(0 \rightarrow 3)} = \frac{2}{1} = 2
\end{align*}

In this case, the trace \( (0 \rightarrow 2 \rightarrow 3) \) has a lower trace rank (cost) compared to \( (0 \rightarrow 3) \). This indicates that the former trace, despite being longer, is preferred because it provides a helpful intermediate state 2 to guide the search.

\paragraph{Remarks}
The UCS algorithm uses trace ranks to select and prioritize traces, effectively guiding the search towards plans that are easier to realize in the planning domain.
While this cost function might result in non-optimality for UCS (e.g., longer traces with lower per-transition costs might be preferred), this is not problematic for our method. The goal of TIDE is not to find the optimal trace but rather to identify traces that are likely to result in feasible and easy-to-solve subproblems. This trace ranking system strikes a balance between trace length and the difficulty of transitions within the trace.

\paragraph{Handling Cycles in the DFA}

One challenge that arises in the DFA search is the presence of cycles, which could cause the UCS algorithm to continually extend traces with cycles due to their smaller transition costs. To discourage UCS from prioritizing cyclic traces, we introduce a penalty mechanism that increases the cost of cyclic transitions based on the number of times a state is revisited within the same trace.

For a DFA transition \( q_i \to q_j \), if the state \( q_j \) is revisited for the \( k \)-th time in the current trace, the cost of the transition is incremented by \( (k-1) \cdot CYCLE\_COST \), where \( CYCLE\_COST \) is a predefined penalty constant. Formally:
\[
\text{cost}(q_i \to q_j) = \text{base\_cost}(q_i \to q_j) + (k-1) \cdot CYCLE\_COST,
\]
where \( \text{base\_cost}(q_i \to q_j) \) is the original edge cost, as defined in Section~\ref{subsec:edge_cost}.

This penalty ensures that traces revisiting states multiple times incur increasingly higher costs, thereby discouraging UCS from prioritizing loopy traces. However, this mechanism does not compromise the completeness of the algorithm: if no lower-cost acyclic traces remain, UCS will eventually explore cyclic traces which might lead to a solution plan.

\paragraph{Proof of Completeness}
The UCS algorithm, applied in our context of DFA trace selection, is guaranteed to be complete. Completeness in this case means that if a solution (i.e., an accepting DFA trace that can be realized in the planning domain) exists, the algorithm will find it; otherwise, it will determine that solution doesn't exist.
\begin{itemize}
    \item Case 1: Solution exists
    \begin{proof}
    Let us consider the set of all possible DFA traces starting from the initial DFA state $q_0$ and terminating at some accepting state $q_a$. The UCS algorithm will enqueue every DFA state reachable from $q_0$, and for each state, it will generate the set of all possible transitions based on the DFA’s transition function. For a given state $q_i$, UCS will explore every possible trace from $q_0$ to $q_i$. Because UCS expands nodes in order of increasing cost, any trace that leads to an accepting state $q_f$ will eventually be reached.

Now, consider the possibility of cycles in the DFA. To ensure that the algorithm does not get stuck in an infinite loop caused by cycles in the DFA, we apply a penalty \( CYCLE\_COST \) to repeated transitions within the same trace. This penalty increases the cost of loopy traces, causing UCS to prioritize acyclic traces.

However, the addition of \( CYCLE\_COST \) does not prevent the algorithm from eventually exploring cycles. If all acyclic traces have been explored and no solution is found, UCS will eventually revisit cyclic traces, as their cumulative cost (even with the penalty) will become competitive. Therefore, UCS will still consider all possible DFA traces, including cyclic ones, ensuring that the search is complete.
\end{proof}
    \item Case 2: Solution does not exist
    \begin{proof}
Consider a situation where no valid trace can be realized in the domain. For each DFA trace selected by UCS, we attempt to solve the corresponding subproblems in the domain. If a subproblem fails (e.g., the planner cannot find a valid plan for a particular DFA transition), this failure acts as feedback that the current trace cannot be realized in the domain. If the planner's search was exhaustive, the failed prefix of the trace is marked as non-realizable, preventing TIDE from further exploring extensions of that trace. This pruning reduces the number of remaining traces and ensures TIDE does not repeatedly attempt to realize the same unfeasible traces.

Additionally, if we track all visited product states and determine that it's impossible to transition to some DFA state (e.g., state 2) from all possible product states corresponding to another DFA state (e.g., state 1), we can mark the DFA transition $(1 \rightarrow 2)$ as unrealizable in any trace. This edge elimination ensures that UCS will eventually exhaust the priority queue, provided there are a finite number of domain states.

As UCS progresses, the priority queue of DFA traces shrinks as traces are either explored or disregarded due to failed subproblems. Once all DFA traces have been explored or pruned, and the priority queue is empty, UCS concludes that no realizable trace exists, and the search terminates, signaling that no solution can be found.
\end{proof}
\end{itemize}

\subsection{Phase 2: Automaton Trace Realization in Domain}
\paragraph{Subproblem Creation}
\noindent
{\bf Subproblem Definition}:

Each subproblem \( \mathcal{P}_i \) in the automaton trace realization process is defined as a planning problem with propositional reach-avoid goals and corresponds to a DFA subtrace \( (q_{i} \rightarrow (q_{i})^* \rightarrow q_{i+1}) \), which is interpreted as:
    \begin{itemize}
        \item Zero or more self-transitions \((q_{i} \rightarrow q_{i})\) followed by,
        \item A transition to a new DFA state \((q_{i} \rightarrow q_{i+1})\).
    \end{itemize}
Formally, a subproblem \( \mathcal{P}_i \) is represented as a tuple \( (\mathcal{D}, \mathbf{s}^{\text{start}}_i, \varphi^{\text{goal}}_i, \varphi^{\text{constraint}}_i) \), where the components are defined as follows:

\begin{itemize}
    \item \textbf{Domain (\(\mathcal{D}\)):} 
    The interpreted task planning domain is denoted as \( \mathcal{D} \doteq (\mathcal{S}, \mathcal{A}, \mathcal{T}, \Prop, \ell) \), where:
    \begin{itemize}
        \item \( \mathcal{S} \): A finite set of states in the domain.
        \item \( \mathcal{A} \): A finite set of actions available in the domain.
        \item \( \mathcal{T} \subseteq \mathcal{S} \times \mathcal{A} \times \mathcal{S} \): A finite set of deterministic transitions such that for any $(\mathbf{s}, a, \mathbf{s}') \in \mathcal{T}$, there does not exist another $\mathbf{s}'' \in \mathcal{S}$ where $(\mathbf{s}, a, \mathbf{s}'') \in \mathcal{T}$.
        \item \( \Prop \): A finite set of atomic propositions.
        \item \( \ell: \mathcal{S} \rightarrow 2^{\Prop} \): A labeling function that maps each state \( s \in \mathcal{S} \) to the set of atomic propositions that are true in \( s \in \mathcal{S}\).
    \end{itemize}
    \item \textbf{Initial State (\(\mathbf{s}^{\text{start}}_i\)):} 
    The initial state \( \mathbf{s}^{\text{start}}_i \in \mathcal{S} \) is the starting point for the subproblem. It is defined as:
    \begin{itemize}
        \item \( \mathbf{s}^{\text{start}}_1 = \mathbf{s}^{\text{start}}\): for the first subproblem (\( i = 1 \)), the initial state is equal to the start state of the problem with LTL$_f$ goals.
        \item \( \mathbf{s}^{\text{start}}_i = \mathbf{s}^{\text{end}}_{i-1} \): for subsequent subproblems (\( i > 1 \)), the initial state is the final state reached during the execution of the plan for the previous subproblem \( \mathcal{P}_{i-1} \).
    \end{itemize}

    \item \textbf{Goal Condition (\(\varphi^{\text{goal}}_i\)):} 
    The goal condition \( \varphi^{\text{goal}}_i \) is a propositional formula over the set of atomic propositions \( \Prop \) that must be satisfied in the final state of the subproblem. Specifically:
    \begin{itemize}
        \item The final state \( \mathbf{s}^{\text{end}}_i \) must satisfy \( \ell(\mathbf{s}^{\text{end}}_i) \vDash \varphi^{\text{goal}}_i \).
        \item This goal is determined by the edge condition (guard) on the DFA transition \( (q_{i} \rightarrow q_{i+1}) \). The edge condition specifies the logical requirements that must be met to move from state \( q_i \) to state \( q_{i+1} \) in the DFA trace.
    \end{itemize}

    \item \textbf{Constraint Condition (\(\varphi^{\text{constraint}}_i\)):} 
    The constraint condition \( \varphi^{\text{constraint}}_i \) is a propositional formula over \( \Prop \) that must hold true in all intermediate states of the subproblem’s execution. Specifically:
    \begin{itemize}
        \item For every intermediate state \( \mathbf{s}^j_i \) in the execution of the plan,\\  \( \ell(\mathbf{s}^j_i) \vDash \varphi^{\text{constraint}}_i \), where \( j \in \{0, \ldots, n-1\} \).
        \item The constraint condition is derived from the self-edge condition \( (q_{i} \rightarrow q_{i}) \) of the current DFA state \( q_i \). This ensures that the subproblem adheres to the logical requirements of remaining in state \( q_i \) until the transition to \( q_{i+1} \) occurs.
    \end{itemize}
\end{itemize}

\paragraph{Generating Subproblems from DFA Transitions}
The process of subproblem creation relies on the edge condition (also known as the edge guard) of the transition connecting two distinct DFA states and, when applicable, the self-edge condition of the current DFA state. In our approach, we use the Spot library to construct DFAs, where edge conditions are represented using Binary Decision Diagrams (BDDs). BDDs provide a compact and efficient representation of boolean conditions that must be satisfied for transitions to occur. These diagrams are particularly useful for encoding logical constraints in a scalable manner. For example, in a Blocksworld domain, a BDD could express conditions such as "block A must be on the table" or "block B must be clear" as necessary predicates for transitioning between DFA states. 

The subproblem creation algorithm consists of four steps:
\paragraph{Step 1: Creating the Goal Expression for the Subproblem.}
This step involves converting the BDD representing the edge condition of the transition connecting two distinct DFA states into a goal expression. This goal represents the condition that must hold true in the final state of the subproblem and corresponds to the next DFA state.

\paragraph{Step 2: Creating a Constraint Expression.}
This step processes the BDD representing the self-edge condition of the current DFA state. The resulting expression acts as a constraint, ensuring that any intermediate state in the subplan adheres to the self-edge condition and corresponds to the current DFA state.

\paragraph{Step 3: Handling Constraints on Intermediate States.}
This step defines how constraints are imposed on intermediate states, addressing all possible scenarios to ensure proper synchronization with the automaton:
\begin{itemize}
    \item \textbf{Case 1: No constraints needed.} If the disjunction of the edge condition and the self-edge condition evaluates to \texttt{true}, no constraints are needed. The self-edge condition is simply the negation of the edge condition in this case.
    \item \textbf{Case 2: No self-edge exists.} If the self-edge condition is \texttt{false}, it indicates that remaining in the current DFA state indefinitely is not allowed; a transition to the next state must occur in a single step. The subproblem is thus simplified to finding a one-step plan that directly achieves the final goal. This approach ensures proper synchronization with the automaton, particularly when the LTL$_f$ formula involves the 'Next' temporal operator.
    \item \textbf{Case 3: Self-edge condition satisfied in the current state.} If the self-edge condition is already satisfied, the subproblem is constructed with the self-edge expression (from Step 2) as a constraint and the goal expression (from Step 1) as the final goal. This ensures that all intermediate states in the subplan adhere to the self-edge condition, allowing for the possibility of remaining in the same DFA state indefinitely; thus, the resulting plan can have an arbitrary length.
    \item \textbf{Case 4: Self-edge condition not satisfied in the current state.} When the self-edge condition is not met in the current state, an extended goal is created as the disjunction of the edge condition and self-edge condition. The planner attempts a single-step plan to satisfy this extended goal:
    \begin{itemize}
        \item If the resulting state satisfies the final goal, the subproblem is solved.
        \item If the resulting state satisfies only the self-edge condition, the scenario transitions to Case 3, where a new subproblem is created with the self-edge expression (from Step 2) as a constraint and the goal expression (from Step 1) as the final goal.
    \end{itemize}
\end{itemize}

\paragraph{Step 4: Instantiating the Subproblem.}
The final step instantiates the subproblem by combining the goal, constraints, and current domain state into a planning problem, which can then be solved using an off-the-shelf planner.

\paragraph{What happens if a current state doesn't have a self-edge?}
When a current state lacks a self-edge, we assume the condition is \textit{bddfalse}, which implies that staying in that state is impossible under any conditions. In this case, the subproblem's constraints, which apply to all intermediate states of the plan (excluding the initial and final states), become \( False \). Consequently, the only valid plan length is 1, indicating a single action that transitions directly from the start to the goal state without any intermediate states. This corresponds to the correct synchronization step in "Next" formulas, ensuring that the plan adheres to the strict step-by-step progression required by the DFA.

\paragraph{Examples of Subproblems}

Consider problems in the Blocksworld domain involving three blocks: \( b1 \), \( b2 \), and \( b3 \). The domain supports four primary actions: \texttt{pick-up}, \texttt{put-down}, \texttt{stack}, and \texttt{unstack}, defined as follows:
\begin{itemize}
    \item \texttt{pick-up}: Lifts a block from the table into the agent's hand if the block is clear and the agent's hand is empty.
    \item \texttt{put-down}: Places a held block onto the table, clearing the block and freeing the agent's hand.
    \item \texttt{stack}: Stacks a held block onto another block if the latter is clear, modifying the state to reflect the new configuration.
    \item \texttt{unstack}: Removes a block from another block and holds it, provided the block being moved is clear and the agent's hand is empty.
\end{itemize}

We analyze three problems involving stacking \( b2 \) on \( b1 \) and \( b3 \) on \( b2 \), each with slightly different logical requirements.

\begin{figure}[H]
    \centering
    \includegraphics[width=0.6\textwidth]{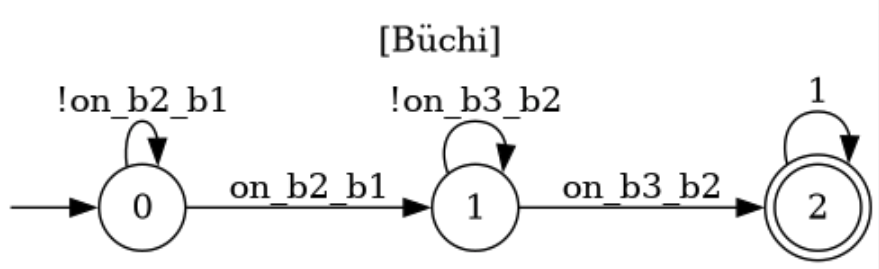} 
    \caption{DFA for Problem 1: \( F (on\_b2\_b1 \land X(F(on\_b3\_b2))) \).}
    \label{fig:dfa-problem1}
\end{figure}

\begin{figure}[H]
    \centering
    \includegraphics[width=0.6\textwidth]{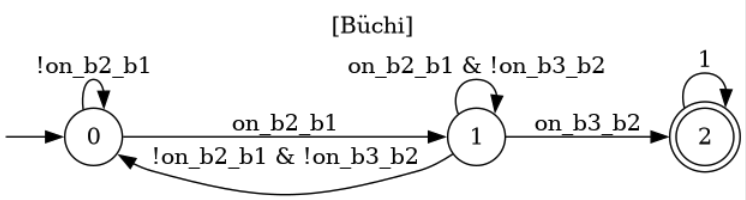} 
    \caption{DFA for Problem 2: \( F (on\_b2\_b1 \land X(on\_b3\_b2)) \).}
    \label{fig:dfa-problem2}
\end{figure}

\begin{figure}[H]
    \centering
    \includegraphics[width=0.6\textwidth]{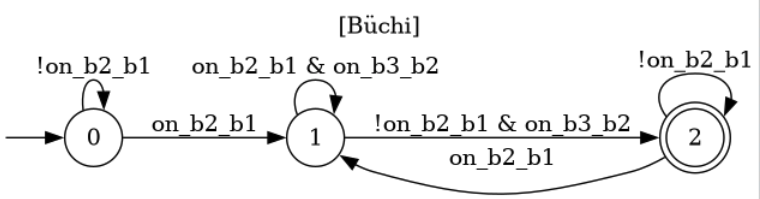} 
    \caption{DFA for Problem 3: \( F(on\_b2\_b1) \land G(on\_b2\_b1 \rightarrow X(on\_b3\_b2)) \).}
    \label{fig:dfa-problem3}
\end{figure}

\begin{itemize}
    \item \textbf{Problem 1}: \( F (on\_b2\_b1 \land X(F(on\_b3\_b2))) \)\\
    This formula requires \( b2 \) to be stacked on \( b1 \) at some point and \( b3 \) to be stacked on \( b2 \) at some later point. The automaton for this formula, shown in Figure~\ref{fig:dfa-problem1}, consists of three states. The transitions reflect the logical progression of satisfying \( on\_b2\_b1 \) and \( on\_b3\_b2 \).

    \item \textbf{Problem 2}: \( F (on\_b2\_b1 \land X(on\_b3\_b2)) \)\\
    This formula requires \( b3 \) to be stacked on \( b2 \) immediately after a certain state in which \( on\_b2\_b1 \) held true. The automaton, shown in Figure~\ref{fig:dfa-problem2}, allows the intermediate actions that do not disrupt \( on\_b2\_b1 \) before achieving \( on\_b3\_b2 \).

    \item \textbf{Problem 3}: \( F(on\_b2\_b1) \land G(on\_b2\_b1 \rightarrow X(on\_b3\_b2)) \)\\
    This formula enforces that \( b3 \) must be stacked on \( b2 \) immediately whenever \( b2 \) is stacked on \( b1 \). The automaton, shown in Figure~\ref{fig:dfa-problem3}, indicates that no intermediate actions are allowed once \( on\_b2\_b1 \) is satisfied.
\end{itemize}

Suppose we select the DFA trace \( 0 \rightarrow 1 \rightarrow 2 \) for each of these problems. We construct subproblems for each DFA transition \( (q_0 \rightarrow q_1) \) and \( (q_1 \rightarrow q_2) \), following the subproblem creation algorithm.

\paragraph{Subproblem 1 (Same for All Problems)}
The first DFA transition \( (q_0 \rightarrow q_1) \) has the edge condition \( on\_b2\_b1 \) and the self-edge condition \( \neg (on\_b2\_b1) \). Using Steps 1 and 2 of the algorithm:
\begin{itemize}
    \item \textbf{Goal}: \( on\_b2\_b1 \)
    \item \textbf{Constraints}: None.\\
    Note: \( ((on\_b2\_b1) \lor (\neg(on\_b2\_b1))) = \texttt{true}\) (\textbf{Case 1})
\end{itemize}

\paragraph{Subproblem 2 for Problem 1}
The second DFA transition \( (q_1 \rightarrow q_2) \) has the edge condition \( on\_b3\_b2 \) and the self-edge condition \( \neg (on\_b3\_b2) \). Using Steps 1 and 2:
\begin{itemize}
    \item \textbf{Goal}: \( on\_b3\_b2 \)
    \item \textbf{Constraints}: None.\\
     Note: \( ((on\_b3\_b2) \lor (\neg(on\_b3\_b2))) = \texttt{true}\) (\textbf{Case 1})
\end{itemize}

\paragraph{Subproblem 2 for Problem 2}
The second DFA transition \( (q_1 \rightarrow q_2) \) has the edge condition \( on\_b3\_b2 \) and the self-edge condition \( (on\_b2\_b1 \land \neg (on\_b3\_b2)) \). Using Steps 1 and 2:
\begin{itemize}
    \item \textbf{Goal}: \( on\_b3\_b2 \)
    \item \textbf{Constraints}: \( on\_b2\_b1 \land \neg (on\_b3\_b2) \) (\textbf{Case 3})
\end{itemize}

\paragraph{Subproblem 2 for Problem 3}
The second DFA transition \( (q_1 \rightarrow q_2) \) has the edge condition \( ((\neg (on\_b2\_b1)) \land on\_b3\_b2) \) and the self-edge condition \( (on\_b2\_b1 \land on\_b3\_b2) \). This scenario corresponds to \textbf{Case 4}: since the self-edge condition \(( on\_b2\_b1 \land on\_b3\_b2) \) is not satisfied in the current state 1, we create an \textit{extended goal expression} that is the disjunction of the edge condition and the self-edge condition:
\[
\text{Extended Goal} = (\neg (on\_b2\_b1) \land on\_b3\_b2) \lor (on\_b2\_b1 \land on\_b3\_b2) = on\_b3\_b2
\]
The planner attempts a single-step plan to achieve this extended goal. The possible outcomes are:
\begin{itemize}
    \item If the resulting state satisfies the final goal \( ((\neg (on\_b2\_b1)) \land on\_b3\_b2) \), the subproblem is solved.
    \item If the resulting state satisfies only the self-edge condition \( (on\_b2\_b1 \land on\_b3\_b2) \), the scenario transitions to \textbf{Case 3}, where a new subproblem is constructed as follows:
    \begin{itemize}
    \item \textbf{Goal}: \( \neg (on\_b2\_b1) \land on\_b3\_b2 \)
    \item \textbf{Constraints}: \( on\_b2\_b1 \land on\_b3\_b2 \)
\end{itemize}
\end{itemize}

By enforcing the extended goal in one step, we ensure proper synchronization with the automaton in the presence of `Next' operator.

Figures~\ref{fig:dfa-problem1},~\ref{fig:dfa-problem2}, and~\ref{fig:dfa-problem3} illustrate the automata for these problems.

\paragraph{Automaton Trace Realization Process}

\label{sec:Phase2}

\begin{definition}[Automaton Trace Realization]
Let \( \mathcal{D} \doteq (\mathcal{S}, \mathcal{A}, \mathcal{T}, \Prop, \ell) \) be an interpreted planning domain, and let \( \varphi_G \) be an LTL$_f$ formula defining the goal. Given a Deterministic Finite Automaton (DFA) \( \mathcal{A} = (Q, q_0, \Sigma, \delta, F) \) constructed from \( \varphi_G \), an automaton trace \( \tau = (q_1, (q_1)^*, q_2, (q_2)^*, q_3, \ldots, (q_{n-1})^*, q_n) \in Q^{+} \) is said to be \textit{realizable} in \( \mathcal{D} \) if the following conditions hold:

\begin{enumerate}
    \item The trace \( \tau \) can be decomposed into a sequence of subproblems \( \{\mathcal{P}_1, \mathcal{P}_2, \ldots, \mathcal{P}_n\} \), where each subproblem \( \mathcal{P}_i \) corresponds to a DFA subtrace \( (q_{i} \rightarrow (q_{i})^* \rightarrow q_{i+1}) \). 

    \item Each subproblem \( \mathcal{P}_i \) has a valid plan \( \pi_i \) that satisfies its goal and constraint conditions. The final state \( \mathbf{s}^{\text{end}}_i \) reached by executing \( \pi_i \) serves as the initial state \( \mathbf{s}^{\text{start}}_{i+1} \) for the next subproblem \( \mathcal{P}_{i+1} \). This ensures continuity across subproblems.
    
    \item The concatenated plan \( \Pi = \pi_1 \circ \pi_2 \circ \ldots \circ \pi_n \), formed by combining the individual plans for each subproblem, satisfies the original temporal goal \( \varphi_G \). Specifically, the sequence of states \( \mathfrak{S}(\Pi) \) generated by executing \( \Pi \) in \( \mathcal{D} \) must satisfy\\ \( \mathfrak{S}(\Pi) \vDash \varphi_G \).
\end{enumerate}

\end{definition}

The process of automaton trace realization can be formally outlined as follows:
\begin{enumerate}
    \item \textbf{Step 1: Creating a sequence of subproblems with reach-avoid goals.}\\
    Construct a series of classical planning subproblems with propositional reach-avoid goals \( \{\mathcal{P}_1, \mathcal{P}_2, \ldots, \mathcal{P}_n\} \) where each \( \mathcal{P}_i =  (\mathcal{D}, \mathbf{s}^{\text{start}}_i, \varphi^{\text{goal}}_i, \varphi^{\text{constraint}}_i) \) corresponds to a subtrace \( (q_{i} \rightarrow (q_{i})^* \rightarrow q_{i+1}) \) in the DFA trace \( \tau \). This structure ensures that the final state \( \mathbf{s}^{\text{end}}_{i} \) of the execution for subproblem \( \mathcal{P}_i \) becomes the initial state \( \mathbf{s}^{\text{start}}_{i+1} \) for the next subproblem \( \mathcal{P}_{i+1} \), representing a goal progression.
    
    \item \textbf{Step 2: Solving subproblems using an off-the-shelf planner.}\\
    Apply a classical planner to each subproblem \( \mathcal{P}_i \) to obtain a valid plan \( \pi_i \) that transitions from \( \mathbf{s}^{\text{start}}_{i} \) to \( \mathbf{s}^{\text{end}}_{i} \) and $\ell(\mathbf{s}^{\text{end}}_{i}) \vDash \varphi^{\text{goal}}_i$. The execution \( E_i \) of each \( \pi_i \) must be feasible within \( \mathcal{D} \) and all intermediate states must satisfy the contraint formula: $\ell(\mathbf{s}^j_i) \vDash \varphi^{\text{constraint}}_i$ for all $j \in \{0, \ldots, n-1\}$.
    
    \item \textbf{Step 3: Concatenating subplans.}\\
    Combine the plans \( \pi_1, \pi_2, \ldots, \pi_n \) to form a single composite plan \( \Pi = \pi_1 \circ \pi_2 \circ \ldots \circ \pi_n \). The structure of the subproblems, where the start state of \( \mathcal{P}_{i+1} \) matches the final state of plan execution for \( \mathcal{P}_i \), enables smooth concatenation of subplans.
\end{enumerate}

\paragraph{Example: Translating a DFA Transition into Subproblems}
To illustrate how our method creates subproblems from a DFA trace, consider the following LTL$_f$ formula:
\[
\varphi_G = (F \left( \texttt{on\_b2\_b1} \land (X \, (F \, \texttt{on\_b3\_b2} \right))))
\]
This formula expresses a temporal goal where block $b2$ must be placed on $b1$, followed at some later point by block $b3$ being placed on $b2$. The automaton constructed from this formula contains three states and transitions that encode the progression of satisfaction of subgoals.

\begin{figure}[H]
    \centering
    \includegraphics[width=0.95\textwidth]{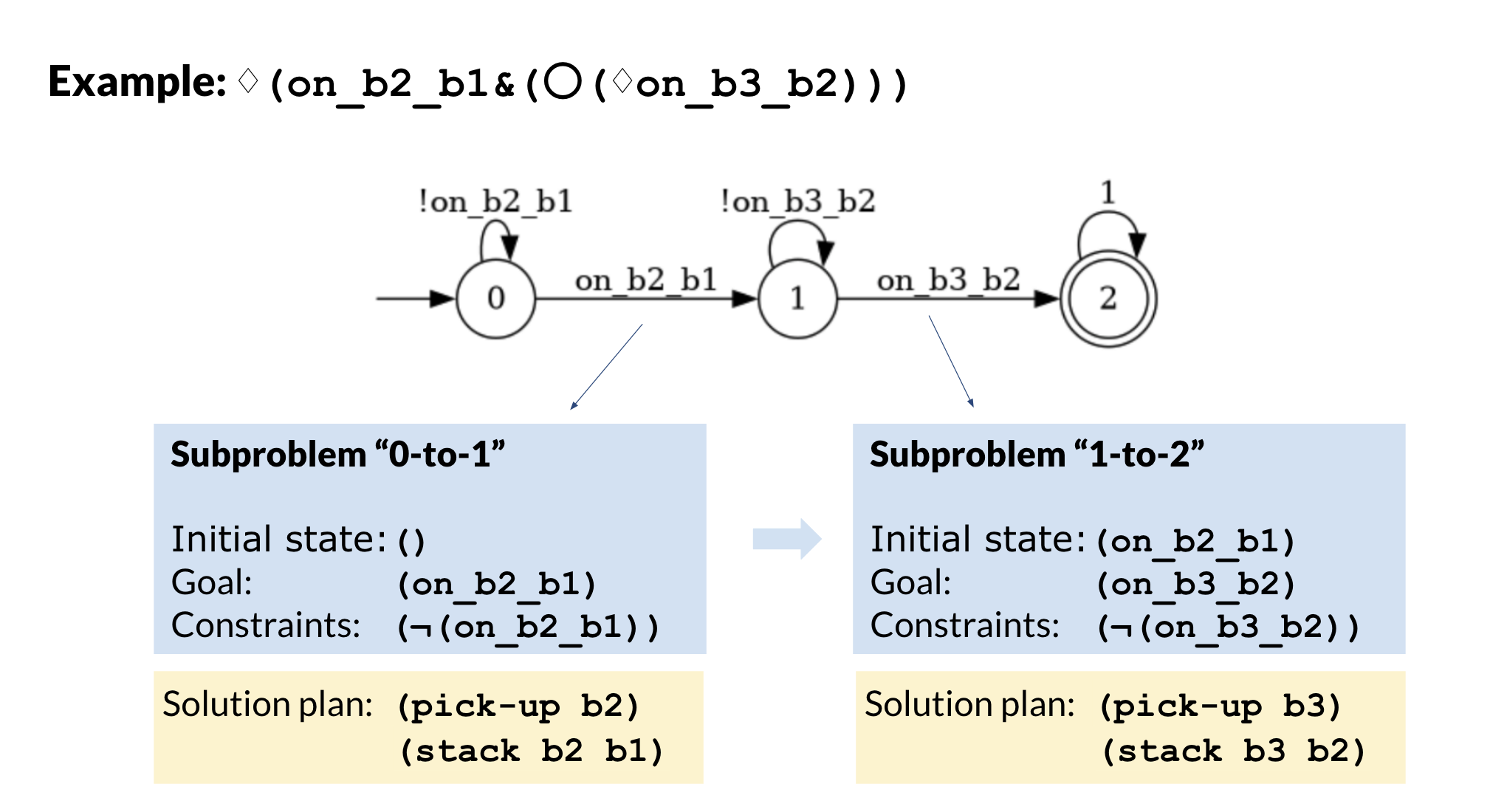}
    \caption{Example of subproblem creation for the LTL$_f$ formula $F(\texttt{on\_b2\_b1} \land (X \, (F \, \texttt{on\_b3\_b2})))$ based on DFA trace $0 \rightarrow 1 \rightarrow 2$. Each subproblem is formulated with goal and constraint expressions derived from the corresponding DFA transitions.}
    \label{fig:subproblem-example}
\end{figure}

Assume we are realizing the DFA trace $0 \rightarrow 1 \rightarrow 2$. This trace can be broken into two subproblems:

\begin{itemize}
    \item \textbf{Subproblem 0-to-1}: This subproblem corresponds to realizing the transition from DFA state $q_0$ to $q_1$.
    \begin{itemize}
        \item \textit{Goal condition}: $\texttt{on\_b2\_b1}$ (derived from the edge condition).
        \item \textit{Constraint condition}: $\neg \texttt{on\_b2\_b1}$ (from the self-loop condition at $q_0$).
        \item \textit{Initial state}: () \quad or \quad $\neg \texttt{on\_b2\_b1} \land \neg \texttt{on\_b3\_b2}$

        \item \textit{Plan}: $\texttt{(pick-up b2)}$, $\texttt{(stack b2 b1)}$
    \end{itemize}
    
    \item \textbf{Subproblem 1-to-2}: This subproblem corresponds to the next DFA transition from $q_1$ to $q_2$.
    \begin{itemize}
        \item \textit{Goal condition}: $\texttt{on\_b3\_b2}$
        \item \textit{Constraint condition}: $\neg \texttt{on\_b3\_b2}$ (from the self-loop at $q_1$)
        \item \textit{Initial state}: ($\texttt{on\_b2\_b1}$) \quad or \quad $\texttt{on\_b2\_b1} \land \neg \texttt{on\_b3\_b2}$

        \item \textit{Plan}: $\texttt{(pick-up b3)}$, $\texttt{(stack b3 b2)}$
    \end{itemize}
\end{itemize}
By solving these subproblems individually and chaining their plans together, we obtain the full plan for realizing the entire DFA trace:
\begin{verbatim}
(pick-up b2)
(stack b2 b1)
(pick-up b3)
(stack b3 b2)
\end{verbatim}
This example demonstrates how each DFA transition—defined by its edge guard and self-loop condition—naturally induces a well-structured subproblem with propositional reach-avoid goals. The subplans are generated independently using classical off-the-shelf planners and then combined to form a coherent plan that satisfies the temporal goal encoded by the DFA. Figure~\ref{fig:subproblem-example} visually depicts the DFA and corresponding subproblems created for this trace.

\paragraph{Methods for DFA Trace Realization}
TIDE provides two methods for realizing DFA traces in the domain: (1) using hierarchical Breadth-First Search (BFS) without a planner and (2) using an off-the-shelf planner to solve each subproblem. These approaches are detailed in Algorithm 3 and Algorithm 4 in the Appendix, respectively.

\paragraph{Hierarchical Breadth-First Search}
In this approach, we incrementally realize the DFA trace \( \tau = (q_0, q_1, \ldots, q_n) \) within the domain \( \mathcal{D} \) using a hierarchical Breadth-First Search (BFS) strategy, as detailed in Algorithm 3. Each DFA state \( q_i \in Q\) in the trace is paired with a set of product states, represented as \( (s, q_i) \), where \( s \in \mathcal{S} \). The hierarchical nature of the approach lies in its step-by-step progression: it first employs a standard BFS algorithm to solve the subproblem \( \mathcal{P}_i =  (\mathcal{D}, \mathbf{s}^{\text{start}}_i, \varphi^{\text{goal}}_i, \varphi^{\text{constraint}}_i) \) corresponding to the current DFA transition \( (q_{i} \rightarrow q_{i+1}) \), before moving on to the subsequent subproblem \( \mathcal{P}_{i+1} \) associated with the next DFA state \( q_{i+1} \).

The process begins by initializing a separate queue for each DFA state \( q_i \) in \( \tau \). These queues handle independent BFS processes for each subproblem. The initial product state \( (s_0, q_0) \), where \( s_0 \) is the domain's initial state, is enqueued. The BFS proceeds by dequeuing the current product state \( (s, q_i) \) and generating successors, which are new product states \( (s', q') \) obtained by applying valid transitions in \( \mathcal{D} \). Each valid transition corresponds to an action \( a \in \mathcal{A} \) such that \( (s, a, s') \in \mathcal{T} \). The exploration can result in the following cases:
\begin{enumerate}
    \item If \( q' = q_i \), the DFA state remains the same, indicating a transition to a new domain state while staying in the current DFA state. In this case, the product state \( (s', q_i) \) is added back to the same queue for continued exploration.
    \item If \( q' = q_{i+1} \), we have successfully transitioned to the next DFA state in the trace. This indicates that the subproblem \( \mathcal{P}_i \) is realized, and the algorithm initializes a new queue for \( q_{i+1} \) to begin solving \( \mathcal{P}_{i+1} \).
    \item If \( q' \) does not match either \( q_i \) or \( q_{i+1} \), it indicates a transition to an unintended DFA state, which is disregarded as it does not align with the current trace \( \tau \).
\end{enumerate}

If a valid product state \( (s', q_{i+1}) \) satisfying the goal \( \varphi^{\text{goal}}_i \) is found, the transition \( (q_{i} \rightarrow q_{i+1}) \) is considered realized, and the algorithm moves on to solve the next subproblem \( \mathcal{P}_{i+1} \). However, if no valid product state is found, the algorithm backtracks to the previous DFA state \( q_{i-1} \) and explores alternative paths. This backtracking ensures completeness by exhaustively searching for feasible solutions without prematurely committing to any single path.

\paragraph{Search with Off-the-Shelf Planners}
The second method for realizing DFA traces \( \tau = (q_0, q_1, \ldots, q_n) \) within \( \mathcal{D} \) involves decomposing the realization into classical planning subproblems \( \{\mathcal{P}_1, \mathcal{P}_2, \ldots, \mathcal{P}_n\} \) and applying an off-the-shelf planner (Algorithm 4). Each subproblem \( \mathcal{P}_i =  (\mathcal{D}, \mathbf{s}^{\text{start}}_i, \varphi^{\text{goal}}_i, \varphi^{\text{constraint}}_i) \) corresponds to a DFA transition \( (q_{i} \rightarrow q_{i+1}) \) and is defined by the edge condition (guard) that determines the goal \( \varphi^{\text{goal}}_i \).

The realization process begins with the subproblem \( \mathcal{P}_1 =  (\mathcal{D}, \mathbf{s}^{\text{start}}_1, \varphi^{\text{goal}}_1, \varphi^{\text{constraint}}_1) \), where \( \mathbf{s}^{\text{start}}_1 \) is the initial state of the original problem with LTL$_f$ goals. A classical planner, such as A* or Fast Downward, is invoked to find a valid plan \( \pi_1 \) that satisfies the subproblem's conditions. The execution \( E_i \) of \( \pi_i \) must be feasible in \( \mathcal{D} \) and must transition from \( \mathbf{s}^{\text{start}}_i \) to \( \mathbf{s}^{\text{end}}_i \) such that $\ell(\mathbf{s}^{\text{end}}_{i}) \vDash \varphi^{\text{goal}}_i$. If a valid plan \( \pi_i \) is found, the next subproblem \( \mathcal{P}_{i+1} = (\mathcal{D}, \mathbf{s}^{\text{end}}_i, \varphi^{\text{goal}}_{i+1}, \varphi^{\text{constraint}}_{i+1}) \) is constructed, where \( \mathbf{s}^{\text{end}}_i \) becomes the initial state for the next transition.

If the planner fails to find a valid plan for \( \mathcal{P}_i \), the subproblem is marked as unsolvable, and the algorithm backtracks to explore alternative DFA transitions from \( q_{i-1} \). This process ensures that all possible paths are considered. The use of an off-the-shelf planner is particularly useful for more complex domains where heuristic-based planning can significantly reduce the search space. By solving each subproblem \( \mathcal{P}_i \) and concatenating the resulting plans \( \pi_1, \pi_2, \ldots, \pi_n \), the method constructs a full plan \( \Pi = \pi_1 \circ \pi_2 \circ \ldots \circ \pi_n \) that realizes \( \tau \) and satisfies the temporal goal \( \varphi_G \) in \( \mathcal{D} \).

\subsection{Phase 3: Backtracking}
\label{sec:Phase3}
The backtracking phase is triggered when the realization of a selected DFA trace fails. In this case, the algorithm must adjust its strategy by incorporating feedback from the failed attempt and re-evaluating the remaining DFA traces. This systematic backtracking mechanism ensures that failed transitions are penalized and successful transitions are rewarded, guiding the search process toward feasible and efficient traces.

\paragraph{Handling Suboptimal DFA Traces}
To handle suboptimal DFA traces efficiently, we implement a strategy that monitors unintended DFA transitions. These deviations occur when the search in the domain leads to a DFA state that does not correspond to the expected next state in the current trace. If the number of these unintended transitions exceeds a predefined threshold, the algorithm temporarily ``gives up'' on the current DFA trace. This failed trace is not permanently discarded; instead, we assign a higher cost to the transition that caused the failure, deprioritizing the trace in future iterations. By temporarily suspending the trace, we prevent suboptimal traces from monopolizing resources, while allowing the search to return to it later if no better solution is found. 

Additionally, to improve efficiency, we could employ parallel searches across multiple DFA traces, distributing computational resources more effectively. In some cases, a time-out mechanism could also be implemented to ensure the search doesn't get stuck on a single trace for too long.

\paragraph{Caching Realized Transitions}
When a transition in the DFA trace is successfully realized in the domain, the algorithm caches subplans corresponding to all successfully-realized prefixes of the trace. This caching mechanism is vital for improving efficiency because it allows us to reuse previously computed subplans in future iterations. Each subproblem solved for a DFA transition corresponds to a specific part of the trace, so caching enables the algorithm to avoid redundant calculations when exploring similar traces or re-attempting previously failed traces. The cached plans are stored along with their associated DFA prefix paths and can be immediately reused if the same prefixes are encountered again. 

\paragraph{Cost Updates for Success and Failure}
The cost update mechanism plays a central role in the backtracking process. When a transition is successfully realized, we assign it a significantly lower cost in the DFA. This incentivizes the algorithm to reuse the transition in future trace selections, as lower costs make it more likely that the trace will be prioritized in the Uniform Cost Search.

Conversely, if a transition fails—meaning the planner could not find a valid plan to satisfy the transition's edge condition—its cost is increased substantially. This higher cost serves as a penalty, disincentivizing the selection of traces that involve this transition. If the failure occurs after an exhaustive search, indicating that the transition cannot be realized under any conditions, we completely discard this trace prefix from further consideration in future UCS iterations.

By recalculating the costs of all traces in the frozen priority queue based on this feedback, we ensure that future trace selections are informed by both successful and failed transitions. 

%% file: experiment_results.tex
\section{Experiment Results}
\label{ch:ExperimentResults}

This section presents a comparative analysis of the proposed method (TIDE) against several established frameworks: FOND4LTL$_f$ \cite{fuggitti_fond_2020}, Exp \cite{baier_planning_2006}, Poly \cite{torres_polynomial-time_2015}, and Plan4Past \cite{bonassi_planning_2023}. Exp and Poly are state-of-the-art methods for planning with LTL$_f$ goals, translating these goals into non-deterministic finite automata (NFA) and alternating finite automata (AFA), respectively. FOND4LTL$_f$ offers a more direct comparison, as it converts LTL$_f$ formulas into deterministic finite automata (DFA) and computes the product graph of the DFA and the domain. Although FOND4LTL$_f$ was originally designed for fully observable non-deterministic (FOND) domains, it can also be applied to classical deterministic domains. Finally, Plan4Past is designed for handling temporal goals expressed in Pure-Past Linear Temporal Logic (PPLTL).

We used two classical off-the-shelf planners for final-state goals in our experiments: 
\begin{itemize}
    \item \textbf{A*}, a heuristic search algorithm that guarantees optimal solutions when using admissible heuristics by prioritizing states based on the combined cost of reaching a state and an estimated cost to the goal \cite{hart_formal_1968}.
    \item \textbf{Fast Downward with LAMA search}, a state-of-the-art planner designed to solve planning problems encoded in PDDL \cite{helmert_fast_2006}. LAMA search accelerates planning by identifying \textit{landmarks}, which are critical subgoals that must be satisfied in any valid plan \cite{richter_lama_2010}. It combines a cost-sensitive landmark heuristic with a variant of the FF (Fast Forward) heuristic, which estimates progress toward the goal. LAMA employs an iterated weighted A* search, where weights are progressively decreased to improve solution quality over time. While LAMA often produces high-quality plans efficiently, it does not guarantee optimality. 

    LAMA offers two planning modes:
    \begin{itemize}
        \item \texttt{lama-first}: returns the first plan found without further refinement;
        \item \texttt{seq-sat-lama-2011}: attempts to optimize the first-found plan to make it shorter.
    \end{itemize}
    The \texttt{lama-first} mode is significantly faster than \texttt{seq-sat-lama-2011}. Notably, FOND4LTL$_f$, Exp, Poly, and Plan4Past are extremely slow when coupled with
    \texttt{seq-sat-lama-2011}, and they frequently time out even on simple problems. TIDE, however, performs better with \texttt{seq-sat-lama-2011} because it decomposes the problem into smaller subproblems that are easier to refine. Since this comparison does not focus on plan length, we use the \texttt{lama-first} mode for our experiments.
\end{itemize}

All experiments in this section were executed on a Linux machine running Ubuntu 22.04.5 LTS, equipped with an AMD Ryzen 7 3700X 8-core processor (16 threads, 3.6 GHz base clock). Each experiment was repeated 50 times, and the results were averaged to ensure statistical reliability. A timeout threshold of 30 minutes was applied to the search phase of each problem across all benchmarks and planning methods.

\subsection{Evaluation on ``Easy'' Problems (TB15)}
\label{sec:EvaluationonEasyProblems}

The TB15 problem set, introduced by Torres and Baier \cite{torres_polynomial-time_2015}, was originally designed to compare the performance of the Poly and Exp methods for planning with LTL$_f$ goals. Subsequently, Plan4Past authors extended its use to evaluate their approach for planning with PPLTL goals in comparison to Poly and Exp. This benchmark set includes PDDL problem instances with equivalent LTL$_f$ and PPLTL goals, making it well-suited for a direct comparison of these methods. 

The problem set consists of three classical planning domains that were introduced in past International Planning Competitions (IPC):
\begin{itemize}
    \item \textbf{Blocksworld}: This domain involves manipulating blocks. Using a robotic arm, blocks can be stacked, unstacked, picked up, or put down. The goal is to arrange the blocks according to a specified pattern.
    \item \textbf{Openstacks}: This domain models a resource management problem where production orders must be fulfilled in a specific sequence. ``Stacks'' represent resources that are temporarily used for tasks. A stack is ``opened'' when resources are allocated for a task and ``closed'' when the task is completed, freeing up the resources. For example, a valid plan might involve configuring machines, producing items, and shipping orders, while carefully managing resource availability to ensure all orders are completed in the correct order.
    \item \textbf{Rovers}: This domain simulates a Mars exploration scenario. A set of rovers gathers data about soil, rocks, and images, then communicates this data back to Earth. The goals enforce a specific order over the communication of data.
\end{itemize}

Tables~\ref{table:time_comparison}, \ref{table:time_comparison_openstacks}, and \ref{table:time_comparison_rovers} report the total planning times for each method across the three domains, including both the formula-to-automaton translation time and the search time. In these tables, TIDE is evaluated with both A* and FD-LAMA, while the competing methods (FOND4LTL$_f$, Exp, Poly, and Plan4Past) are tested only with FD-LAMA. This choice was made because FD-LAMA consistently outperformed A* when paired with these methods.

However, to provide a fair comparison, we also report results for Poly and FOND4LTL$_f$ with both A* and FD-LAMA in Tables~\ref{table:planner_comparison}, \ref{table:planner_comparison_openstacks}, and \ref{table:planner_comparison_rovers}.

\begin{table}
\centering
\caption{Planning times (in seconds) for the Blocksworld TB15 benchmark.\\ Bold values indicate the best performance for each problem.}
\label{table:time_comparison}
\begin{tabular}{c|cccccc}
\toprule
Problem & TIDE+A* & TIDE+LAMA  & FOND4LTL$_f$ & Exp & Poly & Plan4Past \\
\midrule
a03 & \textbf{0.009} & 0.195 & 0.656 & 0.516 & 5.475 & 0.397 \\
a04 & \textbf{0.041} & 0.293 & 23.397 & 4.354 & 5.562 & 0.402 \\
a05 & \textbf{0.558} & 0.986 & 74.291 & timeout & 5.731 & 0.410 \\
b03 & \textbf{0.003} & 0.066 & 0.466 & 1.425 & 5.321 & 0.402 \\
b04 & \textbf{0.004} & 0.066 & 0.486 & 42.655 & 5.590 & 0.406 \\
b05 & \textbf{0.004} & 0.065 & 0.405 & failure & 5.665 & 0.409 \\
c03 & \textbf{0.021} & 0.219 & 1.046 & 0.658 & 6.239 & 0.401 \\
c04 & \textbf{0.080} & 0.250 & 28.672 & 31.264 & 7.861 & 0.408 \\
c05 & \textbf{0.425} & 0.628 & 95.435 & timeout & 10.083 & 0.414 \\
d03 & \textbf{0.092} & 0.239 & failure & 0.471 & 5.372 & 0.404 \\
d04 & \textbf{0.147} & 0.332 & failure & 0.482 & 5.452 & 0.416 \\
d05 & \textbf{0.183} & 0.455 & failure & 0.495 & 5.599 & 0.421 \\
e03 & \textbf{0.012} & 0.197 & 0.662 & 0.514 & 5.443 & 0.398 \\
e04 & \textbf{0.024} & 0.284 & 23.411 & 4.383 & 5.558 & 0.412 \\
e05 & \textbf{0.206} & 0.640 & 76.559 & timeout & 5.587 & 0.411 \\
\midrule
Average & \textbf{0.121} & 0.328 & - & - & 6.036 & 0.407 \\
\bottomrule
\end{tabular}
\end{table}

\begin{table}
\centering
\caption{Planning times (in seconds) for the Openstacks TB15 benchmark.\\ Bold values indicate the best performance for each problem.}
\label{table:time_comparison_openstacks}
\begin{tabular}{c|cccccc}
\toprule
Problem & TIDE+A* & TIDE+LAMA  & FOND4LTL$_f$ & Exp & Poly & Plan4Past \\
\midrule
a03 & \textbf{0.043} & 0.698 & 7.821 & 5.019 & 9.826 & 2.571 \\
a04 & \textbf{0.062} & 0.950 & 8.890 & 5.297 & 9.472 & 3.183 \\
a05 & \textbf{0.562} & 1.679 & 16.154 & 11.375 & 12.919 & 4.519 \\
e03 & \textbf{0.045} & 0.698 & 8.882 & 5.019 & 10.034 & 2.567 \\
e04 & \textbf{0.053} & 0.939 & 8.919 & 5.277 & 9.994 & 3.191 \\
e05 & \textbf{0.207} & 1.308 & 16.394 & 11.371 & 10.745 & 4.530 \\
f03 & \textbf{0.042} & 0.697 & 7.848 & 4.222 & 10.195 & 1.557 \\
f05 & \textbf{0.052} & 1.169 & 16.311 & 4.942 & 10.687 & 3.203 \\
g02 & \textbf{0.106} & 1.866 & 9.512 & 5.015 & 9.474 & 4.040 \\
g03 & \textbf{0.066} & 2.188 & 13.362 & 4.943 & 9.559 & 5.156 \\
\midrule
Average & \textbf{0.124} & 1.219 & 11.409 & 6.248 & 10.290 & 3.452 \\
\bottomrule
\end{tabular}
\end{table}

\begin{table}
\centering
\caption{Planning times (in seconds) for the Rovers TB15 benchmark.\\ Bold values indicate the best performance for each problem.}
\label{table:time_comparison_rovers}
\begin{tabular}{c|cccccc}
\toprule
Problem & TIDE+A* & TIDE+LAMA  & FOND4LTL$_f$ & Exp & Poly & Plan4Past \\
\midrule
e03 & 1.306 & \textbf{0.231} & 0.532 & 0.552 & 6.003 & 0.433 \\
e04 & 1.750 & \textbf{0.314} & 17.428 & 0.808 & 6.330 & 0.442 \\
f01 & 0.120 & \textbf{0.075} & 0.415 & 0.470 & 0.735 & 0.401 \\
f02 & 1.433 & \textbf{0.153} & 0.420 & 0.489 & 7.454 & 0.409 \\
f03 & 1.314 & \textbf{0.229} & 0.531 & 0.512 & 5.876 & 0.419 \\
i03 & \textbf{0.030} & 0.075 & 0.418 & 0.537 & 5.628 & 0.439 \\
i04 & \textbf{0.034} & 0.081 & 0.420 & 0.533 & 5.569 & 0.430 \\
\midrule
Average & 0.855 & \textbf{0.165} & 2.881 & 0.557 & 5.371 & 0.425 \\
\bottomrule
\end{tabular}
\end{table}

\begin{table}
\centering
\caption{Planning times (in seconds) for the Blocksworld TB15 benchmark.\\ 
Comparison of TIDE, Poly, and FOND4LTL$_f$ when paired with either the A* or FD-LAMA planner. 
Bold values indicate the best performance for each problem.}
\label{table:planner_comparison}
\begin{tabular}{c|ccc|ccc}
\toprule
Problem & \multicolumn{3}{c|}{A* planner} & \multicolumn{3}{c}{FD-LAMA planner} \\
 & TIDE & Poly & FOND4LTL$_f$ & TIDE & Poly & FOND4LTL$_f$ \\
\midrule
a03 & \textbf{0.009} & 2.842 & 25.391 & 0.195 & 5.475 & 0.656 \\
a04 & \textbf{0.041} & 19.664 & timeout & 0.293 & 5.562 & 23.397 \\
a05 & \textbf{0.558} & 101.409 & timeout & 0.986 & 5.731 & 74.291 \\
b03 & \textbf{0.003} & 0.749 & 0.844 & 0.066 & 5.321 & 0.466 \\
b04 & \textbf{0.004} & 1.477 & 3.872 & 0.066 & 5.590 & 0.486 \\
b05 & \textbf{0.004} & 2.651 & 3.820 & 0.065 & 5.665 & 0.405 \\
c03 & \textbf{0.021} & 10.701 & 1508.380 & 0.219 & 6.239 & 1.046 \\
c04 & \textbf{0.080} & 36.182 & timeout & 0.250 & 7.861 & 28.672 \\
c05 & \textbf{0.425} & 200.199 & timeout & 0.628 & 10.083 & 95.435 \\
d03 & \textbf{0.092} & 1.298 & failure & 0.239 & 5.372 & failure \\
d04 & \textbf{0.147} & 7.937 & failure & 0.332 & 5.452 & failure \\
d05 & \textbf{0.183} & 37.340 & failure & 0.455 & 5.599 & failure \\
e03 & \textbf{0.012} & 4.955 & 60.065 & 0.197 & 5.443 & 0.662 \\
e04 & \textbf{0.024} & 27.608 & timeout & 0.284 & 5.558 & 23.411 \\
e05 & \textbf{0.206} & 119.635 & timeout & 0.640 & 5.587 & 76.559 \\
\midrule
Average & \textbf{0.121} & 38.310 & - & 0.328 & 6.036 & - \\
\bottomrule
\end{tabular}
\end{table}

\begin{table}
\centering
\caption{Planning times (in seconds) for the Openstacks TB15 benchmark.\\ 
Comparison of TIDE, Poly, and FOND4LTL$_f$ when paired with either the A* or FD-LAMA planner. 
Bold values indicate the best performance for each problem.}
\label{table:planner_comparison_openstacks}
\begin{tabular}{c|ccc|ccc}
\toprule
Problem & \multicolumn{3}{c|}{A* planner} & \multicolumn{3}{c}{FD-LAMA planner} \\
 & TIDE & Poly & FOND4LTL$_f$ & TIDE & Poly & FOND4LTL$_f$ \\
\midrule
a03 & \textbf{0.043} & 6.446 & 0.599 & 0.698 & 9.826 & 7.821 \\
a04 & \textbf{0.062} & 13.768 & 0.695 & 0.950 & 9.472 & 8.890 \\
a05 & \textbf{0.562} & 30.274 & 0.930 & 1.679 & 12.919 & 16.154 \\
e03 & \textbf{0.045} & 7.253 & 0.603 & 0.698 & 10.034 & 8.882 \\
e04 & \textbf{0.053} & 13.876 & 0.691 & 0.939 & 9.994 & 8.919 \\
e05 & \textbf{0.207} & 27.996 & 0.916 & 1.308 & 10.745 & 16.394 \\
f03 & \textbf{0.042} & 5.097 & 0.602 & 0.697 & 10.195 & 7.848 \\
f05 & \textbf{0.052} & 24.188 & 0.908 & 1.169 & 10.687 & 16.311 \\
g02 & \textbf{0.106} & 8.461 & 0.917 & 1.866 & 9.474 & 9.512 \\
g03 & \textbf{0.066} & 10.479 & 1.133 & 2.188 & 9.559 & 13.362 \\
\midrule
Average & \textbf{0.124} & 14.784 & 0.799 & 1.219 & 10.290 & 11.409 \\
\bottomrule
\end{tabular}
\end{table}

\begin{table}
\centering
\caption{Planning times (in seconds) for the Rovers TB15 benchmark.\\ 
Comparison of TIDE, Poly, and FOND4LTL$_f$ when paired with either the A* or FD-LAMA planner. 
Bold values indicate the best performance for each problem.}
\label{table:planner_comparison_rovers}
\begin{tabular}{c|ccc|ccc}
\toprule
Problem & \multicolumn{3}{c|}{A* planner} & \multicolumn{3}{c}{FD-LAMA planner} \\
 & TIDE & Poly & FOND4LTL$_f$ & TIDE & Poly & FOND4LTL$_f$ \\
\midrule
e03 & 1.306 & timeout & 107.69 & \textbf{0.231} & 6.003 & 0.532 \\
e04 & 1.750 & timeout & timeout & \textbf{0.314} & 6.330 & 17.428 \\
f01 & 0.120 & 1.375 & 0.423 & \textbf{0.075} & 0.735 & 0.415 \\
f02 & 1.433 & 491.044 & 10.788 & \textbf{0.153} & 7.454 & 0.420 \\
f03 & 1.314 & timeout & 108.346 & \textbf{0.229} & 5.876 & 0.531 \\
i03 & \textbf{0.030} & 2.749 & 0.357 & 0.075 & 5.628 & 0.418 \\
i04 & \textbf{0.034} & 1.375 & 0.351 & 0.081 & 5.569 & 0.420 \\
\midrule
Average & 0.855 & - & - & \textbf{0.165} & 5.371 & 2.881 \\
\bottomrule
\end{tabular}
\end{table}

\paragraph{Discussion}
The experimental results highlight the consistent performance advantages of TIDE. Across the TB15 problem set in three domains (Blocksworld, Openstacks, and Rovers), TIDE combined with A* demonstrated superior planning times in the Blocksworld and Openstacks domains. Meanwhile, when combined with the Fast Downward planner using LAMA search (denoted as FD-LAMA), TIDE exhibited a distinct advantage in the Rovers domain.

To evaluate TIDE's performance, we compared it against competing methods combined with FD-LAMA in the \texttt{lama-first} mode, which returns the first valid plan without further optimization. Additionally, we tested Poly and FOND4LTL$_f$ with the A* planner. However, we were unable to integrate A* with Exp or Plan4Past because these planners reformulate the problem into a final-state goal PDDL representation that the A* planner (implemented in the \texttt{pddlboat} library) failed to parse. Even when used with Poly and FOND4LTL$_f$, A* underperformed, whereas our method excelled. This disparity is likely due to the lack of an effective directional heuristic in Poly and FOND4LTL$_f$.

Overall, the results indicate that TIDE’s trace-guided approach, whether combined with A* or Fast Downward, is both versatile and effective for planning with LTL$_f$ goals. While Plan4Past performed reasonably well in comparison to FOND4LTL$_f$, Exp, and Poly, it was generally outperformed by TIDE. Notably, Plan4Past exhibited significantly slower performance in the Openstacks domain.

\subsection{Scaling Experiments in the Blocksworld Domain}
\label{sec:EvaluationonMyProblems}

This subsection evaluates the performance of TIDE, FOND4LTL$_f$, Exp, Poly, and Plan4Past on two custom scaling benchmarks in the Blocksworld domain. Both benchmarks involve stacking a tower of \( n \) blocks according to specified LTL$_f$ goals, with \( n \) ranging from 3 to 25. These benchmarks are designed to evaluate the scalability of planning methods and their ability to handle increasingly complex subgoals.

\begin{figure}[h]
    \centering
    \begin{minipage}{0.48\linewidth}
        \centering
        \includegraphics[width=\linewidth]{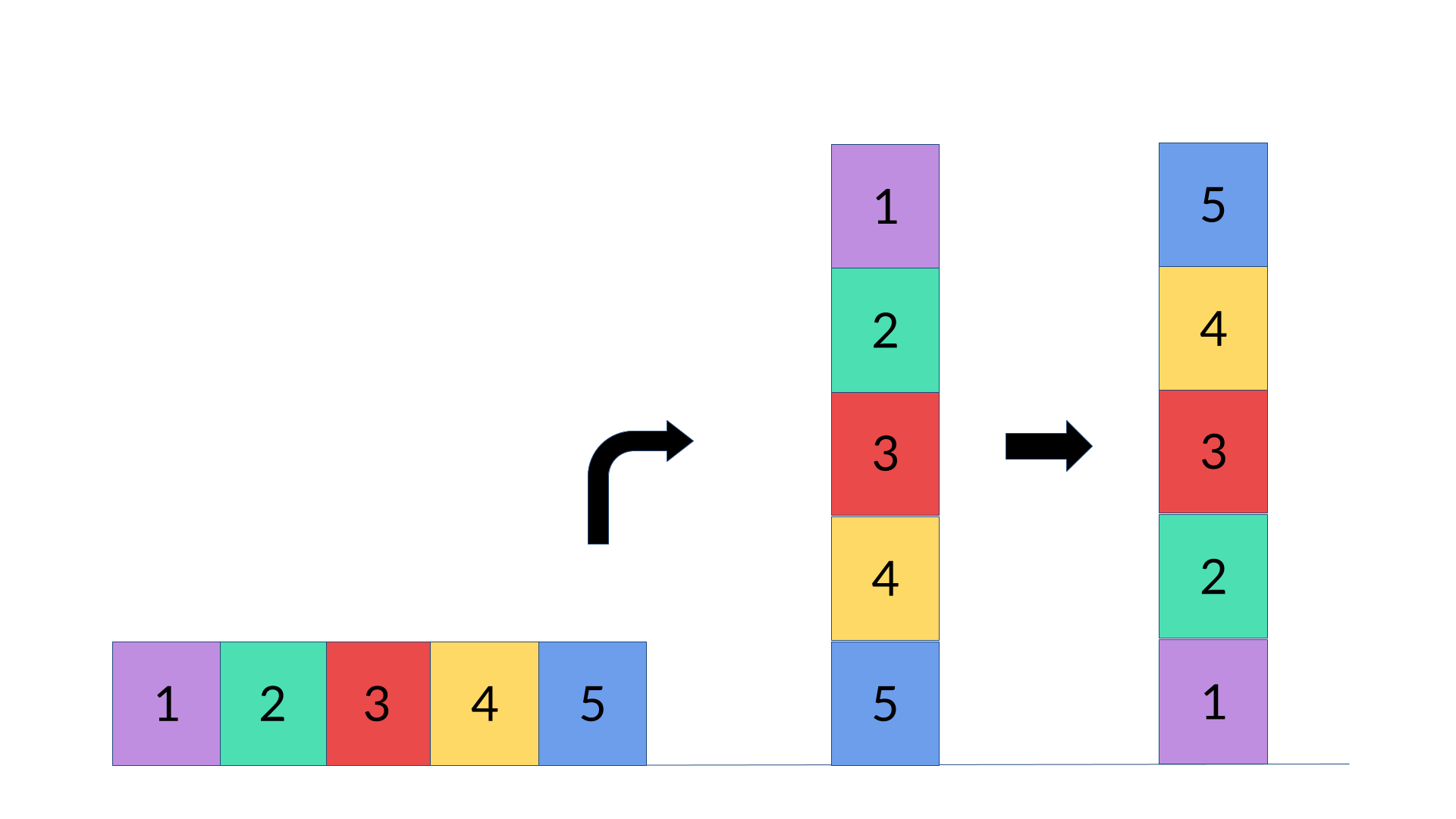}
        \caption*{(a) Benchmark 1: Tower Reversal}
        \label{fig:goal_b1}
    \end{minipage}%
    \hspace{0.02\linewidth}
    \begin{minipage}{0.48\linewidth}
        \centering
        \includegraphics[width=\linewidth]{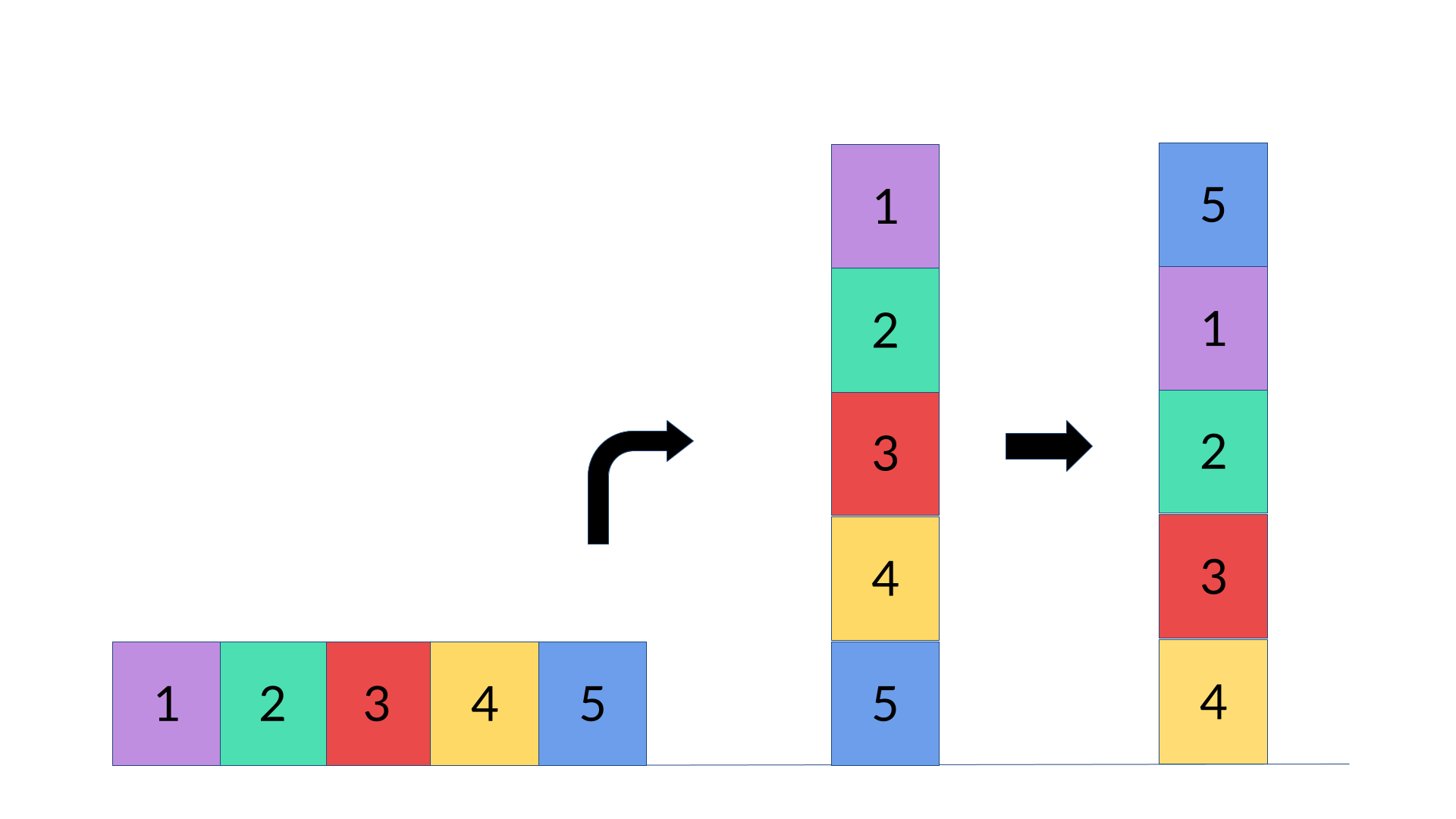}
        \caption*{(b) Benchmark 2: Base Block Relocation}
        \label{fig:goal_b2}
    \end{minipage}
    \caption{Visual goals for the two Blocksworld benchmarks with \( n = 5\) blocks.\\
    (a) The goal for Benchmark~1 is to stack a tower in the ascending order and then reverse it.\\
    (b) The goal for Benchmark~2 is to stack a tower in the ascending order and then move the base block to the top while preserving the order of the other blocks.}
    \label{fig:blocksworld_goals}
\end{figure}

Figure~\ref{fig:blocksworld_goals} illustrates the goal specifications for both benchmarks when \( n = 5 \). In Benchmark~1, the goal is to reverse a tower initially stacked in ascending order. In Benchmark~2, the goal is to reposition the original base block to the top of the tower, while preserving the order of the other blocks.

\paragraph{Benchmark 1: Tower Reversal}
The first benchmark requires stacking a tower of \( n \) blocks in ascending order and then reversing the tower's order. This task is defined using the following LTL$_f$ goal:
\[
F(\text{tower\_ascending\_order} \land X(F(\text{tower\_descending\_order}))),
\]
where:
\begin{itemize}
    \item \(\text{tower\_ascending\_order}\): ensures that the blocks are stacked in ascending order, and
    \item \(\text{tower\_descending\_order}\): requires the stack to be reordered in descending order.
\end{itemize}

Although this goal is relatively simple, the complexity of solving the subgoals grows exponentially as the number of blocks increases.

\paragraph{Benchmark 2: Base Block Relocation}
The second benchmark introduces a more complex subgoal. After stacking the tower in ascending order, the ``lowest'' base block of the initial tower must be moved to the top of the tower, while the order of all other blocks remains unchanged. This task is defined using the following LTL$_f$ goal:
\[
F(\text{tower\_ascending\_order} \land X(F(\text{tower\_ascending\_order\_with\_base\_block\_on\_top}))),
\]
where:
\begin{itemize}
    \item \(\text{tower\_ascending\_order}\): ensures that the blocks are stacked in ascending order, and
    \item \(\text{tower\_ascending\_order\_with\_base\_block\_on\_top}\): requires that the tower is re-stacked in ascending order, except the original base block must now be placed at the top.
\end{itemize}

\paragraph{Known Theoretical Plan Length}
The shortest possible theoretical plan length for both benchmarks can be derived based on the required actions in the Blocksworld domain. The domain includes four actions: \texttt{pick-up}, \texttt{stack}, \texttt{unstack}, and \texttt{put-down}. Below, we derive the theoretical minimum plan lengths for each benchmark.

\noindent
{\bf Benchmark 1: Tower Reversal}
The goal for Benchmark 1 is to stack a tower of \( n \) blocks in ascending order and then reverse the order of the tower. The plan length is derived as follows:

\paragraph{Step 1: Stack the tower in ascending order}
\begin{itemize}
    \item The first block (base) is already on the table and does not require any action. 
    \item Each of the remaining \( n-1 \) blocks requires two actions: \texttt{pick-up} and \texttt{stack}.
\end{itemize}
The total number of actions for this step is: \(2(n-1).\)

\paragraph{Step 2: Reverse the tower}
\begin{itemize}
    \item Each block requires two actions to reverse its position: one to remove it from the current position and one to place it in the reversed order.
\end{itemize}
The total number of actions for this step is: \(2n\).

\paragraph{Total Theoretical Plan Length for Benchmark 1}
\[
\text{Plan Length} = 2(n-1) + 2n = 4n - 2.
\]

\noindent
{\bf Benchmark 2: Base Block Relocation}
The goal for Benchmark 2 is to stack a tower of \( n \) blocks in ascending order and then relocate the base block to the top, while preserving the relative order of the other blocks. The plan length is derived as follows:

\paragraph{Step 1: Stack the tower in ascending order}  \(2(n-1)\) actions (as before).

\paragraph{Step 2: Unstack all blocks above the base block} \(2(n-1)\) actions.\\
To relocate the base block to the top, all \( n-1 \) blocks above the base block must be unstacked and placed on the table. Each block requires two actions: \texttt{unstack} and \texttt{put-down}. 

\paragraph{Step 3: Stack the tower in the new order} \(2(n-1)\) actions.\\
With the base block already on the table, the remaining \( n-1 \) blocks must be stacked in the new order. Each block requires two actions: \texttt{pick-up} and \texttt{stack}. 

\paragraph{Total Theoretical Plan Length for Benchmark 2}
\[
\text{Plan Length} = 2(n-1) + 2(n-1) + 2(n-1) = 6(n-1).
\]

These theoretical plan lengths provide a baseline to evaluate the performance of planning methods and their ability to produce solution plans of reasonable lengths.

% Insert here

\newpage
\begin{figure}
    \centering
    \includegraphics[width=0.805\textwidth]{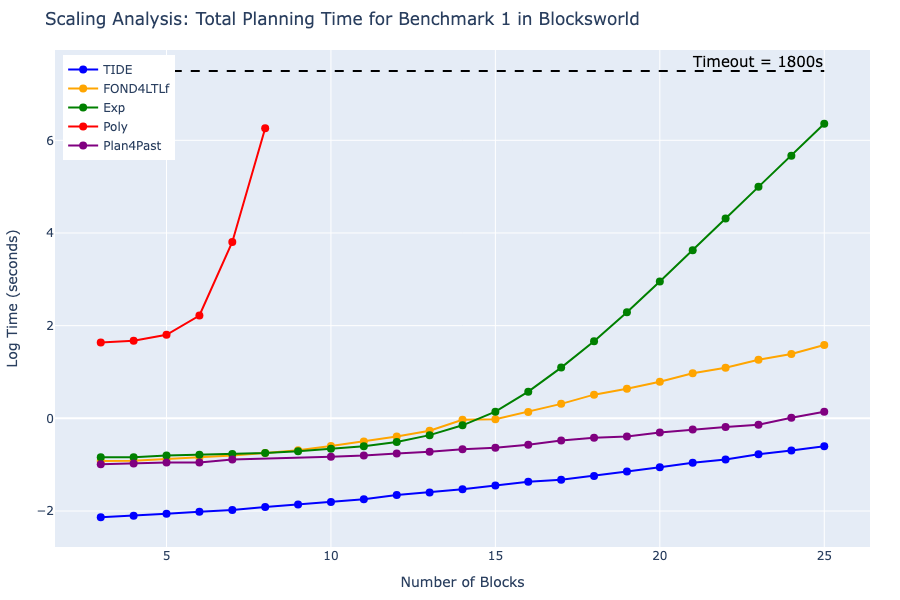}
    \caption{Scalability of total planning times for Benchmark 1 (Tower Reversal). The x-axis represents the number of blocks in the problem (\( n \)), and the y-axis shows total planning time—comprising both formula-to-automaton translation and planner search—on a logarithmic scale.}
    \label{fig:scalability_b1}
\end{figure}

\begin{figure}
    \centering
    \includegraphics[width=0.805\textwidth]{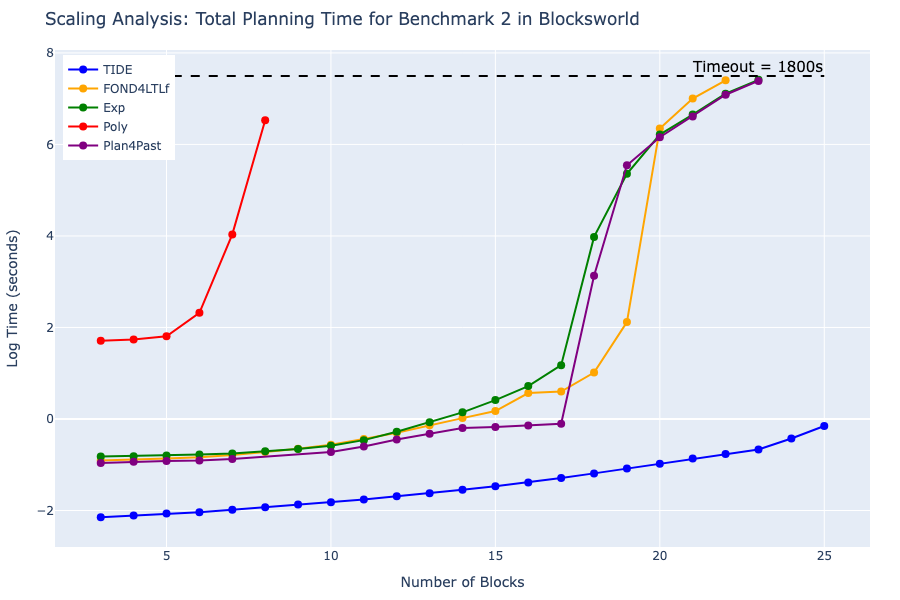}
    \caption{Scalability of total planning times for Benchmark 2 (Base Block Relocation). The x-axis represents the number of blocks in the problem (\( n \)), and the y-axis shows total planning time—comprising both formula-to-automaton translation and planner search—on a logarithmic scale.}
    \label{fig:scalability_b2}
\end{figure}

\FloatBarrier

\newpage
\begin{table}
\centering
\caption{Planning times (in seconds) for Blocksworld Benchmark 1: Tower Reversal.\\
Each problem involves stacking a tower of $i$ blocks and then reversing it,\\
where $i$ is the problem number. Bold values indicate the best performance.}
\label{table:time_comparison_blocksworld_new_b1}
\begin{tabular}{c|ccccc}
\toprule
Problem & TIDE  & FOND4LTL$_f$ & Exp & Poly & Plan4Past \\
\midrule
     t3 &    \textbf{0.119} & 0.398 & 0.431 &  5.142 &  0.371 \\
     t4 &  \textbf{0.122} &  0.400 & 0.433 &  5.340 &  0.378 \\
     t5 &  \textbf{0.128} &  0.416 & 0.448 &  6.074 &  0.386 \\
     t6 &  \textbf{0.133} &  0.430 & 0.457 &  9.202 &  0.386 \\
     t7 &   \textbf{0.139} &  0.450 & 0.464 &  45.000 &  0.411 \\
     t8 &  \textbf{0.148} & 0.474 & 0.473 &  523.02 &   failure \\
     t9 &    \textbf{0.156} & 0.504 & 0.491 &  timeout &  failure \\
     t10 &  \textbf{0.164} & 0.550 & 0.519 &  timeout &  0.436 \\
     t11 &  \textbf{0.174} & 0.605 & 0.547 &  timeout &  0.448 \\
     t12 &  \textbf{0.191} & 0.674 & 0.602 &  timeout &  0.470 \\
     t13 &   \textbf{0.202} & 0.766 & 0.696 &  timeout &  0.486 \\
     t14 &  \textbf{0.216} & 0.969 & 0.861 &  timeout &   0.513 \\
     t15 &  \textbf{0.234} & 0.979 & 1.155 &  timeout &  0.531 \\
     t16 &  \textbf{0.255} & 1.150 & 1.778 &  timeout &  0.566 \\
     t17 &  \textbf{0.266} & 1.367 & 2.989 &  timeout &  0.620 \\
     t18 &   \textbf{0.288} & 1.668 & 5.286 &  timeout &  0.657 \\
     t19 &  \textbf{0.318} & 1.892 & 9.851 &  timeout &   0.676 \\
     t20 &  \textbf{0.348} & 2.205 & 19.188 &  timeout &  0.737 \\
     t21 &  \textbf{0.383} & 2.642 & 37.581 &  timeout &  0.788 \\
     t22 &   \textbf{0.410} & 2.975  & 74.634 & timeout &  0.831 \\
     t23 &  \textbf{0.462} & 3.541 & 147.74 &  timeout &   0.873 \\
     t24 &  \textbf{0.499} & 4.006 & 289.12 &  timeout &  1.011 \\
     t25 &  \textbf{0.548} & 4.875 & 576.08 &  timeout &  1.151 \\
\bottomrule
\end{tabular}
\end{table}

\begin{table}
\centering
\caption{Comparison of Plan Length for Blocksworld Benchmark 1: Tower Reversal.\\
Each problem involves stacking a tower of $i$ blocks and then reversing it,\\
where $i$ is the problem number. Bold values indicate the shortest plan.}
\label{table:length_comparison_blocksworld_new_b1}
\begin{tabular}{c|c|ccccc}
\toprule
Problem & Optimal & TIDE & FOND4LTL$_f$ & Exp & Poly & Plan4Past \\
\midrule
     t3 & 10 & \textbf{10} & \textbf{10} & \textbf{10} &  \textbf{10} &  \textbf{10} \\
     t4 & 14 & \textbf{14} & \textbf{14} & 20 &  \textbf{14} &  20 \\
     t5 & 18 & 22 & \textbf{18} & 28 &  \textbf{18} &  28 \\
     t6 & 22 & 26 & 32 & 30 &  \textbf{22} &  30 \\
     t7 & 26 & 30 & 40 & 54 &  \textbf{26} &  58 \\
     t8 & 30 & 34 & 60 & 64 &  \textbf{30} &   failure \\
     t9 & 34 & \textbf{38} & 52 & 70 &  timeout &  failure \\
     t10 & 38 & \textbf{42} & 82 & 64 &  timeout &  78 \\
\bottomrule
\end{tabular}
\end{table}

\FloatBarrier

\begin{table}
\centering
\caption{Planning times (in seconds) for Benchmark 2: Base Block Relocation.\\
Each problem requires stacking a tower of $i$ blocks in ascending order and relocating the base block to the top. Bold values indicate the best performance.}
\label{table:time_comparison_blocksworld_new_b2}
\begin{tabular}{c|ccccc}
\toprule
Problem & TIDE  & FOND4LTL$_f$ & Exp & Poly & Plan4Past \\
\midrule
m3  & \textbf{0.117} & 0.403 & 0.439 & 5.525 & 0.383 \\
m4  & \textbf{0.121} & 0.410 & 0.444 & 5.679 & 0.393 \\
m5  & \textbf{0.127} & 0.421 & 0.451 & 6.084 & 0.399 \\
m6  & \textbf{0.130} & 0.434 & 0.460 & 10.181 & 0.404 \\
m7  & \textbf{0.138} & 0.456 & 0.470 & 56.501 & 0.417 \\
m8  & \textbf{0.145} & 0.486 & 0.493 & 683.184 & failure \\
m9  & \textbf{0.154} & 0.521 & 0.519 & timeout & failure \\
m10 & \textbf{0.162} & 0.571 & 0.556 & timeout & 0.487 \\
m11 & \textbf{0.172} & 0.645 & 0.630 & timeout & 0.548 \\
m12 & \textbf{0.185} & 0.742 & 0.757 & timeout & 0.639 \\
m13 & \textbf{0.198} & 0.874 & 0.929 & timeout & 0.725 \\
m14 & \textbf{0.213} & 1.019 & 1.156 & timeout & 0.819 \\
m15 & \textbf{0.230} & 1.189 & 1.512 & timeout & 0.839 \\
m16 & \textbf{0.250} & 1.762 & 2.053 & timeout & 0.867 \\
m17 & \textbf{0.276} & 1.822 & 3.241 & timeout & 0.902 \\
m18 & \textbf{0.305} & 2.760 & 53.486 & timeout & 22.838 \\
m19 & \textbf{0.338} & 8.318 & 211.998 & timeout & 255.880 \\
m20 & \textbf{0.376} & 572.370 & 500.787 & timeout & 471.694 \\
m21 & \textbf{0.421} & 1098.469 & 775.433 & timeout & 743.749 \\
m22 & \textbf{0.466} & 1636.141 & 1223.995 & timeout & 1194.657 \\
m23 & \textbf{0.514} & timeout & 1654.612 & timeout & 1617.297 \\
m24 & \textbf{0.653} & timeout & timeout & timeout & timeout \\
m25 & \textbf{0.857} & timeout & timeout & timeout & timeout \\
\bottomrule
\end{tabular}
\end{table}

\begin{table}
\centering
\caption{Comparison of Plan Length for Benchmark 2: Base Block Relocation.\\
Each problem requires stacking a tower of $i$ blocks in ascending order and relocating the base block to the top. Bold values indicate the shortest plan.}
\label{table:length_comparison_blocksworld_new_b2}
\begin{tabular}{c|c|ccccc}
\toprule
Problem & Optimal & TIDE  & FOND4LTL$_f$ & Exp & Poly & Plan4Past \\
\midrule
     m3 &    12 & \textbf{12} & \textbf{12} & \textbf{12} &  \textbf{12} &  \textbf{12} \\
     m4 &  18 & 22 & 22 & 26 &  \textbf{18} &  22 \\
     m5 &  24 & 40 & 52 & 48 &  \textbf{24} &   52 \\
     m6 &  30 & 46 & 66 & 78 &  \textbf{30} &  66 \\
     m7 &  36 & 52 & 76 & 76 &  \textbf{36} &  92 \\
     m8 &  42 & 58 & 110 & 150 & \textbf{42} &   failure \\
     m9 &  48 & \textbf{64} & 140 & 180 &  timeout &  failure \\
     m10 &  54 & \textbf{70} & 158 & 158 &  timeout &  158 \\
\bottomrule
\end{tabular}
\end{table}

\FloatBarrier

\paragraph{Results and Discussion}
The experimental results, summarized in Tables~\ref{table:time_comparison_blocksworld_new_b1} and \ref{table:length_comparison_blocksworld_new_b1} for Benchmark 1, and Tables~\ref{table:time_comparison_blocksworld_new_b2} and \ref{table:length_comparison_blocksworld_new_b2} for Benchmark 2, provide insights into the trade-offs between computational efficiency and plan quality for each method. The results for planning times and plan lengths are analyzed separately for the two benchmarks.

The planning time results highlight the scalability advantages of TIDE across both benchmarks, consistently solving all problems up to \( n = 25 \) without timing out, while other methods exhibit varying degrees of inefficiency.

\noindent
{\bf Planning Time Analysis}

The total planning times—comprising both translation and search—for Benchmark 1 (Tower Reversal) and Benchmark 2 (Base Block Relocation) are shown in Tables~\ref{table:time_comparison_blocksworld_new_b1} and~\ref{table:time_comparison_blocksworld_new_b2}. Key observations include:

\begin{itemize}
    \item \textbf{TIDE}: Demonstrates superior performance across both benchmarks, solving all problems up to \( n = 25 \) with the shortest planning times. 
    \item \textbf{Plan4Past}: Achieves strong performance, particularly for smaller problem sizes, but becomes slower than TIDE as problem complexity increases, especially in Benchmark 2.
    \item \textbf{FOND4LTL$_f$}: Shows moderate scalability but is consistently slower than TIDE and Plan4Past. It handles Benchmark 1 better but struggles with the additional unstacking and restacking steps in Benchmark 2, timing out for problems with \( n > 22 \).
    \item \textbf{Exp}: Performs better than Poly but fails to scale, timing out for \( n > 23 \) in Benchmark 2.
    \item \textbf{Poly}: Struggles with scalability in both benchmarks, timing out for \( n > 8\). 
\end{itemize}

The scalability plots (Figures~\ref{fig:scalability_b1} and~\ref{fig:scalability_b2}) clearly illustrate the differences in performance, particularly the dominance of TIDE and Plan4Past in solving larger problem instances more efficiently.

\noindent
{\bf Plan Length Analysis}

The plan lengths for each method, shown in Table~\ref{table:length_comparison_blocksworld_new_b1} and Table~\ref{table:length_comparison_blocksworld_new_b2}, provide additional insights into the strengths and weaknesses of the evaluated methods:
\begin{itemize}
    \item \textbf{Poly}: Consistently produces the shortest plans, matching the theoretical minimum of \( 4n - 2 \) for Benchmark 1 and \( 6(n - 1) \) for Benchmark 2. However, its computational cost makes it impractical for larger problems.
    \item \textbf{FOND4LTL$_f$}, \textbf{Exp}, and \textbf{Plan4Past}: Generate significantly longer plans with many unnecessary actions. For example, in problem \( t7 \) (\( n = 7 \)), Poly’s plan length is more than twice as short as those produced by Exp and Plan4Past.
    \item \textbf{TIDE}: Balances plan length and planning time, producing plans much closer to the theoretical minimum than those of FOND4LTL$_f$, Exp and Plan4Past, while maintaining computational efficiency.
\end{itemize}

This balance makes TIDE particularly effective for practical scenarios where both reasonable plan lengths and computational efficiency are required.

\subsection{Breakdown of Planning Times: Translation vs. Search}
\label{sec:BreakdownTranslationSearch}

To gain deeper insights into the behavior of each method, we analyze the total planning time by separating it into two distinct phases:
\begin{enumerate} \item \textbf{Translation Phase}: includes the time required to convert the LTL$_f$ goal into a finite automaton; \item \textbf{Search Phase}: includes the time spent solving the resulting final-state planning problem using an off-the-shelf planner (A* or FD-LAMA). \end{enumerate}

The breakdown results are shown in Table~\ref{table:time_breakdown} for the TB15 Blocksworld benchmark and in Table~\ref{table:time_breakdown_blocksworld_b2} for the custom Benchmark~2 (Base Block Relocation).

\begin{table}
\centering
\small
\caption{Breakdown of planning times (in seconds) for the Blocksworld TB15 benchmark. Each cell reports the time spent in the translation and search phases separately. All methods are paired with the FD-LAMA off-the-shelf planner.}
\label{table:time_breakdown}
\begin{tabular}{c|cc|cc|cc|cc|cc}
\toprule
Problem
& \multicolumn{2}{c|}{\textbf{TIDE}} 
& \multicolumn{2}{c|}{\textbf{FOND4LTL$_f$}} 
& \multicolumn{2}{c|}{\textbf{Exp}} 
& \multicolumn{2}{c|}{\textbf{Poly}} 
& \multicolumn{2}{c}{\textbf{Plan4Past}} \\
& Trans. & Search & Trans. & Search & Trans. & Search & Trans. & Search & Trans. & Search \\
\midrule
a03 & 0.003 & 0.192 & 0.308 & 0.347 & 0.364 & 0.151 & 0.215 & 5.259 & 0.286 & 0.111 \\
a04 & 0.024 & 0.269 & 0.313 & 23.083 & 0.489 & 3.864 & 0.220 & 5.342 & 0.286 & 0.116 \\
a05 & 0.543 & 0.444 & 0.340 & 73.951 & 2.507 & timeout & 0.220 & 5.510 & 0.289 & 0.121 \\
b03 & 0.003 & 0.062 & 0.356 & 0.110 & 1.324 & 0.102 & 0.216 & 5.105 & 0.289 & 0.113 \\
b04 & 0.003 & 0.063 & 0.373 & 0.114 & 42.554 & 0.101 & 0.219 & 5.371 & 0.288 & 0.118 \\
b05 & 0.003 & 0.062 & 0.306 & 0.099 & 56.669 & failure & 0.220 & 5.445 & 0.287 & 0.123 \\
c03 & 0.005 & 0.214 & 0.308 & 0.738 & 0.369 & 0.289 & 0.216 & 6.023 & 0.288 & 0.113 \\
c04 & 0.015 & 0.235 & 0.313 & 28.360 & 0.518 & 30.746 & 0.218 & 7.643 & 0.290 & 0.118 \\
c05 & 0.238 & 0.390 & 0.335 & 95.101 & 2.925 & timeout & 0.221 & 9.862 & 0.288 & 0.126 \\
d03 & 0.002 & 0.237 & 0.309 & failure & 0.349 & 0.122 & 0.213 & 5.159 & 0.288 & 0.116 \\
d04 & 0.006 & 0.326 & 0.322 & failure & 0.348 & 0.134 & 0.217 & 5.234 & 0.292 & 0.124 \\
d05 & 0.012 & 0.443 & 0.330 & failure & 0.350 & 0.146 & 0.220 & 5.379 & 0.291 & 0.130 \\
e03 & 0.004 & 0.193 & 0.309 & 0.354 & 0.364 & 0.151 & 0.215 & 5.228 & 0.287 & 0.110 \\
e04 & 0.012 & 0.272 & 0.311 & 23.100 & 0.489 & 3.894 & 0.218 & 5.340 & 0.294 & 0.118 \\
e05 & 0.176 & 0.464 & 0.325 & 76.234 & 2.555 & timeout & 0.220 & 5.367 & 0.289 & 0.122 \\
\midrule
Average & 0.070 & 0.258 & 0.324 & - & 7.478 & - & 0.218 & 5.818 & 0.289 & 0.119 \\
\bottomrule
\end{tabular}
\end{table}

\begin{table}
\centering
\small
\caption{Breakdown of planning times (in seconds) for Benchmark 2: Base Block Relocation. 
Each cell reports the time spent in the translation and search phases separately. All methods are paired with the FD-LAMA off-the-shelf planner.\\
The benchmark requires stacking a tower of $i$ blocks in ascending order and relocating the base block to the top.}
\label{table:time_breakdown_blocksworld_b2}
\begin{tabular}{c|cc|cc|cc|cc|cc}
\toprule
Problem 
& \multicolumn{2}{c|}{\textbf{TIDE}} 
& \multicolumn{2}{c|}{\textbf{FOND4LTL$_f$}} 
& \multicolumn{2}{c|}{\textbf{Exp}} 
& \multicolumn{2}{c|}{\textbf{Poly}} 
& \multicolumn{2}{c}{\textbf{Plan4Past}} \\
& Trans. & Search & Trans. & Search & Trans. & Search & Trans. & Search & Trans. & Search \\
\midrule
m3  & 0.002 & 0.115 & 0.309 & 0.094 & 0.346 & 0.093 & 0.212 & 5.312 & 0.288 & 0.095 \\
m4  & 0.002 & 0.119 & 0.309 & 0.101 & 0.346 & 0.098 & 0.216 & 5.463 & 0.292 & 0.101 \\
m5  & 0.002 & 0.125 & 0.309 & 0.111 & 0.347 & 0.104 & 0.220 & 5.864 & 0.293 & 0.107 \\
m6  & 0.002 & 0.129 & 0.309 & 0.125 & 0.349 & 0.111 & 0.224 & 9.957 & 0.290 & 0.115 \\
m7  & 0.002 & 0.136 & 0.312 & 0.144 & 0.350 & 0.120 & 0.222 & 56.279 & 0.291 & 0.126 \\
m8  & 0.002 & 0.143 & 0.311 & 0.175 & 0.354 & 0.139 & 0.227 & 682.96 & 0.202 & - \\
m9  & 0.002 & 0.152 & 0.314 & 0.207 & 0.360 & 0.159 & 0.226 & - & 0.203 & - \\
m10 & 0.002 & 0.160 & 0.313 & 0.258 & 0.369 & 0.187 & 0.229 & - & 0.291 & 0.195 \\
m11 & 0.002 & 0.172 & 0.312 & 0.333 & 0.385 & 0.245 & 0.234 & - & 0.294 & 0.254 \\
m12 & 0.002 & 0.185 & 0.312 & 0.429 & 0.424 & 0.333 & 0.236 & - & 0.294 & 0.345 \\
m13 & 0.002 & 0.196 & 0.313 & 0.561 & 0.494 & 0.435 & 0.241 & - & 0.293 & 0.432 \\
m14 & 0.002 & 0.211 & 0.313 & 0.706 & 0.638 & 0.517 & 0.248 & - & 0.295 & 0.525 \\
m15 & 0.002 & 0.228 & 0.311 & 0.878 & 0.927 & 0.585 & 0.248 & - & 0.294 & 0.544 \\
m16 & 0.002 & 0.248 & 0.312 & 1.450 & 1.501 & 0.552 & 0.254 & - & 0.294 & 0.573 \\
m17 & 0.002 & 0.274 & 0.313 & 1.509 & 2.651 & 0.590 & 0.260 & - & 0.293 & 0.609 \\
m18 & 0.002 & 0.303 & 0.314 & 2.446 & 4.925 & 48.561 & 0.271 & - & 0.294 & 22.544 \\
m19 & 0.002 & 0.336 & 0.314 & 8.004 & 9.436 & 202.56 & 0.272 & - & 0.303 & 255.58 \\
m20 & 0.002 & 0.374 & 0.312 & 572.05 & 18.634 & 482.15 & 0.278 & - & 0.297 & 471.40 \\
m21 & 0.002 & 0.419 & 0.310 & 1098.2 & 37.310 & 738.12 & 0.283 & - & 0.304 & 743.44 \\
m22 & 0.002 & 0.464 & 0.311 & 1635.8 & 73.265 & 1150.7 & 0.298 & - & 0.307 & 1194.4 \\
m23 & 0.002 & 0.512 & 0.310 & - & 146.82 & 1507.8 & 0.306 & - & 0.297 & 1617.0 \\
m24 & 0.002 & 0.653 & 0.311 & - & 287.56 & - & 0.321 & - & 0.297 & - \\
m25 & 0.002 & 0.857 & 0.312 & - & 575.64 & - & 0.320 & - & 0.295 & - \\
\bottomrule
\end{tabular}
\end{table}

\newpage
This breakdown shows trade-offs between translation and search across methods:
\begin{itemize}
    \item \textbf{TIDE}: Translation times are consistently negligible, as TIDE performs a straightforward LTL$_f$-to-DFA conversion without any costly post-processing, such as modifying domain and problem files or introducing new predicates. The translation relies on the highly efficient Spot library, implemented in C++17 \cite{artho_spot_2016, shoham_spot_2022}. The search phase dominates the total runtime but scales well due to the simplified structure of the subproblems. Even in Benchmark~2, where the problems are more challenging, TIDE completes all instances well below the 30-minute timeout.

    \item \textbf{FOND4LTL$_f$}: Translation is noticeably more expensive—around 0.3 seconds—but remains relatively constant as problem complexity increases. This fixed overhead suggests that converting to a DFA can be a viable strategy for real-world problems. However, search time grows rapidly with problem size, leading to consistent timeouts in Benchmark~2 beyond \( n = 22 \). The method’s scalability is limited by the size of the product graph.

    \item \textbf{Exp}: Exp incurs the highest translation cost among all methods. It converts LTL$_f$ formulas into NFAs and applies additional post-processing, resulting in translation times that grow exponentially in Benchmark~2. Its search phase is also relatively slow and consistently times out for \( n > 23 \).

    \item \textbf{Poly}: Translation is faster than Exp and FOND4LTL$_f$, but still grows with problem size. This is expected, as Poly compiles LTL$_f$ formulas into AFAs using a polynomial-time procedure. However, its translation phase includes extensive post-processing that injects many spurious actions and auxiliary predicates into the domain, making it slower than TIDE’s lightweight translation. The main bottleneck remains the search phase, which performs poorly due to the size of the compiled problem and the lack of effective heuristics. Poly frequently times out in both benchmarks.

    \item \textbf{Plan4Past}: Translation times are similar to those of FOND4LTL$_f$, as it also compiles the formula into a DFA. However, search time increases rapidly with problem size in Benchmark~2. While it performs competitively on smaller problems, Plan4Past scales less effectively than TIDE in Benchmark~2.
\end{itemize}

\begin{figure}[h]
    \centering
    \begin{minipage}{0.48\linewidth}
        \centering
        \includegraphics[width=\linewidth]{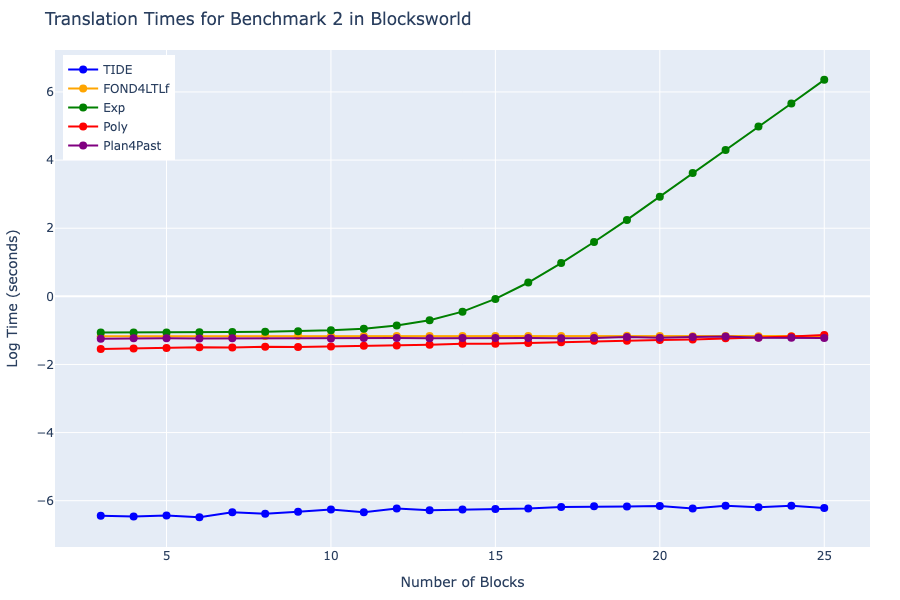}
        \caption*{(a) Translation times (log scale)}
        \label{fig:scaling_trans}
    \end{minipage}%
    \hspace{0.02\linewidth}
    \begin{minipage}{0.48\linewidth}
        \centering
        \includegraphics[width=\linewidth]{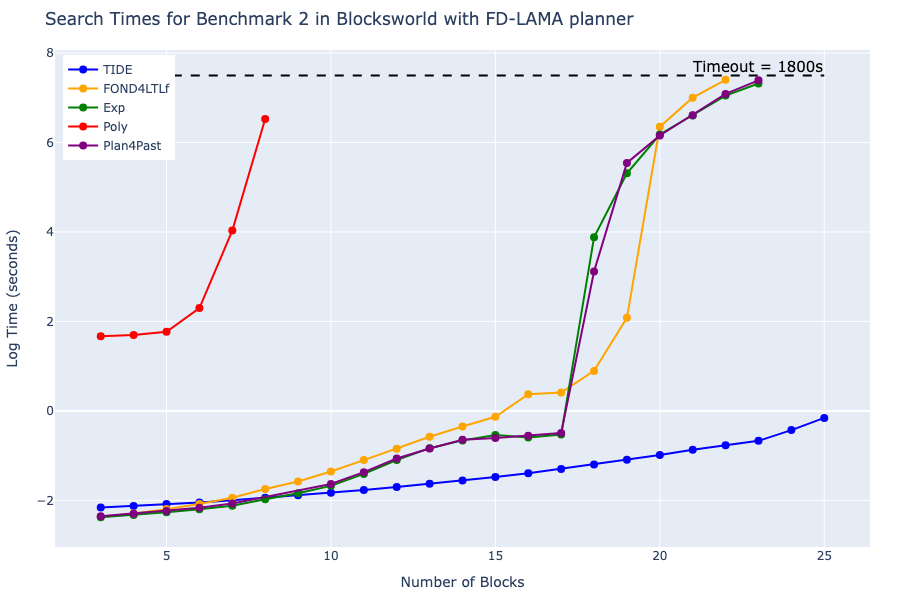}
        \caption*{(b) Search times (log scale)}
        \label{fig:scaling_search}
    \end{minipage}
    \caption{Scaling behavior of planning phases in Benchmark~2.\\
The x-axis shows the number of blocks in the problem (\( n \)); the y-axis indicates: \\
(a) translation time for converting the LTL$_f$/PPLTL goal into a final-state goal; \\
(b) search time for solving the resulting final-state goal using the FD-LAMA planner.}
    \label{fig:scaling_breakdown}
\end{figure}

These trends are further illustrated in Figure~\ref{fig:scaling_breakdown}, which separately plots translation and search times for Benchmark~2 across different methods.

\subsection{Evaluation on Backtracking-Heavy Openstacks Problems}
\label{sec:EvaluationonMyOpenstacksProblems}

This subsection evaluates the performance of TIDE, Exp, and Poly on a custom benchmark designed for the Openstacks domain. These problems were specifically crafted to expose a known weakness of TIDE: the handling of scenarios requiring extensive backtracking. While TIDE excels when the number of backtracking steps is small, its efficiency declines in cases involving numerous backtracking steps \textbf{with differing prefixes}, where cached results cannot be reused effectively.

\paragraph{Benchmark Details}
The custom Openstacks problems are designed to evaluate the performance of planning methods under constraints that require fulfilling a sequence of five orders while managing limited resources. These benchmarks impose dependencies between actions, ensuring that the orders are completed in specific sequences.

For example, an order can only be shipped after certain production stages are completed, and the start of an order may depend on the completion of previous orders. The problems also enforce constraints on resource usage, such as managing the availability of machines or stacks used for production and shipping. 

These problems are particularly challenging because they involve backtracking to explore alternative plans. As the number of dependencies and constraints increases, the need for backtracking grows significantly. This benchmark is designed to highlight TIDE’s strength in handling simpler scenarios with fewer backtracking steps but also exposes its limitations when extensive backtracking with unique prefixes is required, where caching cannot be effectively utilized.

\paragraph{Results and Discussion}

The first set of results, shown in Table~\ref{table:time_comparison_openstacks_custom}, reveal key insights into the performance of different planning approaches in backtracking-heavy Openstacks problems. In this table, TIDE is evaluated with both A* and FD-LAMA, while all competing methods (FOND4LTL$_f$, Exp, Poly, Plan4Past) are paired only with FD-LAMA, as FD-LAMA generally provides better heuristic guidance for these approaches. 

As expected, TIDE+A* outperforms all other methods in most cases, particularly in simpler problems such as \texttt{n01}--\texttt{n06} and \texttt{s01}--\texttt{s06}, where backtracking is minimal or where caching allows for effective reuse of computed subplans. TIDE+FD-LAMA remains competitive, though it is generally slower than TIDE+A*, reflecting the difference in how these planners approach subproblem solution.

However, as the complexity increases, TIDE's efficiency declines when excessive backtracking with differing prefixes is required. This effect is particularly pronounced in problems like \texttt{r02} and \texttt{r03}, where caching cannot be effectively leveraged, leading to a significant increase in planning time. 

To provide more insight into advantages of A* vs. FD-LAMA across multiple methods, we present Table~\ref{table:planner_comparison_openstacks_custom}, where TIDE, Poly, and FOND4LTL$_f$ are tested with both off-the-shelf planners. TIDE+A* consistently achieves the best performance across all problems, confirming that A* is a strong fit for TIDE’s trace-guided search. However, for Poly and FOND4LTL$_f$, A* performs poorly, as these methods generate large search spaces without effective heuristics to guide A*'s expansion. In contrast, the FD planner with LAMA search performs better for these approaches, as LAMA identifies landmarks—critical subgoals for planning—that compilation-based methods inherently lack. TIDE, however, formally generates such subgoals using the DFA trace, making its approach more efficient for A* search.

The second set of results, detailed in Table~\ref{table:tide_comparison_openstacks}, focuses on the performance of TIDE with the FD planner using LAMA search, comparing the impact of caching on planning times. We chose to evaluate caching with the FD planner because A* is extremely fast at solving subproblems in the Openstacks domain, and caching showed minimal impact on its performance. When TIDE is paired with A*, the bottleneck shifts to maintaining the uniform-cost search priority queue for potential candidate traces, rather than subproblem solving itself. Conversely, the FD planner with LAMA search is slower for Openstacks subproblems, making it easier to observe the performance improvements brought by caching. The columns \texttt{No Cache} and \texttt{With Cache} illustrate that caching significantly reduces planning times when prefix plans can be effectively reused (e.g., in the problem \texttt{p01}). Additionally, the table reports the average number of backtracking steps required to realize each automaton trace, providing further insight into when caching is most effective.

\begin{table}
\centering
\caption{Planning times (in seconds) for the custom Openstacks benchmark.\\ Bold values indicate the best performance for each problem.}
\label{table:time_comparison_openstacks_custom}
\begin{tabular}{c|cccccc}
\toprule
Problem & TIDE+A* & TIDE+LAMA & FOND4LTL$_f$ & Exp & Poly & Plan4Past\\
\midrule
n01 & \textbf{0.049} & 1.136 & 3.388 & 4.205 & 10.806 & 5.186 \\
n02 & \textbf{0.258} & 3.032 & 24.400 & 5.079 & 10.482 & 5.657 \\
n03 & \textbf{1.345} & 4.764 & 63.790 & 5.207 & 12.267 & 5.541 \\
n04 & \textbf{2.870} & 6.731 & failure & 5.225 & 17.569 & 5.759 \\
n05 & \textbf{5.490} & 9.925 & failure & \textbf{5.420} & 50.310 & 5.827 \\
n06 & 11.109 & 15.987 & failure & \textbf{5.576} & 183.233 & 6.007 \\
s01 & \textbf{0.088} & 2.713 & 25.278 & 4.936 & 8.852 & 5.199 \\
s02 & \textbf{0.073} & 3.177 & 35.508 & 5.149 & 7.666 & 5.190\\
s03 & \textbf{0.239} & 3.222 & 37.117 & 4.948 & 10.346 & 5.252 \\
s04 & \textbf{0.128} & 3.238 & 57.316 & 5.148 & 9.083 & 5.164 \\
s05 & \textbf{0.185} & 4.062 & 90.419 & 5.209 & 9.485 & 5.290 \\
s06 & \textbf{1.006 } & 5.631 & 288.334 & 5.224 & 10.384 & 5.393\\
p01 & \textbf{0.231} & 5.927 & failure & 5.360 & 248.845 & 5.413\\
r01 & 10.378 & 17.091 & 1017.85 & \textbf{5.316} & 17.671 & 5.428\\
r02 & 54.490 & 66.228 & 2160.31 & \textbf{5.441} & 40.561 & 5.540\\
r03 & 228.449 & 244.183 & timeout & \textbf{5.597} & 141.193 & 5.724\\
\bottomrule
\end{tabular}
\end{table}

\begin{table}
\centering
\caption{Planning times (in seconds) for the custom Openstacks benchmark.\\ 
Comparison of TIDE, Poly, and FOND4LTL$_f$ when paired with either the A* or FD-LAMA planner. 
Bold values indicate the best performance for each problem.}
\label{table:planner_comparison_openstacks_custom}
\begin{tabular}{c|ccc|ccc}
\toprule
Problem & \multicolumn{3}{c|}{A* planner} & \multicolumn{3}{c}{FD-LAMA planner} \\
 & TIDE & Poly & FOND4LTL$_f$ & TIDE & Poly & FOND4LTL$_f$ \\
\midrule
n01 & \textbf{0.049} & 6.526 & 0.579 & 1.136 & 10.806 & 3.388 \\
n02 & \textbf{0.258} & 18.349 & 1.508 & 3.032 & 10.482 & 24.400 \\
n03 & \textbf{1.345} & 47.826 & 3.088 & 4.764 & 12.267 & 63.790 \\
n04 & \textbf{2.870} & 166.538 & failure & 6.731 & 17.569 & failure \\
n05 & \textbf{5.490} & 619.631 & failure & 9.925 & 50.310 & failure \\
n06 & \textbf{11.109} & timeout & failure & 15.987 & 183.233 & failure \\
s01 & \textbf{0.088} & 8.852 & 1.926 & 2.713 & 8.852 & 25.278 \\
s02 & \textbf{0.073} & 7.666 & 2.707 & 3.177 & 7.666 & 35.508 \\
s03 & \textbf{0.239} & 10.346 & 2.415 & 3.222 & 10.346 & 37.117 \\
s04 & \textbf{0.128} & 9.083 & 4.292 & 3.238 & 9.083 & 57.316 \\
s05 & \textbf{0.185} & 9.485 & 6.725 & 4.062 & 9.485 & 90.419 \\
s06 & \textbf{1.006} & 10.384 & 17.011 & 5.631 & 10.384 & 288.334 \\
p01 & \textbf{0.231} & timeout & failure & 5.927 & 248.845 & failure \\
r01 & \textbf{10.378} & 216.944 & 79.768 & 17.091 & 17.671 & 1017.85 \\
r02 & 54.490 & 722.209 & 161.025 & 66.228 & \textbf{40.561} & 2160.31 \\
r03 & 228.449 & timeout & timeout & 244.183 & \textbf{141.193} & timeout \\
\bottomrule
\end{tabular}
\end{table}

\begin{table}
\centering
\caption{Planning times (in seconds) for TIDE on the custom Openstacks benchmark, paired with the FD planner using LAMA search. The table compares performance with and without caching, while the ``Backtracking Steps'' column indicates the average number of backtracking steps required to solve each problem.}
\label{table:tide_comparison_openstacks}
\begin{tabular}{c|cc|c}
\toprule
Problem & No Cache & With Cache & Backtracking Steps\\
\midrule
n01  & 1.137 & 1.136 & 0 \\
n02  & 3.668 & 3.032 & 2.68 \\
n03  & 6.752 & 4.764 & 4.42 \\
n04  & 10.619 & 6.731 & 5.82 \\
n05  & 15.122 & 9.925 & 6.46 \\
n06  & 15.987 & 23.147 & 7.98 \\
s01  & 3.390 & 2.713 & 5.28 \\
s02  & 4.970 & 3.177 & 6.06 \\
s03  & 3.436 & 3.222 & 10.24 \\
s04  & 4.277 & 3.238 & 7.08 \\
s05  & 5.789 & 4.062 & 8.10 \\
s06  & 9.461 & 5.631 & 9.14 \\
p01  & 14.392 & 5.927 & 15.48 \\
r01  & 33.123 & 17.091 & 21.92 \\
r02  & 92.472 & 66.228 & 30.12 \\
r03  & 289.170 & 244.183 & 42.26 \\
\bottomrule
\end{tabular}
\end{table}

Overall, the results demonstrate that caching improves efficiency when prefixes can be reused, but it does not resolve the inherent challenges of scenarios requiring extensive backtracking with diverse prefixes.

% In our future work, to address this limitation, which stems from the ``exploration vs. exploitation'' trade-off, we aim to implement a parallel search strategy that explores multiple DFA traces simultaneously. Since the realization process for each automaton trace is independent, this approach could reduce the impact of exploring incorrect DFA traces with varied prefixes.

%% file: conclusion.tex
\section{Conclusion}
\label{ch:Conclusion}

\subsection{Summary}
This thesis presented \textit{TIDE (Trace-Informed Depth-first Exploration)}, a novel planning framework for solving planning problems with temporally extended goals (TEGs) expressed in LTL$_f$ in the fully observable, deterministic domains. Unlike traditional methods that construct and search the full product graph between a planning domain and a goal automaton, TIDE incrementally explores this space by focusing on individual DFA traces. It decomposes each trace into a sequence of reach-avoid subproblems, which are then solved using off-the-shelf planners such as A* or Fast Downward with LAMA search.

Through this trace-guided approach, TIDE significantly improves scalability by avoiding full product construction and leveraging trace structure to guide search. Our experiments demonstrate that TIDE outperforms state-of-the-art methods such as Exp, Poly, FOND4LTL$_f$, and Plan4Past across a variety of domains, including the standard TB15 benchmarks, two custom Blocksworld scaling suites, and a custom Openstacks benchmark designed to evaluate backtracking behavior. This shows that TIDE is a valuable addition to the portfolio of planning methods for temporally extended goals.

\subsection{Future Work}
There are several promising directions for extending this work:

\begin{itemize} \item \textbf{Parallel exploration of DFA traces.} One limitation of the current TIDE framework is its sequential search over DFA traces. In cases where failed traces have distinct prefixes, caching provides limited benefit, as subplans cannot be reused. A parallel trace exploration strategy—where multiple candidate traces are explored concurrently—could significantly reduce total planning time. Since the realization of each trace is independent, this approach is naturally well-suited to multicore or distributed systems.

\item \textbf{Extension to FOND domains.} While this thesis focuses on fully observable deterministic domains, TIDE’s approach could be generalized to fully observable non-deterministic (FOND) settings. In such cases, each trace could be realized by decomposing into a structure of FOND reach-avoid subproblems. Existing FOND planners, such as MyND~\cite{mattmuller_pattern_2010}, could be used to solve these subproblems. 

\item \textbf{On-the-fly automaton construction.} Recent advances in LTL$_f$ synthesis suggest that it is possible to avoid constructing the entire DFA up front. For example, Xiao et al.~\cite{xiao_--fly_2021} and Di Giacomo et al.~\cite{giacomo_ltlf_2022} propose forward-search approaches that generate only relevant portions of the automaton on demand using SAT-based encodings or Sentential Decision Diagrams (SDDs). Adapting such ideas to TIDE could reduce the cost of automaton compilation, especially for complex LTL$_f$ goals.

\item \textbf{Heuristic improvements.}
TIDE can benefit from heuristic improvements at both levels of its framework—tightening the feedback loop between high-level trace selection and low-level subproblem planning:

\begin{itemize} \item \textit{Trace selection heuristics.} At the high level, TIDE selects a single DFA trace to realize at a time. Currently, this selection is based on a generic cost model, but the process is highly customizable. TIDE provides a natural mechanism for assigning or updating heuristic costs on DFA transitions, making it easy to inject domain-specific knowledge. For example, we can penalize transitions with undesirable conditions (e.g., ``risky'' predicates) or promote transitions with desirable conditions (preferences). 

\item \textit{Planner-level heuristics.} At the lower level, once a trace is selected, it is realized as a classical planning subproblem solved using existing off-the-shelf planners (e.g., A* or Fast Downward). These planners typically operate independently of the temporal goal’s structure. A promising direction is to develop a modified version of these planners—a “specialized” off-the-shelf planner—that incorporates DFA progress information (e.g., the current automaton state and the set of neighboring states) into the heuristic function. By incorporating this trace-specific knowledge, such a planner could offer a more focused search.
\end{itemize}
\end{itemize}

%% file: ack.tex
\section{Acknowledgments}
\label{ch:ack}

This work has been supported in part by NSF-IIS-1830549,
NSF ITR-21-27309, and Rice University Funds. 

%% file: appendix.tex
\clearpage
\onecolumn
\appendix
\textbf{\Large \center Appendix}
\section{Pseudocode}

This section contains the pseudocodes used in the proposed method.

\begin{algorithm}
\caption{High-Level Pseudocode for TIDE Approach}
\begin{algorithmic}[1]
\State \textbf{Initialize:} DFA start state and priority queue for DFA traces
\While{not all product states are visited}
    \If{no more DFA traces available} \Comment{All traces have been explored}
        \State \textbf{Terminate Search}
    \EndIf
    \State $dfa\_trace \gets generate\_dfa\_trace()$: select the most promising DFA trace
    \State $plan \gets realize\_dfa\_trace(dfa\_trace)$: attempt to realize the selected DFA trace in the domain
    \If{$plan$ is found} \Comment{Valid plan has been found}
        \State \textbf{Return:} $plan$
    \EndIf
\EndWhile
\State \textbf{Return:} \textbf{null}
\end{algorithmic}
\end{algorithm}

\begin{algorithm}
\caption{\textit{generate\_dfa\_trace()}: Modified Uniform Cost Search for Selecting a DFA Trace}
\begin{algorithmic}[1]
\Require Priority queue of DFA traces, each with an associated trace cost
\State \textbf{Initialize:} $best\_trace \gets \text{empty trace}$, $best\_trace.cost \gets \infty$
\While{priority queue is not empty}
    \State $current\_trace \gets $ extract the trace with the lowest cost from the queue
    \State $current\_dfa\_state \gets $ get the end DFA state in this trace
    \If{$current\_dfa\_state$ is an accepting state}
        \If{$current\_trace.cost < best\_trace.cost$} 
            \If{$best\_trace$ is not empty}
                \State Add $best\_trace$ back into the priority queue
            \EndIf
            \State $best\_trace \gets current\_trace$ \Comment{Hill Climbing Component}
        \Else \Comment{We found an accepting trace with a larger or equal cost than before}
            \State Add $current\_trace$ back into the priority queue
            \State \textbf{Return} $best\_trace$ \Comment{Return a previously found accepting trace}
        \EndIf
        \State \textbf{Continue} to explore other traces (we don't expand accepting DFA states)
    \EndIf
    \For{each transition edge from $current\_dfa\_state$} \Comment{Expand the trace}
        \State $next\_dfa\_state \gets$ get the destination DFA state from the edge
        \If{$next\_dfa\_state \neq current\_dfa\_state$}  
            \State $extended\_trace \gets$ create a new trace that includes this new DFA edge
            \State $extended\_trace.cost \gets$  calculate the cost of this trace
            \State Add $extended\_trace$ to the priority queue
        \EndIf
    \EndFor
\EndWhile
\State \textbf{Return:} $best\_trace$, or \textbf{null} if none is found
\end{algorithmic}
\end{algorithm}

\begin{algorithm}
\caption{\textit{realize\_dfa\_trace\_without\_planner($dfa\_trace$)}: Realize a DFA trace in the domain with hierarchical Breadth-First Search }
\begin{algorithmic}[1]
\Require $dfa\_trace$: a DFA path from start state to the accepting state
\State \textbf{Initialize:} $queuesMap \gets \text{a map of queues for each DFA state in $dfa\_trace$}$
\State \textbf{Initialize:} $visited \gets \{\}$
\State \textbf{Initialize:} $start\_product\_state \gets(start\_domain\_state, start\_dfa\_state)$ 
\newline \Comment{Cartesian product of domain and DFA start states}
\newline \Comment{Queue $queuesMap[start\_dfa\_state]$ represents a Breadth-First Search for realizing a DFA transition from $start\_dfa\_state$}
\newline \Comment{Initialize BFS for this DFA state}
\State Push $start\_product\_state$ onto the $queuesMap[start\_dfa\_state]$ queue 
\State Add $start\_product\_state$ to the $visited$ set
\While{the queue for the $start\_dfa\_state$ is not empty} 
    \State $current\_queue \gets queuesMap[current\_dfa\_state]$ 
    \If{$current\_queue$ is empty} \Comment{BFS failed to find a solution}
        \State Backtrack to the previous DFA state 
        \State \textbf{continue}
    \EndIf
    \State $current\_product\_state \gets current\_queue.pop()$  
    \If{$current\_product\_state$ has not been expanded} 
        \State Generate successors for $current\_product\_state$
    \EndIf
    \For{each valid transition from $current\_product\_state$}
        \State Get the $next\_product\_state = (next\_domain\_state, next\_dfa\_state)$
        \If{$next\_product\_state$ has been visited}
            \State \textbf{continue} \Comment{Skip already visited states}
        \EndIf
        \If{$next\_dfa\_state \ne current\_dfa\_state$}
            \State Cache this path for future reuse
            \State Update cost for this transition to $SUCCESS\_COST$
            \If{$next\_dfa\_state$ is not the immediate next state in our $dfa\_trace$}
                \State \textbf{continue} \Comment{Handle it as an obstacle}
            \Else  \Comment{A desired DFA transition was realized}
            \State $current\_dfa\_state \gets next\_dfa\_state$
            \If{$next\_dfa\_state$ is the final accepting state in $dfa\_trace$}
                \State \textbf{Return:} the full path
            \EndIf
            \EndIf
        \EndIf

        \State Push $next\_product\_state$ onto the $queuesMap[next\_dfa\_state]$ queue
        \State Add $next\_product\_state$ to the $visited$ set
    \EndFor
\EndWhile
\State Update cost for the failed transition to $FAILURE\_COST$
\State \textbf{Return:} \textbf{null} \Comment{Failed to realize the trace}
\end{algorithmic}
\end{algorithm}

\begin{algorithm}
\caption{\textit{realize\_dfa\_trace\_with\_planner($dfa\_trace$)}: Realize a DFA trace using an off-the-shelf planner}
\begin{algorithmic}[1]
\Require $dfa\_trace$: a DFA path from start state to the accepting state
\State \textbf{Initialize:} $subproblemsMap \gets \{\} \text{, a map of reachability subproblems for each DFA state}$
\State \textbf{Initialize:} $visited \gets \{\}$
\State \textbf{Initialize:} $start\_product\_state \gets(start\_domain\_state, start\_dfa\_state)$ 
\newline \Comment{Cartesian product of domain and DFA start states}
\newline \Comment{$subproblemsMap[start\_dfa\_state]$ is a classical problem that needs to be solved to realize a DFA transition from $start\_dfa\_state$}
\State $start\_subproblem \gets $ create a subproblem for the first DFA transition in $dfa\_trace$
\State $subproblemsMap[start\_dfa\_state] \gets start\_subproblem$
\State Add $start\_product\_state$ to the $visited$ set
\While{$subproblemsMap$ is not empty}
    \State $current\_subproblem \gets subproblemsMap[current\_dfa\_state]$ 
    \State $plan \gets planner.solve(current\_subproblem)$  \Comment{Use an off-the-shelf planner}
     \If{$plan$ is \textbf{null}} \Comment{No solution is found}
        \State Remove $current\_dfa\_state$ from $subproblemsMap$
        \State Backtrack to the previous DFA state
        \State \textbf{continue}
    \EndIf
    \If{$plan$ does not satisfy the edge condition for $(current\_dfa\_state \rightarrow next\_dfa\_state)$ transition}
    \State \textbf{continue} 
    \EndIf
    \State Update cost for $(current\_dfa\_state \rightarrow next\_dfa\_state)$ transition to $SUCCESS\_COST$
    \If{$next\_dfa\_state$ is the final accepting state in $dfa\_trace$}
        \State \textbf{Return:} the full path
    \EndIf
    \State $end\_domain\_state \gets$ final state of $plan$
    \State $next\_product\_state \gets (end\_domain\_state, next\_dfa\_state)$
    \State Add $next\_product\_state$ to $visited$
    \State $next\_subproblem \gets $ create a subproblem for the next DFA transition from $next\_dfa\_state$ in $dfa\_trace$
\State $subproblemsMap[next\_dfa\_state] \gets next\_subproblem$
\State $current\_dfa\_state \gets next\_dfa\_state$
\EndWhile
\State Update cost for the failed transition to $FAILURE\_COST$
\State \textbf{Return:} \textbf{null} \Comment{Failed to realize the trace}
\end{algorithmic}
\end{algorithm}